\documentclass[10pt,twocolumn,letterpaper]{article}

\usepackage{cvpr}              %
\usepackage{tikz}
\usepackage{comment}
\usepackage{dblfloatfix}
\usepackage{bbm}

\usepackage{pifont}%
\newcommand{\cmark}{\ding{51}}%
\newcommand{\xmark}{\ding{55}}%

\usepackage{graphicx}
\graphicspath{ {images/} }

\usepackage{multirow}

\definecolor{cvprblue}{rgb}{0.21,0.49,0.74}
\usepackage[pagebackref,breaklinks,colorlinks,citecolor=cvprblue]{hyperref}

\def\paperID{15452} %
\def\confName{CVPR}
\def\confYear{2024}

\title{A Unified and Interpretable Emotion Representation and Expression Generation}
\author{\!\!\!Reni Paskaleva$^{3*}$, Mykyta Holubakha$^{1}$, Andela Ilic$^{2}$, Saman Motamed$^{1}$, Luc Van Gool$^{1,2}$, Danda Paudel$^{1}$\\
$^{1}$INSAIT, Sofia University, Bulgaria\hspace{5mm} $^{2}$ETH Zurich, Switzerland \\ $^{3}$First Private Mathematical High School, Sofia, Bulgaria\\
{\tt\small firstname.lastname@insait.ai, anilic@student.ethz.ch}
}

\begin{document}
\twocolumn[{%
\renewcommand\twocolumn[1][]{#1}%
\maketitle
\begin{center}
    \centering
    \captionsetup{type=figure}
 \vspace{-7mm}
 {
 \addtolength{\tabcolsep}{-4.5pt}
\begin{tabular} {c||ccccc||cc|}
 \cline{2-8} & \multicolumn{5}{|c||}{\textbf{Compared methods}}  & \multicolumn{2}{|c|}{\textbf{3D model (ours)}} \\ \cline{2-8}
 & DreamBooth & DALL.E 3 & {Stable Diffusion} & AUs model  & 2D model &  & +X \\
 \rotatebox{90}{Happy-Fearful} &
 \includegraphics[width=0.130\linewidth]{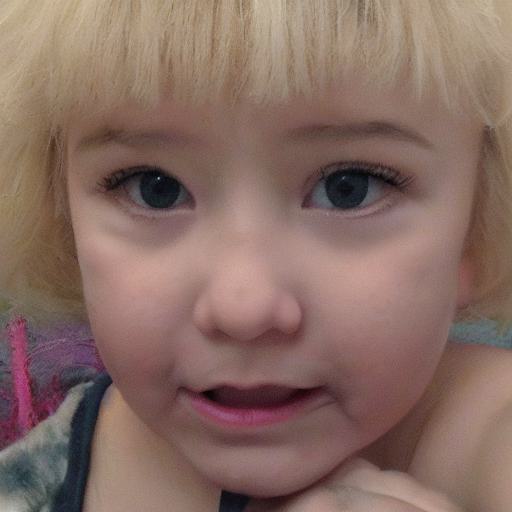}&
 \includegraphics[width=0.130\linewidth]{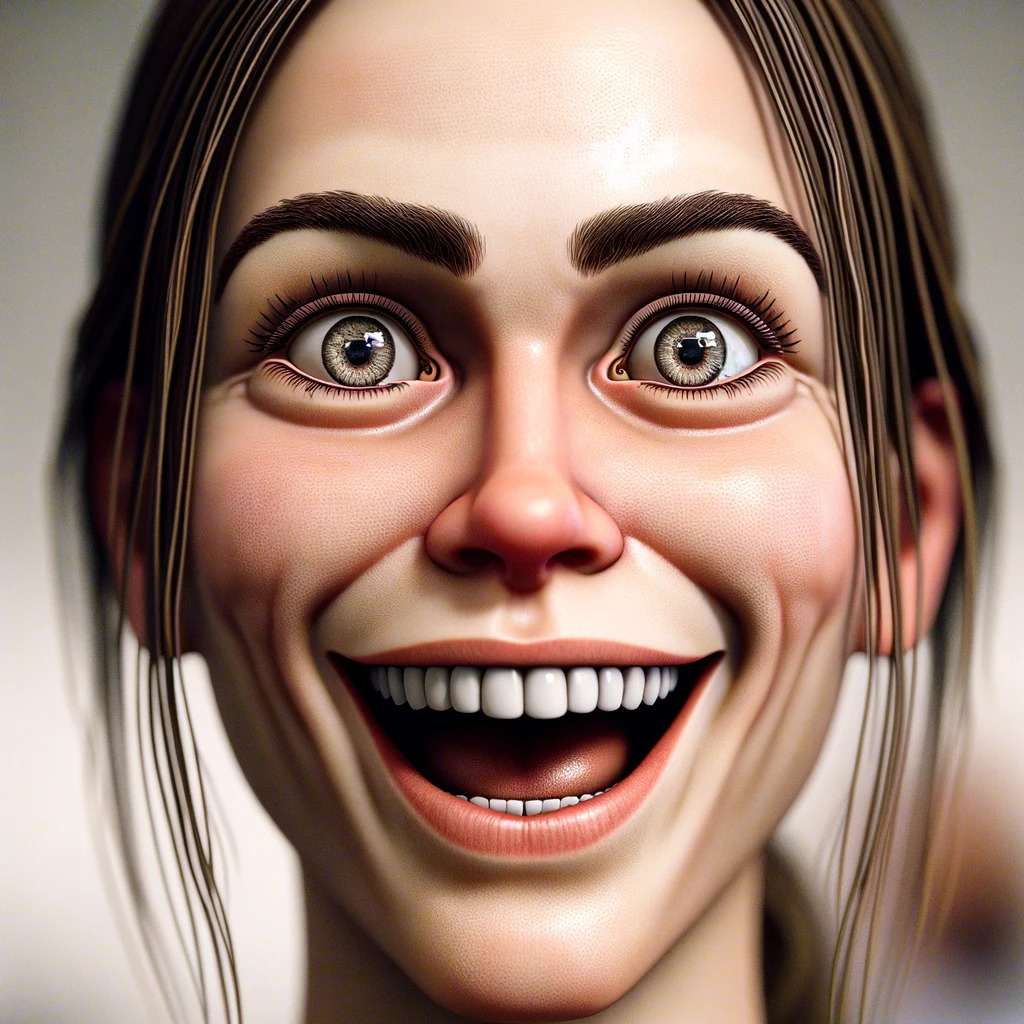}&
 \includegraphics[width=0.130\linewidth]{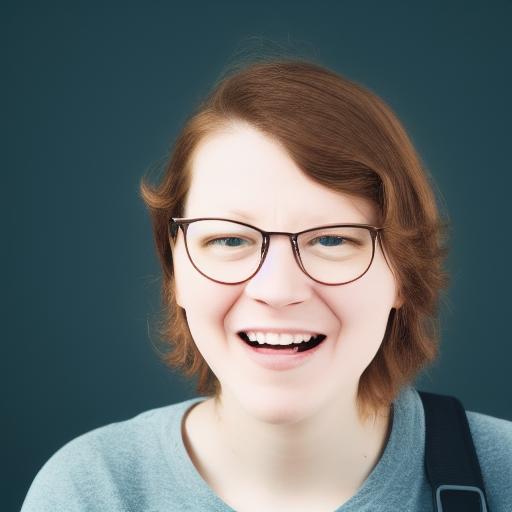}&

\includegraphics[width=0.130\linewidth]{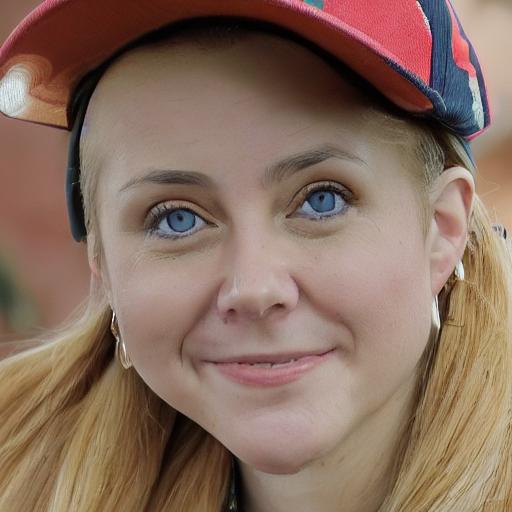}&
 \includegraphics[width=0.130\linewidth]{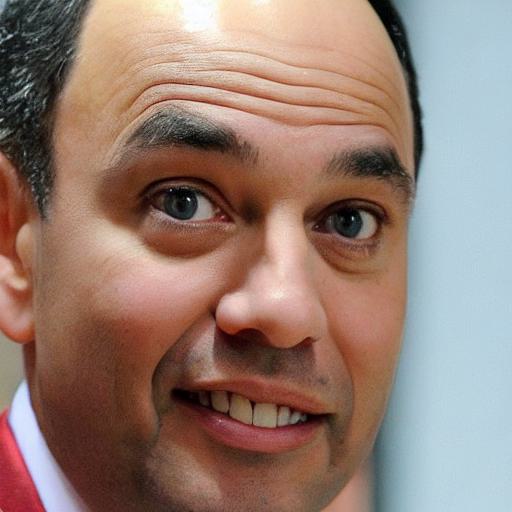}&
 
 \includegraphics[width=0.130\linewidth]{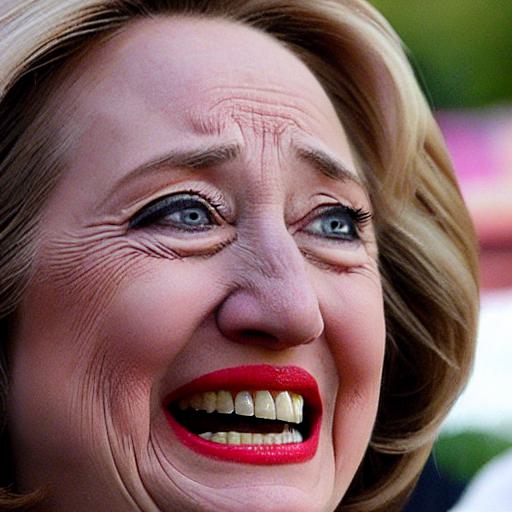}&
 \includegraphics[width=0.130\linewidth]{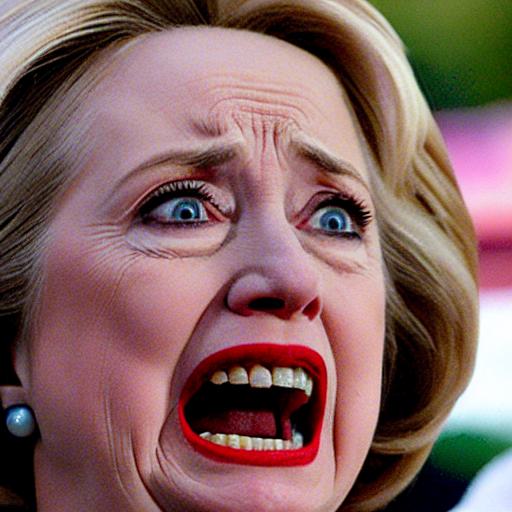} \\
 
 \rotatebox{90}{Surprised-Sad} &
 \includegraphics[width=0.130\linewidth]{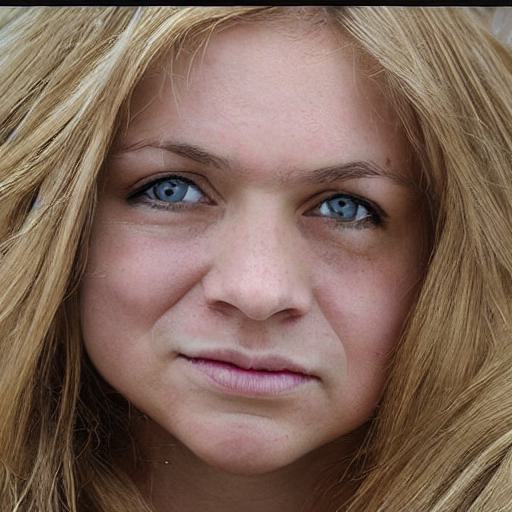}&
 \includegraphics[width=0.130\linewidth]{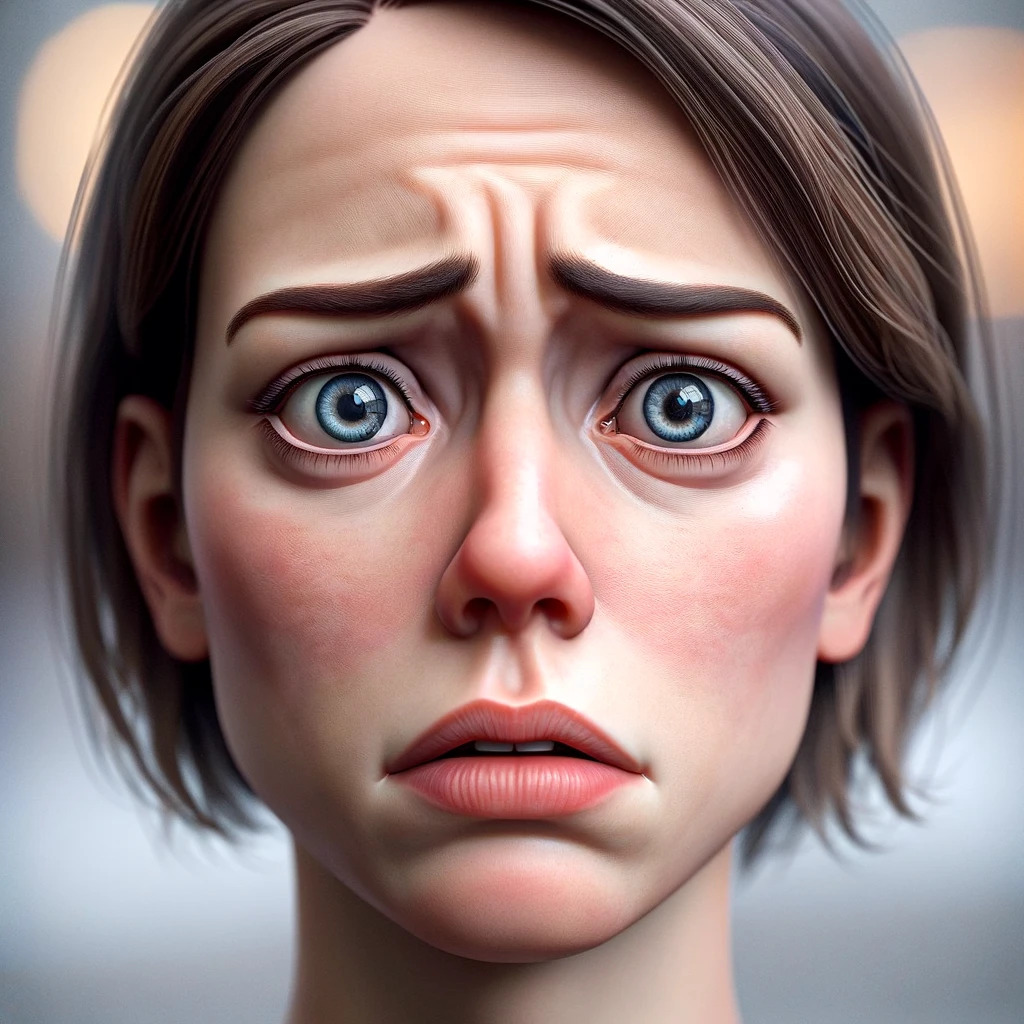}&
 \includegraphics[width=0.130\linewidth]{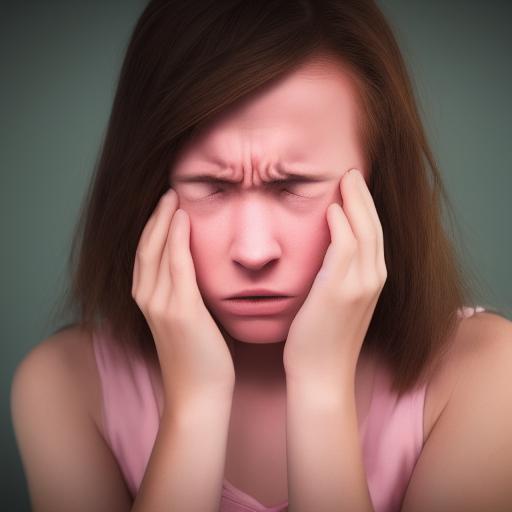}&
 
 \includegraphics[width=0.130\linewidth]{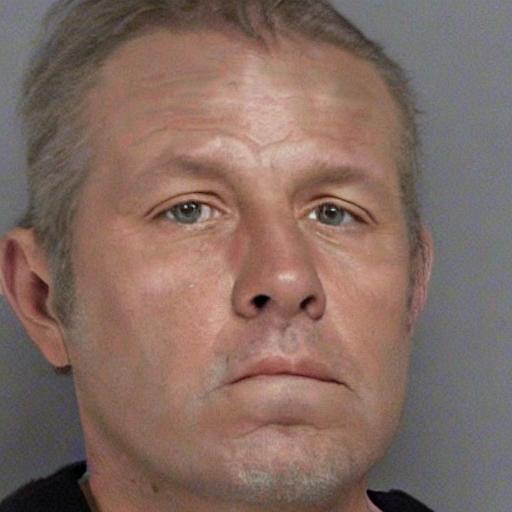}&
 \includegraphics[width=0.130\linewidth]{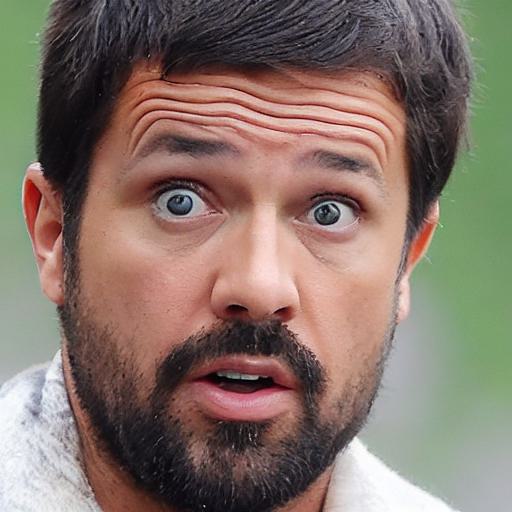}&
 
 \includegraphics[width=0.130\linewidth]{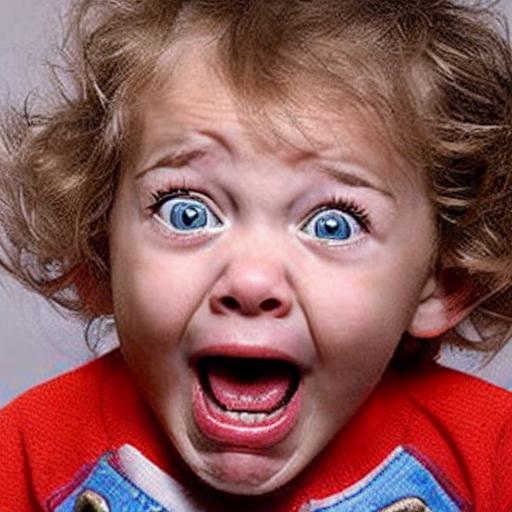}&
 \includegraphics[width=0.130\linewidth]{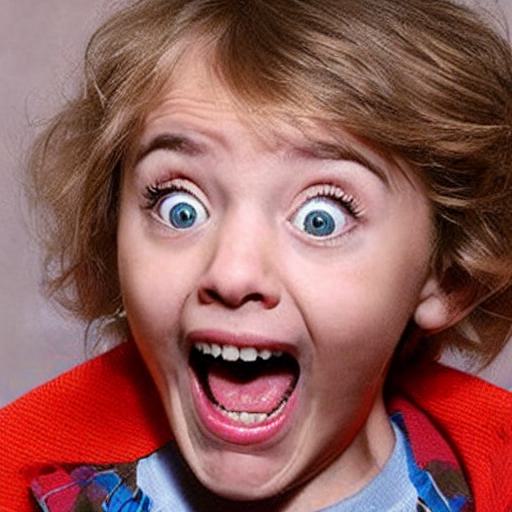} \\
 
 \rotatebox{90}{Sad-Angry} &
 \includegraphics[width=0.130\linewidth]{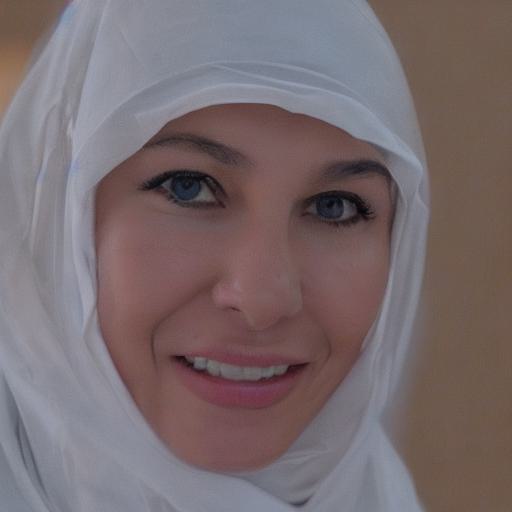}&
 \includegraphics[width=0.130\linewidth]{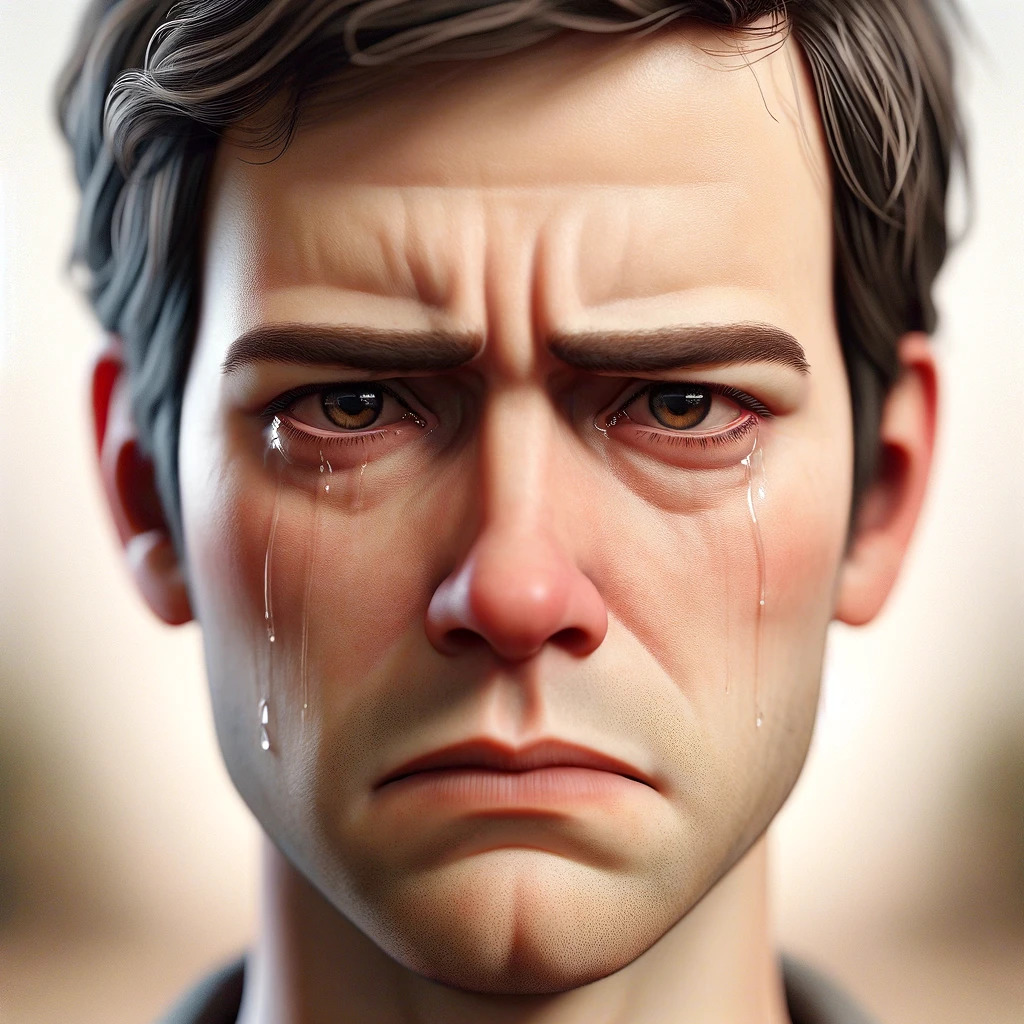}&
 \includegraphics[width=0.130\linewidth]{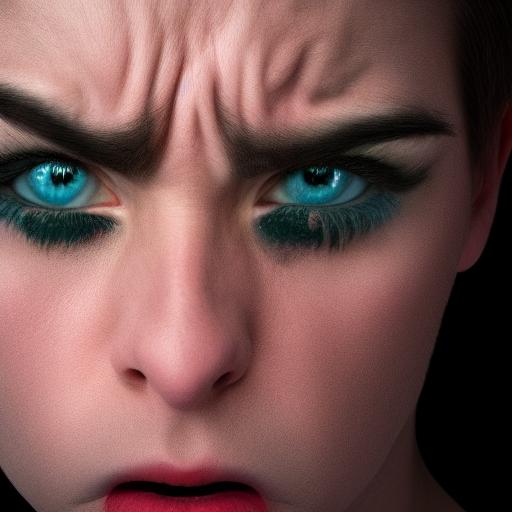}&
 \includegraphics[width=0.130\linewidth]{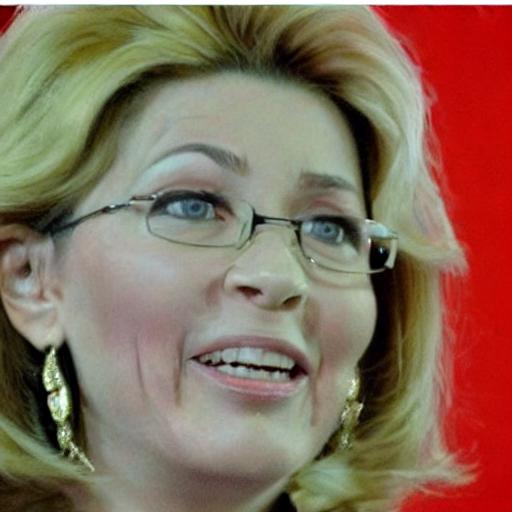}&
 \includegraphics[width=0.130\linewidth]{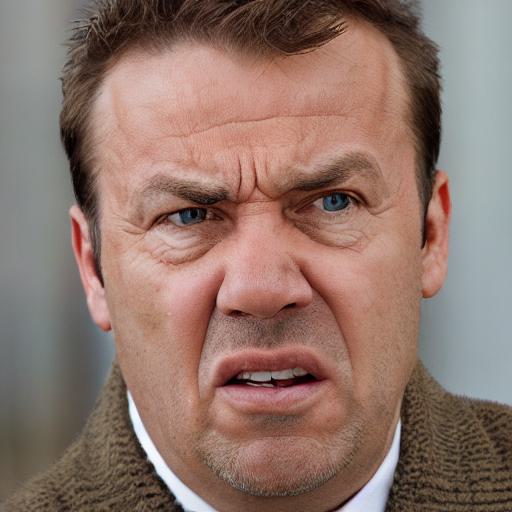}&
 
 \includegraphics[width=0.130\linewidth]{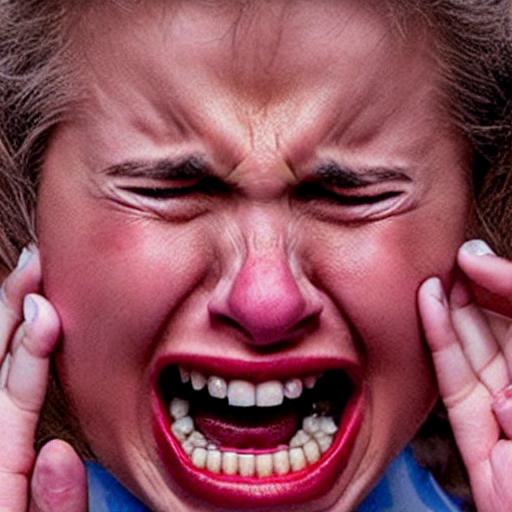}&
 \includegraphics[width=0.130\linewidth]{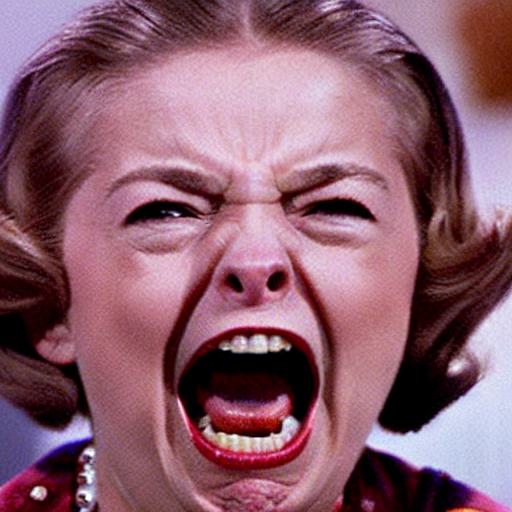} \\
 
 \rotatebox{90}{Fearful-Disgd.} &
 \includegraphics[width=0.130\linewidth]{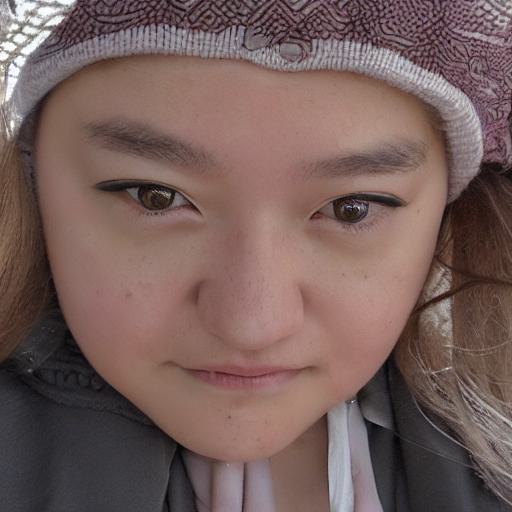}&
 \includegraphics[width=0.130\linewidth]{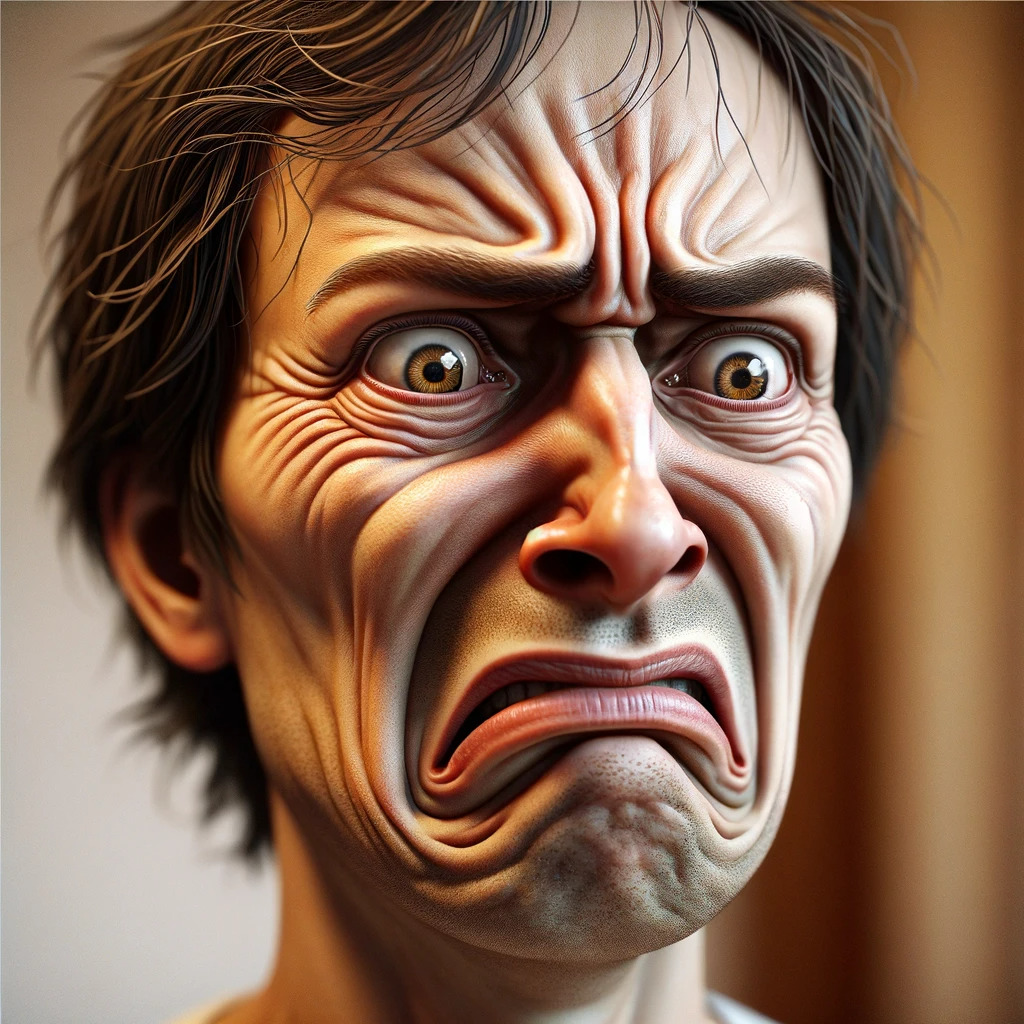}&
 \includegraphics[width=0.130\linewidth]{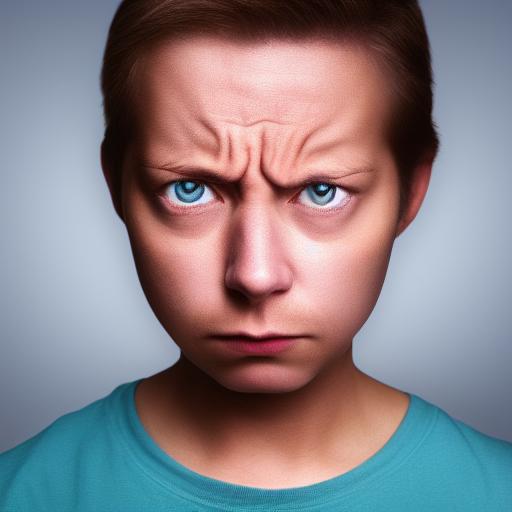}&
 \includegraphics[width=0.130\linewidth]{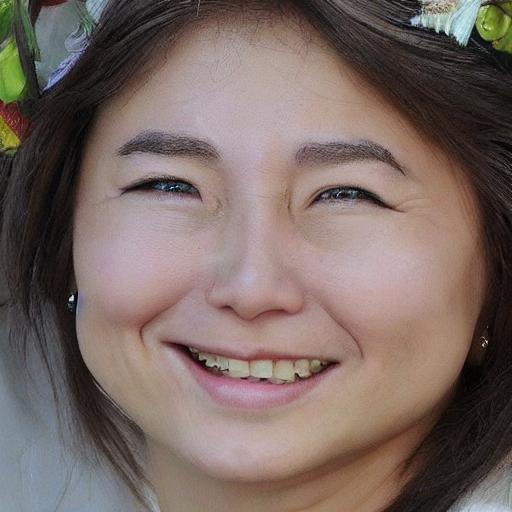}&
 \includegraphics[width=0.130\linewidth]{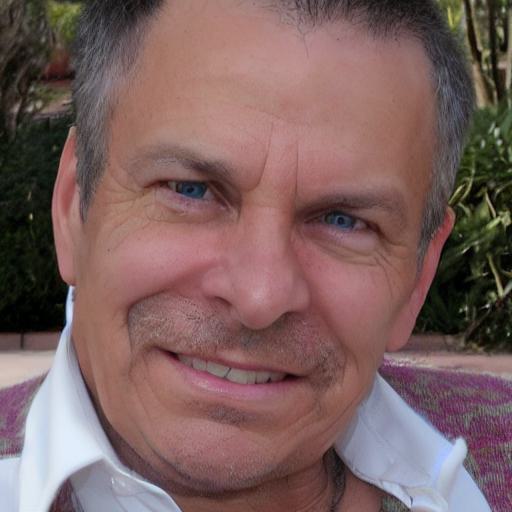}&

\includegraphics[width=0.130\linewidth]{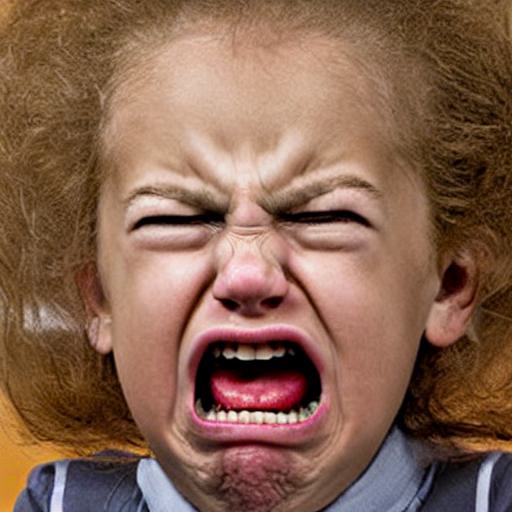}&

 \includegraphics[width=0.130\linewidth]{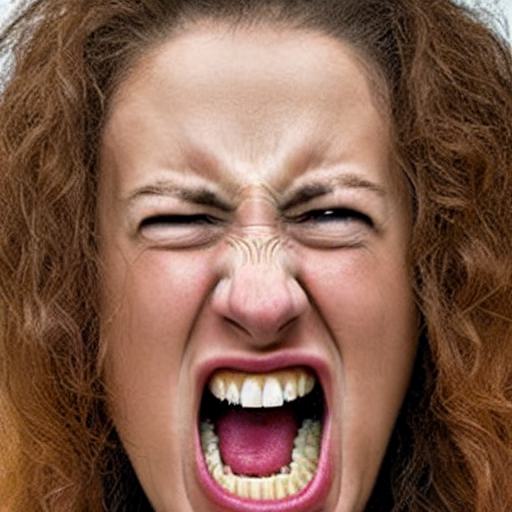}
\end{tabular}
    }
\vspace{-3mm}
    \captionof{figure}{\label{fig:teaser} We show the capability of our continuous 3D-representation-based expression generation method in generating rich and compound expressions. An extra arbitrarily  chosen  expression component (+X) is added to the targeted compound on the left. The proposed 3D model performs the best compared to the 2D model and other competing methods. Our model shares the same settings with DreamBooth. 
    }
    \vspace{-2mm}
\end{center}%
 }]
 
\maketitle
\let\thefootnote\relax\footnotetext{\hspace{-5mm}Project \& code: \url{https://emotion-diffusion.github.io/}\\ $^*$This work was done as a part of INSAIT internship.}
\begin{abstract}
Canonical emotions, such as happy, sad, and fearful, are easy to understand and annotate. However, emotions are often compound, e.g. happily surprised, and can be mapped to the action units (AUs) used for expressing emotions, and trivially to the canonical ones. Intuitively, emotions are continuous as represented by the arousal-valence (AV) model. An interpretable unification of these four modalities —namely, Canonical, Compound, AUs, and AV— is highly desirable, for a better representation and understanding of emotions. However, such unification remains to be unknown in the current literature. 
In this work, we propose an interpretable and unified emotion model, referred as \emph{C2A2}. We also develop a method that leverages labels of the non-unified models to annotate the novel unified one. Finally, we modify the text-conditional diffusion models to understand continuous numbers, which are then used to generate continuous expressions using our unified emotion model. Through quantitative and qualitative experiments, we show that our generated images are rich and capture subtle expressions. Our work allows a fine-grained generation of expressions in conjunction with other textual inputs and offers a new label space for emotions at the same time. 

\end{abstract}

\begin{figure*}[!htp]
\begin{tikzpicture}
\node at (0,0){
\scriptsize
\begin{tabular}{
|p{1.45cm}|p{1.0cm}||p{1.6cm}|p{1.2cm}| }
 \hline
\textbf{Category}&\textbf{AUs} &\textbf{Category} &\textbf{AUs}\\
 \hline
\hline
Happy&12,25&Sadly disgd.&4,10\\
\hline
Sad&4,15&Fearfully angry&4,20,25\\
\hline
Fearful&1,4,20,25&Fearfully surpd.&1,2,5,20,25\\
\hline
Angry&4,7,24&Fearfully disgd.&1,4,10,20,25\\
\hline
Surprised&1,2,25,26&Angrily surpd.&4,25,26\\
\hline
Disgusted&9,10,17&Disgd. surpd.&1,2,5,10\\
\hline
Happily sad&4,6,12,25&Happily fearful&1,2,12,25,26\\
\hline
Happily surpd.&1,2,12,25&Angrily disgd.&4,10,17\\
\hline
Happily disgd.&10,12,25&Awed&1,2,5,25\\
\hline
Sadly fearful&1,4,15,25&Appalled&4,9,10\\
\hline
Sadly angry&4,7,15&Hatred&4,7,10\\
\hline
Sadly surpd.&1,4,25,26&-&-\\
\hline
\end{tabular}};
\node at (6.3,0) {\includegraphics[width=0.28\linewidth]{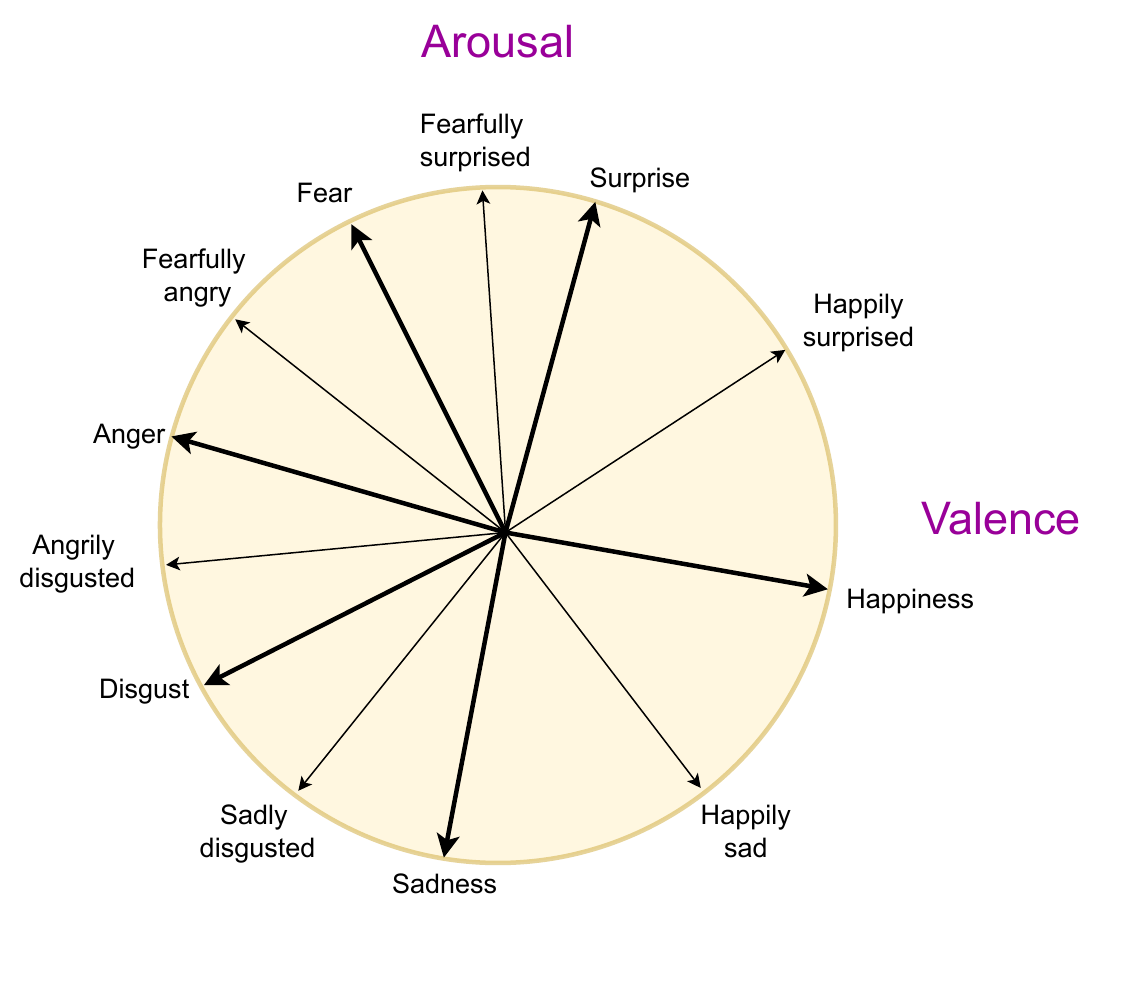}};
\node at (11.25,0) {\includegraphics[width=0.3\linewidth]{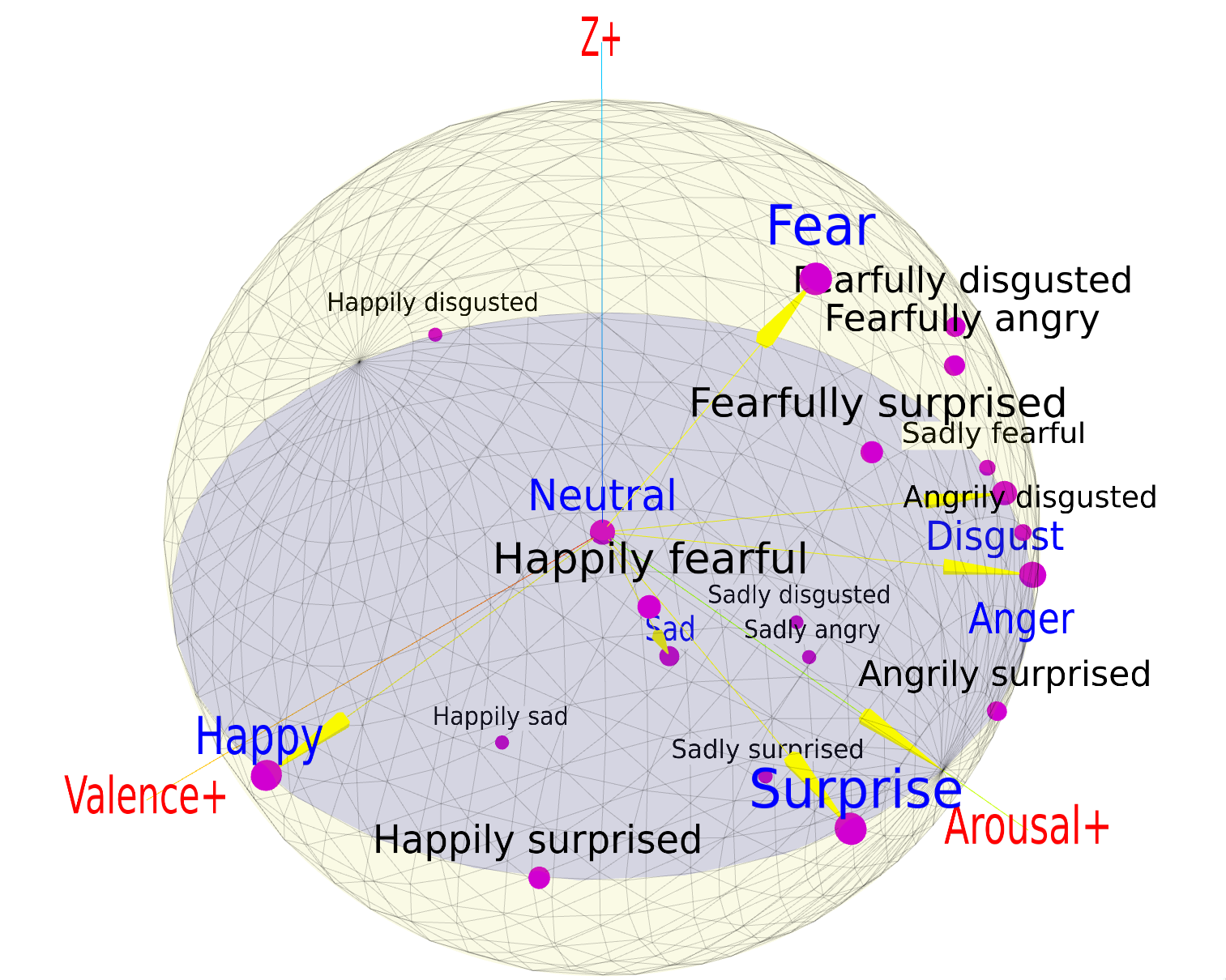}};
\end{tikzpicture}
\vspace{-7mm}
\caption{The compound emotion model on the left unifies the categorical emotions and the AUs based expressions~\cite{du2014compound}. The continuous emotion model of arousal-valence (middle) allows the mapping of some of the categorical emotion on the continuous space~\cite{PADrusselmehrabis}. The proposed 3D-based emotion modelling largely unifies the both thereby allowing more combination of the compound emotions (right).   \label{fig:introFig} }
\end{figure*}

\section{Introduction}
\label{sec:intro}

Expressing emotions affects our lives by playing a vital role in day-to-day communications. We are interested in facial expressions, which are a primary means of such communication. Therefore, a generative model must generate realistic expressions for human-like communications. Human emotions and expressions however, are very complex even for humans to articulate with natural language. Among many, one possible reason is the used language for describing them – in particular the existing different modalities, which are even inconsistent with each other. This paper aims for a unified emotion model that is consistent and mappable to the existing ones, which makes our model also interpretable. The unified model is then used to enable generating rich facial expressions using text-to-image models, as shown in Figure~\ref{fig:teaser}. Further, we propose a method capable of understanding continuous expressions.

The most commonly used emotion models include basic Categorical~\cite{basicEmotion}, Compound~\cite{du2014compound}, and arousal-valence (AV)~\cite{PADrusselmehrabis} -based. The categorical model is simple, intuitive, and easy to annotate, while the compound emotion model is more complete. On the other hand, AV-based models are continuous where Categorical models can also be mapped, as shown in Figure~\ref{fig:introFig}~(middle).  A popular physics-based modeling of expressions is Action Units (AUs)~\cite{FACS}, which relies on the activation of the facial muscles. In fact, Compound emotions can be mapped to AUs, as shown in Figure~\ref{fig:introFig} (left). We aim to map all Categorical, Compound, AV, and AUs in a common unified representation, as shown in Figure~\ref{fig:introFig} (right), which we refer to as C2A2 (for Canonical, Compound, Action units, Arousal-valence).  To the best of our knowledge, such unification is proposed for the first time in this paper. Our proposed unification offers a better representation, leading to more versatile emotion generation, in the context of this paper. 

One major challenge of using a new emotion model is the missing associated labels. To address this problem, we first propose a 3D model such that it can exploit the existing 2D AV labels. Then, we propose a method to learn the additional third dimension without requiring any explicit supervision. In the language of basic categorical emotions, we lift up the ``fear" towards the positive third-dimension, and the ``sad" towards the negative side, as shown in Figure~\ref{fig:introFig} (right). The choice of these two particular emotions is made to best cover the compound emotions presented in the Figure~\ref{fig:introFig} (left). To learn the third dimension, our method leverages AUs-based modeling, where 3D vectors representing some compound emotions are first mapped to the action units followed by their supervision within a learning framework, inspired by GANmut~\cite{ganmut}. This allows us to first generate the third-dimension labels (Z), which we later use to learn the conditional image generation.   

On the image generation side, 
large text-to-image diffusion models \cite{ldm,glide,saharia2022photorealistic,yu2022scaling, balaji2022ediffi} have emerged as a powerful way of generating high-quality images. However, the existing models cannot understand the continuous number required to represent the facial expressions of our interest. Therefore, we also develop a method that facilitates generating images conditioned upon text and a vector of continuous numbers that represents the target emotions of interest. More specifically, we propose to use a number encoder that maps the emotion-condition vector into the common text embedding space. The embedded numbers are then used together with the text embeddings to generate text and emotion-conditioned images. 

We use the latent diffusion model~\cite{ldm} in the training setting of DreamBooth~\cite{dreambooth}. In this setting, we perform two parallel loss computations, one with and other without the emotion embedding. This regularizes the training and helps to preserve the knowledge of base diffusion model, thus allowing us to generate images with a rich expression and additional attributes described by the conditioning text input. Our experiments clearly demonstrate the superiority of the proposed 3D emotion model over the existing 2D AV, in the very same setup. Furthermore, our model that can understand both text and numeral inputs provides very very convincing expression generation results.

Our major contributions can be summarized as:
\begin{itemize}
    \item We propose an emotion model that unifies four different existing models in a common interpretable framework.
    \item We propose a method to annotate the emotion in the proposed emotion space by leveraging the AV and AUs.
    \item A number+text-to-image diffusion model is proposed to accommodate the proposed 3D numerical representation.
    \item Our results validate the 3D emotion model, annotation method, and number+text-to-image generation, by offering better quality and fine-grained control of expressions.
    
\end{itemize}

\section{Related Work}
\label{sec:related_work}
Modeling human emotions is a century-long ongoing topic of study~\cite{cross_culture,Circumplex,FacsProposal,PADrusselmehrabis,Kraut1979SocialAE,EmotionExpression,tang2023emotional}.
A commonly used model is basic categorical emotions~\cite{basicEmotion,larsen1992promises,rosenwein2010problems}. Other models are also used~\cite{scherer2000psychological,moods}. Among these, the most commonly used are compound emotions~\cite{du2014compound}, arousal-valence (AV)~\cite{PADrusselmehrabis}, and Action Units (AUs)~\cite{FACS}. The compound emotion representation also embeds the AUs, up to an extent, as shown in Figure~\ref{fig:introFig}. Similarly, AUs representation also embeds basic categorical and some compound emotions. However, to the best of our knowledge, there is no emotion model that offers better unification than the mentioned above.

Understanding and generating expressions are of high interest in computer vision~\cite{calder2016understanding,acharya2018covariance,kim2022emotion,stargan,ganimation,ganmut,ned}. 
Many generative works are utilized for the purpose of realistic manipulation of human emotions, including StarGAN~\cite{stargan}, GANimation~\cite{ganimation}, SMIT~\cite{ganimation}, GANmut~\cite{ganmut}, ICface~\cite{icface} and Neural Emotion Director~\cite{ned}.
These methods use generative adversarial networks (GANs)~\cite{goodfellow2020generative} to generate or manipulate the expressions expressed in the existing emotion models. Differently, a recent work~\cite{zou20234d} uses a diffusion model to generate landmark controlled 3D meshes for facial expressions. In this work, we introduce a new emotion model and propose a text-to-image diffusion method to generate images with expressions representing our targeted emotions. Nevertheless, we use the framework of GANmut to annotate images in our emotion representation space.

Diffusion-based methods~\cite{ddpm, gu2022vector, song2020score,song2020denoising, sohl2015deep} have become the go-to choice for image generation due to their synthesis quality and stable training. Recently, text-to-image  diffusion methods \cite{nichol2021glide, ramesh2021zero, ramesh2022hierarchical, rombach2022high, saharia2022photorealistic} have shown promise in enabling an intuitive interface for users to control image generation, using natural language descriptions. However, fine-grained control and customized image generation has proven difficult with natural language descriptions alone \cite{feng2022training, liu2022compositional, Wu_2023_CVPR,lee2023aligning}. To address this problem, some existing works adapt pre-trained models to their targeted examples, either to find pseudo-words\cite{t2i, motamed2023lego} or fine-tune some parts of the pre-trained model~\cite{dreambooth, Kumari_2023}. The pseudo-words are searched in the text embedding space of the text encoder (e.g, CLIP~\cite{clip}). On the other hand, the fine-tuning methods such as DreamBooth~\cite{dreambooth} fine-tunes only the attention layers while preserving the generation capabilities of the original network. Therefore, we are interested in this setting and thus develop our method to augment DreamBooth~\cite{dreambooth} with fine-grain control over facial expressions.

\section{The C2A2 Emotion Model}

\begin{table}[]
\scriptsize
    \centering
\begin{tikzpicture}
\node at (0,0) {
\begin{tabular}{
|p{1.6cm}|p{0.4cm}|p{0.4cm}||p{2cm}|p{0.4cm}|p{0.4cm}| }
 \hline
\textbf{Category}&\textbf{2D}&\textbf{3D} &\textbf{Category} &\textbf{2D}&\textbf{3D}\\
 \hline
 \hline
Sadly disgd.&\cmark&\cmark&Happily disgd.&\xmark&\cmark\\
\hline
Fearfully angry&\cmark&\cmark&Sadly feraful&\xmark&\cmark\\
\hline
Fearfully surpd.&\cmark&\cmark&Sadly angry&\xmark&\cmark\\
\hline
Angrily disgd.&\cmark&\cmark&Fearfully disgd.&\xmark&\cmark\\
\hline
Happily surpd.&\cmark&\cmark&Angrily surpd.&\xmark&\cmark\\
\hline
Happily sad&\cmark&\cmark&Happily fearful&\xmark&\cmark\\
\hline
-&-&-&Sadly surpd.&\xmark&\cmark\\
\hline
\hline
Awed&-&-&Happy+surprise+fear&\xmark&\cmark\\
\hline
Hatred&-&-&Disgust+anger+fear&\xmark&\cmark\\
\hline
\hline
Appalled&-&-&Disgust+surprise&\xmark&\xmark\\
\hline
Disgd. surpd.&\xmark&\xmark&-&-&-\\
\hline
\end{tabular}
};
\draw[->, thick] (-1.8,-0.8) -- (0,-0.8);
\draw[->, thick] (-1.8,-1.1) -- (0,-1.1);
\draw[->, thick] (-1.8,-1.48) -- (0,-1.48);
\end{tikzpicture}
    \caption{The compound emotions that can and cannot be represented by the proposed 3D representation of emotions. Our 3D model can represent 15/17 desired emotions (after mapping ``Awed" and ``Hatred" to composition of three basic emotions), whereas, 2D representation of AV can represent only 6/17.}
    \label{tab:2d-3D_emotions}
    \vspace{-3mm}
\end{table}

While aiming to generate compound and continuous expressions, we realized that the existing interpretable representations do not support our needs. Therefore, we proceed to modify the most suitable existing model, namely arousal-valence, as it already embeds the basic emotions in the 2D continuous space. In fact, this representation allows six compound emotions to be expressed, as shown in Table~\ref{tab:2d-3D_emotions}. In the same Table, it can be seen that other compound emotions of interest are not expressed using the 2-dimensional AV model. Therefore, we propose to represent the emotions in 3-dimensional space, while preserving the structure of the AV-based 2D model. More specifically, we lift the ``fear" towards the positive third-dimension, and the ``sad" towards the negative side, as shown in Figure~\ref{fig:introFig} (right). This choice is made to best cover the most number of compound emotions, i.e. 15/17, as shown in the Table~\ref{tab:2d-3D_emotions}. In fact, we decompose two terminologies of~\cite{FACS}, ``Awed", ``Hatred", and ``Appalled", into the composition of basic emotions, happy+surprise+fear, disgust+anger+fear, and disgust+surprise, respectively. This leads to the compatibility of two additional compound emotions. Unfortunately, the emotion ``Appalled" and ``Disgustedly surprised" are not yet compatible with our emotion model. We choose to avoid modelling these two emotions to simplify our emotion model and make use of the AV labels. In fact, in our 3D representation, we use AV labels as the 2D coordinate and learn the third dimension, which we denote with a variable $Z$, using a method inspired by GANmut~\cite{ganmut}, with the help of the Action Units' labels, presented below.  

\subsection{Implicit Supervision for $Z$ of C2A2}
\label{sec:method-model}
To learn the third dimension of  C2A2, we extend the idea proposed in 2D linear GANmut model that has a conditional space parameterized with polar coordinates $\theta$ and $\rho$, which are both uniformly distributed  $\theta\sim U([0,2\pi])$, $\rho\sim U([0,1])$. The angle  $\theta$ indicates the category of the emotion, while the radius $\rho$ tells more about its intensity. The expected behavior of the model is that the intensity increases with the distance from the center and the emotion transition between two basic emotions is smooth and continuous, reflecting the compound emotions in between. 

 Our method aims to learn a linear 3D model by fixing the positions of basic categorical emotions. We use the AffectNet~\cite{affectnet} dataset. From the AV labels,  the angles of the basic emotions are determined by averaging the AV labels corresponding to them. We fix the angles of basic emotions happiness, surprise, disgust, and anger vectors following the estimations. The extreme point of each of these emotions is set at the maximum distance from the center. On the other hand, fear and sadness, we lift above and under the AV plane respectively. The lifting is constrained in such a way that their projection to the AV plane corresponds to the expected 2D position, and makes $60^\circ$ angle with the AV plane. Now, any emotion represented in our 3D space can be mapped back to AV by projecting on the XY-plane.

By lifting the two emotions to the third dimension, we have created a void in labels along the $Z$ dimension. We wish to learn the labels along this dimension in an implicit manner, so as to avoid the need of direct annotations. Note from Table~\ref{tab:2d-3D_emotions} that by making the modification we are able to represent nine additional compound emotions. More importantly, \emph{these compound emotions can be mapped to the AUs, which are also continuous in nature} (please, refer the left side of Figure~\ref{fig:introFig}). Now, our interest is to exploit the continuous labels of AUs, which can conversely help us in the 3D space C2A2. In this work, we first design the mapping process for AUs, and then exploit them to learn the 3D space of C2A2 by using the conditional space learning framework, from weak labels, of GANmut~\cite{ganmut}. 
\begin{figure*}[!htp]
\includegraphics[width=1.2\linewidth]{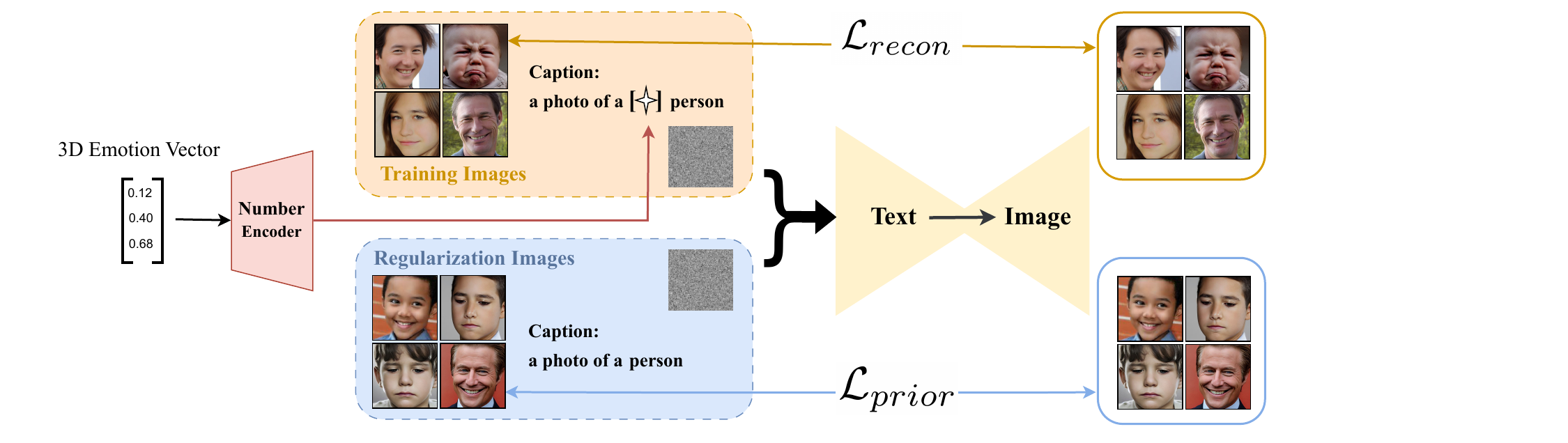}
\vspace{-0.7cm}
\caption{\label{fig:diffusion} We use a number encoder that embeds the continuous 3D representations of emotions. The embedded numbers are fused with the text embedding before decoding into number+text-to-image generation. The learning is done using the frozen text-encoder and shared image decoder. During learning, our method uses prior preservation and emotion reconstruction loss, similar to DreamBooth~\cite{dreambooth}. }
\vspace{-0.3cm}
\end{figure*}

Learning the conditional space is based
on conditioning the samples from mini batches that correspond to the basic categorical emotions, but this time we include also the compound ones. 
While learning $Z$, we supervise our model also by AV labels, say $v_{va}$. Therefore, we add the following new loss to the discriminator's objective function,
\begin{equation} \label{eq15}
\mathcal {L}_{av}=\mathbb {E}_{x,z}[||D_{coor}(x)-v_{av}||_2^2],
\end{equation}
where, $D(.)$ is the discriminator network, and $x$ is the input image. The second loss added to the GANmut discriminator’s objective function is the AU loss. Since AffectNet does not provide action units’ labels, they were manually mapped from the valence and arousal values. Therefore, labels of the real images are limited to 12 possible sets of AUs, which could be found in the AV plane (please, refer to the Figure~\ref{fig:introFig} (left)). The mapping starts by dividing the space between each basic and surrounding compound emotions into two parts, such that one half could be still considered the basic emotion, while the other half goes into the part covered by the compound one.\\

\begin{tikzpicture}
\node at (0,0) {$Y$};
\draw[->, thick] (0.3,0) -- (1,0);
\node at (2.3,0) {Emotion category};
\draw[->, thick] (3.7,0) -- (4.5,0);
\node at (5.0,0) {$AU$}; 
\draw[<->, thick] (5.5,0) -- (6.4,0);
\node at (7,0) {$\hat{AU}_{real}$}; 
\node at (6.0,0.3) {\scriptsize $\mathcal{L}_{AU_Y}$};
\end{tikzpicture}

We obtain the pseudo-labels $\hat{AU}_{real}$ using the OpenGraphAU tool~\cite{AUgraph}. The OpenGraphAU provide us the activation probability of all 41 actions units. Based on  Figure~\ref{fig:introFig} (left), we need only 15 action units in total to decide on the compound emotions of our interests. The pseudo-labels of AUs are used to compute an additional loss between the labeled activation probabilities and the estimated ones, 
\begin{equation} 
\label{eq16}
\mathcal L_{AU_{Y}}=\mathbb {E}_{x,Y}[KL(D_{au}||AU_{Y}) + KL(AU_{Y}|| D_{au})].
\end{equation}
Here, $KL(.||.)$ is the Kullback-Leibler Divergence, and $Y = [A,V,Z]^\intercal$ is the 3D conditional vector. During training, the generator is conditioned on two mentioned mini batches and sampling was performed along the basic and compound emotion vectors (or in their proximity).

\begin{figure*}
\begin{tikzpicture}
    \node at (0,0)
 {
 \addtolength{\tabcolsep}{-4.5pt}
\begin{tabular} {cccccccccccc}
 & $\theta=0.11\pi$ & $0.32\pi$ & $0.53\pi$ & $0.74\pi$ & $0.95\pi$ & $1.16\pi$ & $1.37\pi$ & $1.58\pi$ & $1.79\pi$ & $2\pi$ \\

 \rotatebox{90}{\,\,\,\, Z=0.0} &
 \includegraphics[width=0.09\linewidth]{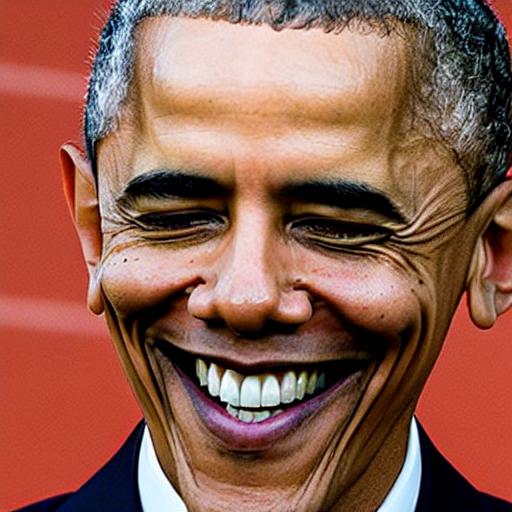}&
 \includegraphics[width=0.09\linewidth]{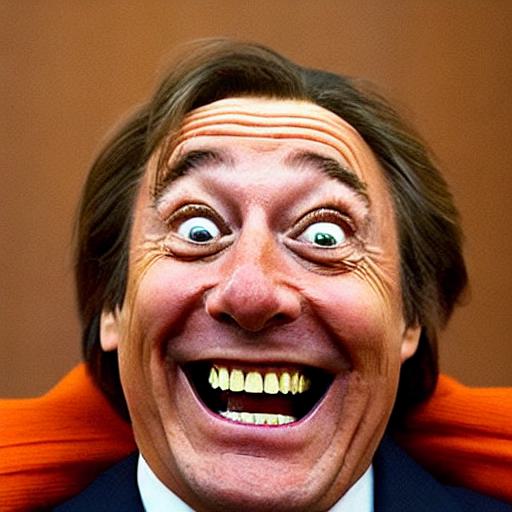}&
 \includegraphics[width=0.09\linewidth]{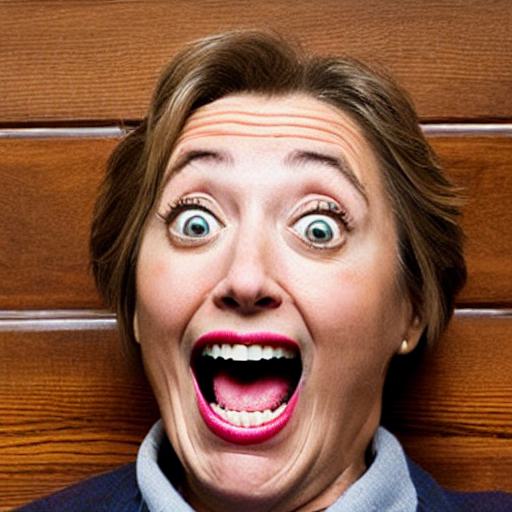}&
 \includegraphics[width=0.09\linewidth]{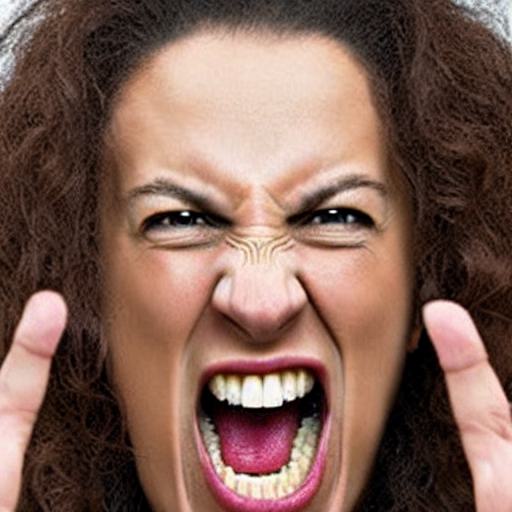}&
 \includegraphics[width=0.09\linewidth]{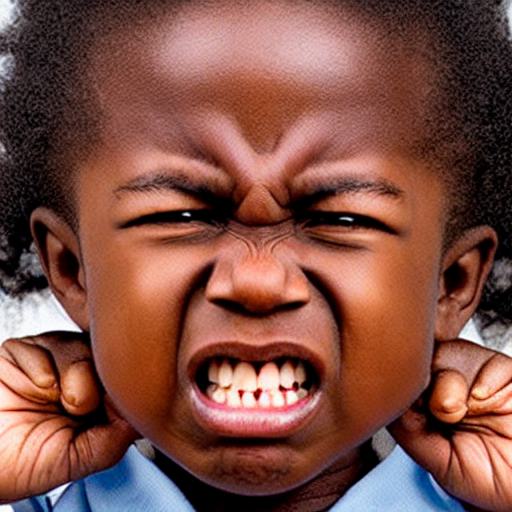}&
 \includegraphics[width=0.09\linewidth]{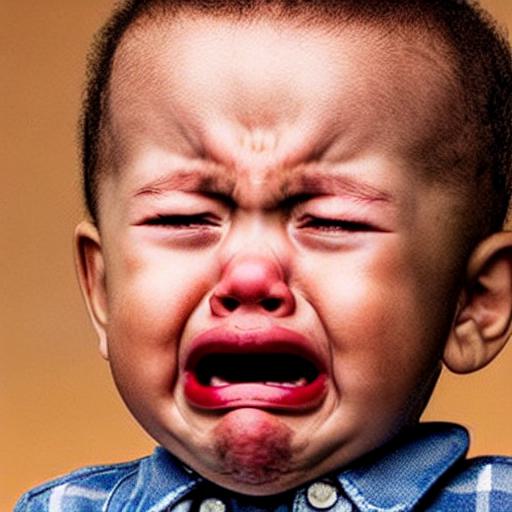}&
 \includegraphics[width=0.09\linewidth]{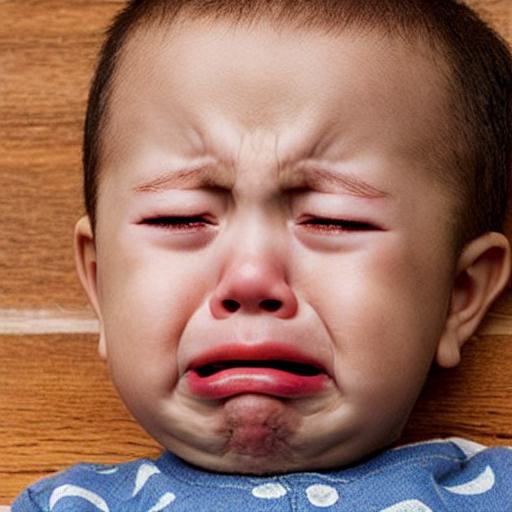}&
 \includegraphics[width=0.09\linewidth]{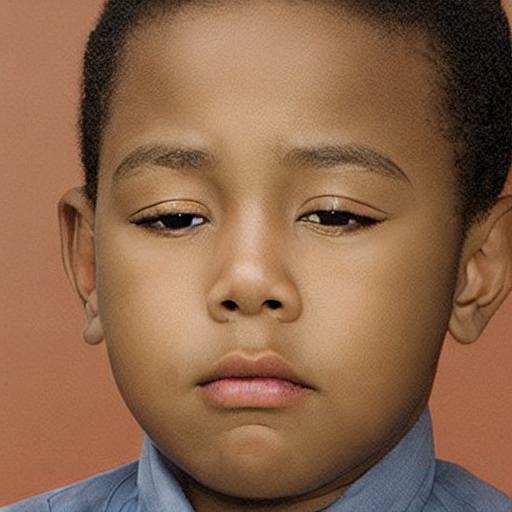}&
 \includegraphics[width=0.09\linewidth]{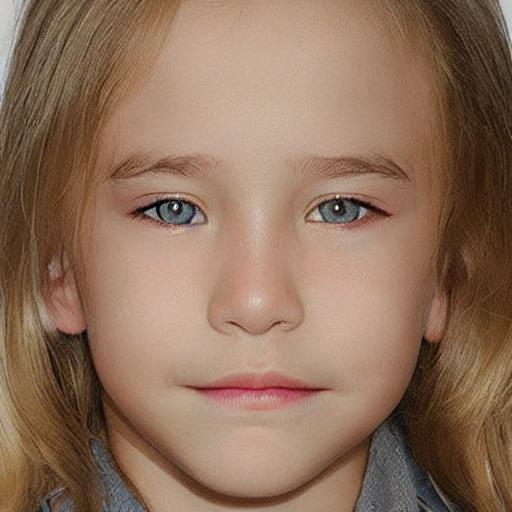}&
 \includegraphics[width=0.09\linewidth]{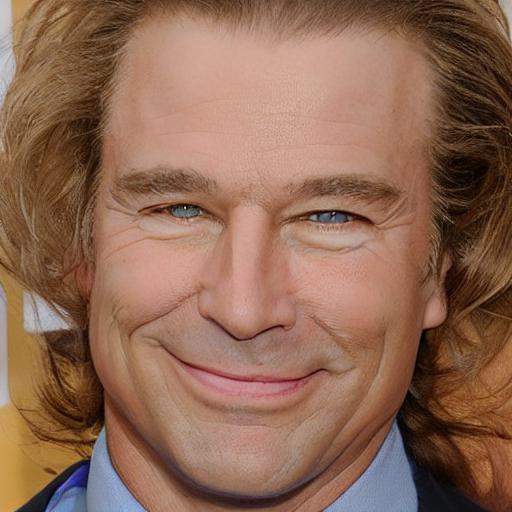}
 \\ 

 \rotatebox{90}{\,\,\,\, Z=0.5} &
 \includegraphics[width=0.09\linewidth]{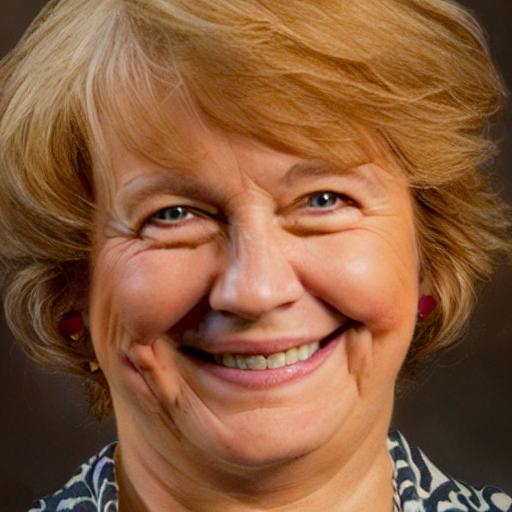}&
 \includegraphics[width=0.09\linewidth]{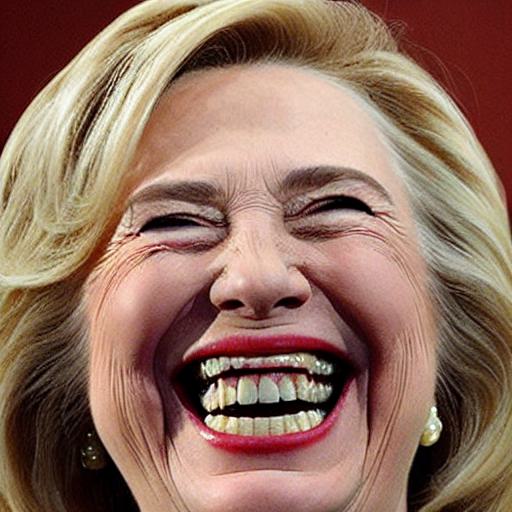}&
 \includegraphics[width=0.09\linewidth]{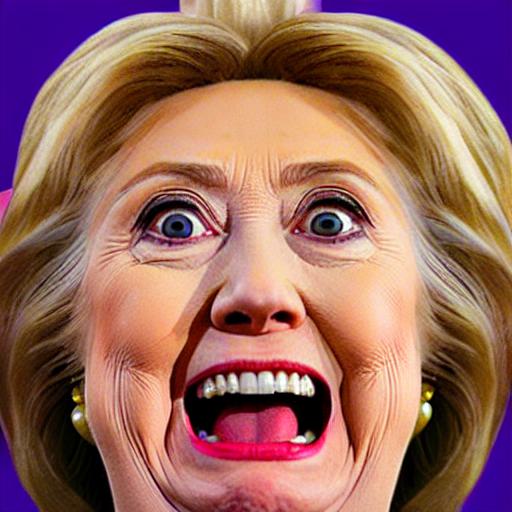}&
 \includegraphics[width=0.09\linewidth]{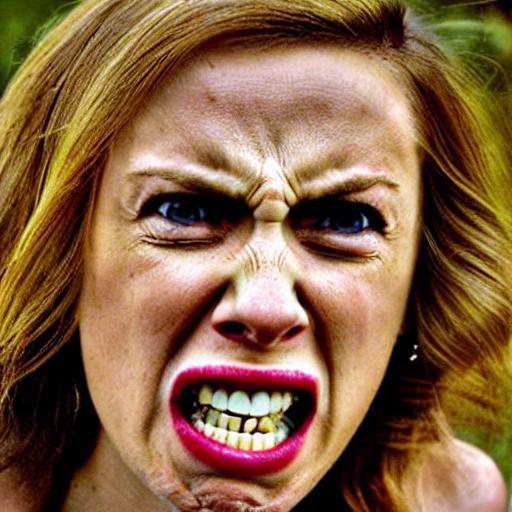}&
 \includegraphics[width=0.09\linewidth]{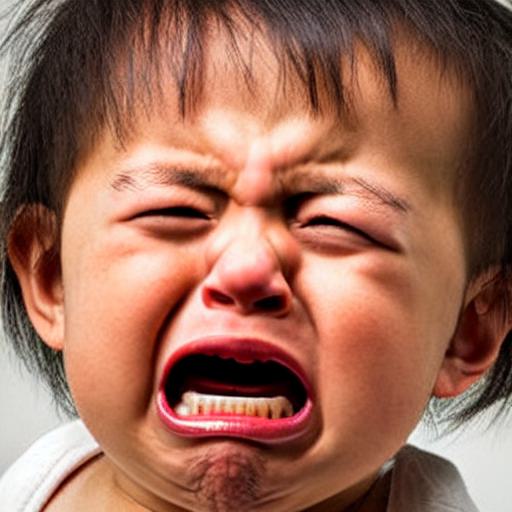}&
 \includegraphics[width=0.09\linewidth]{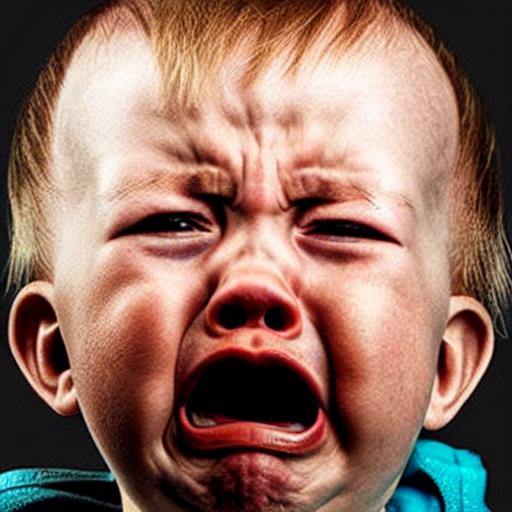}&
 \includegraphics[width=0.09\linewidth]{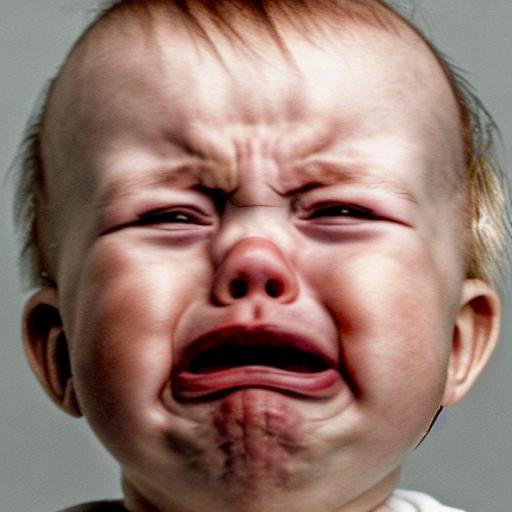}&
 \includegraphics[width=0.09\linewidth]{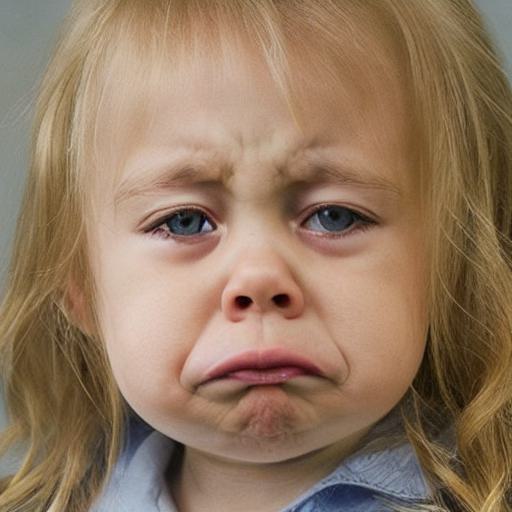}&
 \includegraphics[width=0.09\linewidth]{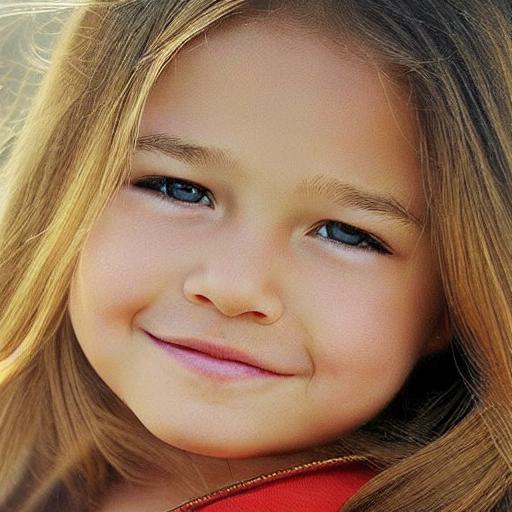}&
 \includegraphics[width=0.09\linewidth]{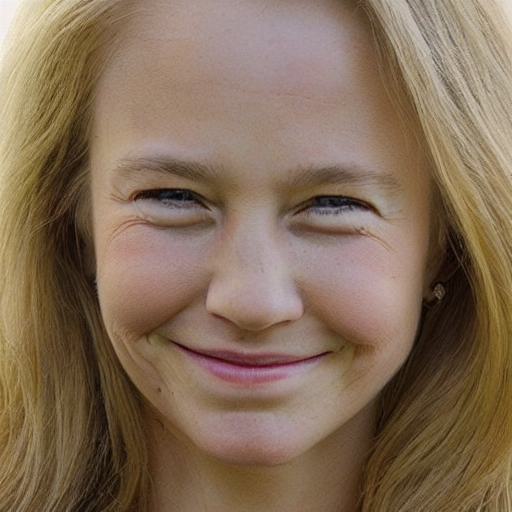}
 \\

  \rotatebox{90}{\,\,\,\, Z=-0.8} &
 \includegraphics[width=0.09\linewidth]{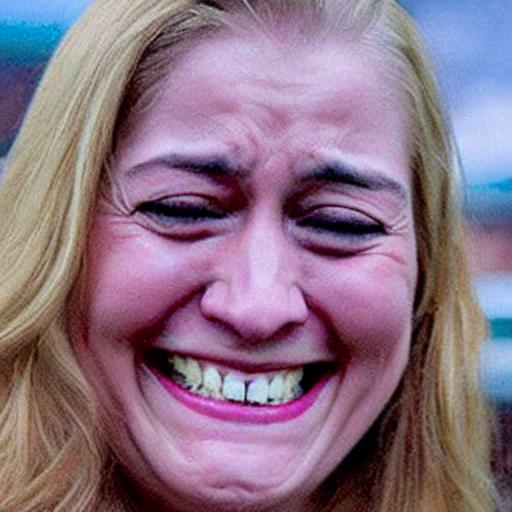}&
 \includegraphics[width=0.09\linewidth]{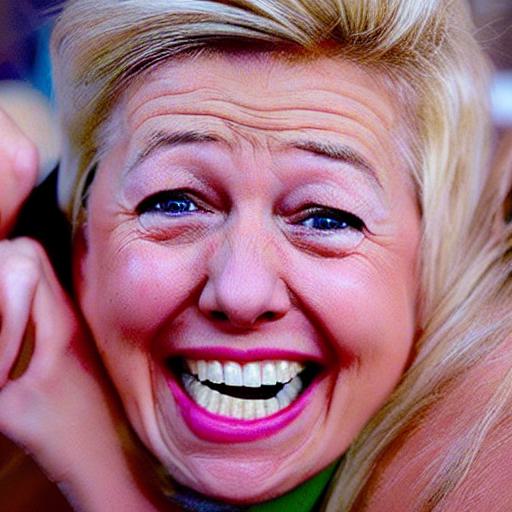}&
 \includegraphics[width=0.09\linewidth]{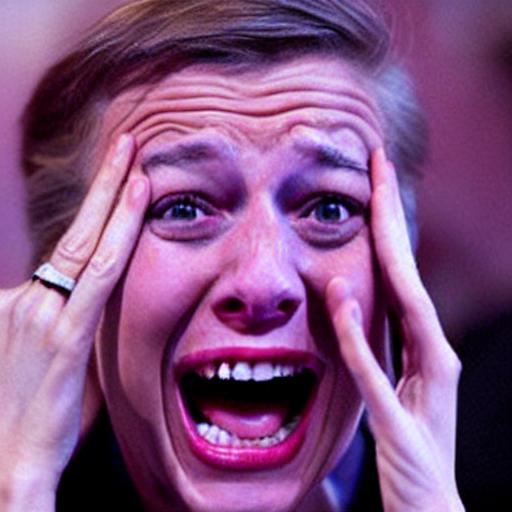}&
 \includegraphics[width=0.09\linewidth]{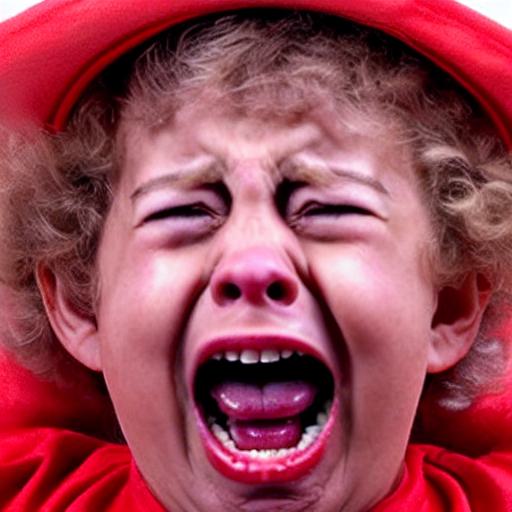}&
 \includegraphics[width=0.09\linewidth]{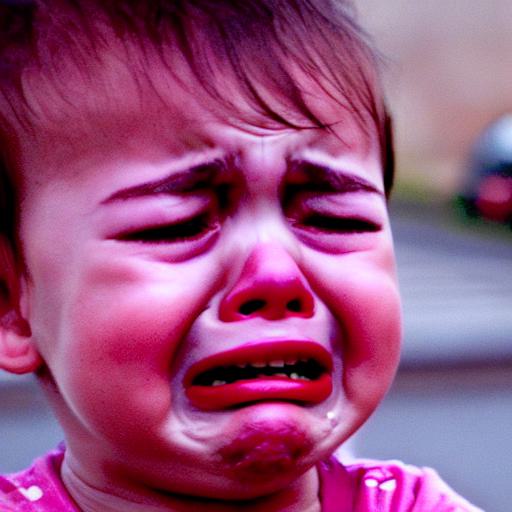}&
 \includegraphics[width=0.09\linewidth]{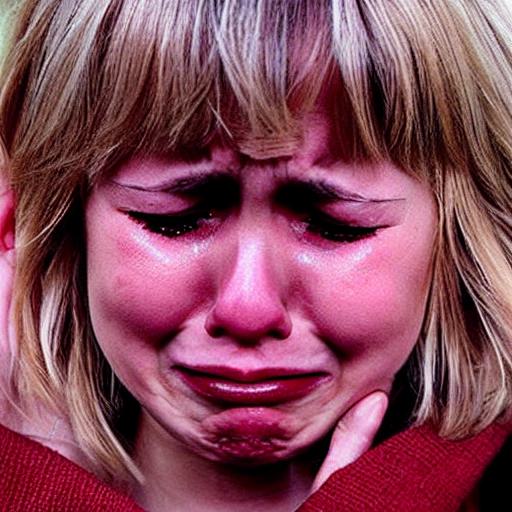}&
 \includegraphics[width=0.09\linewidth]{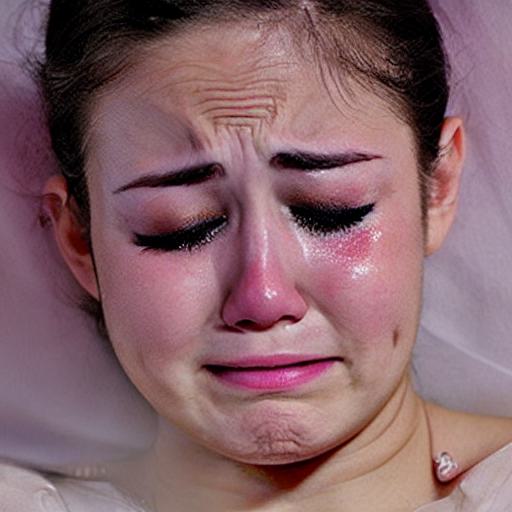}&
 \includegraphics[width=0.09\linewidth]{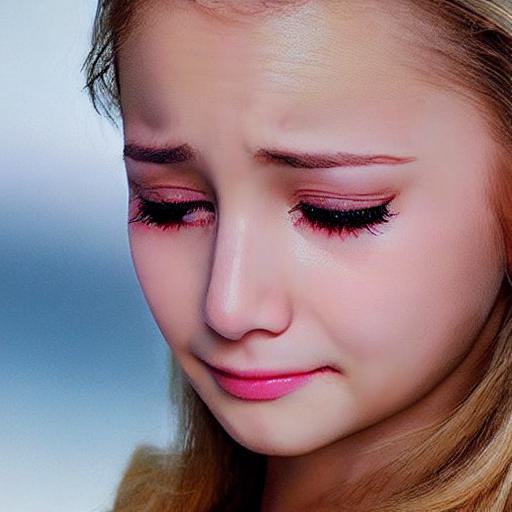}&
 \includegraphics[width=0.09\linewidth]{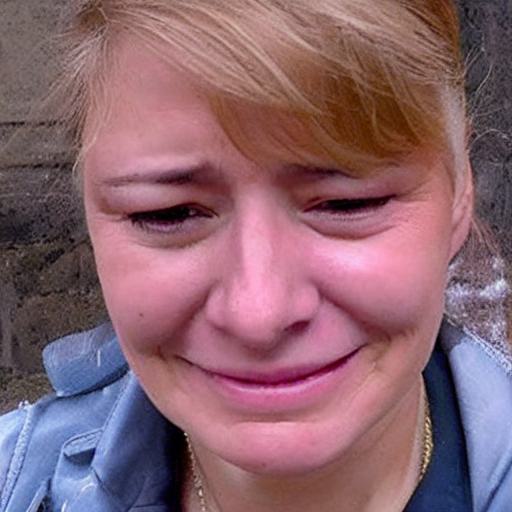}&
 \includegraphics[width=0.09\linewidth]{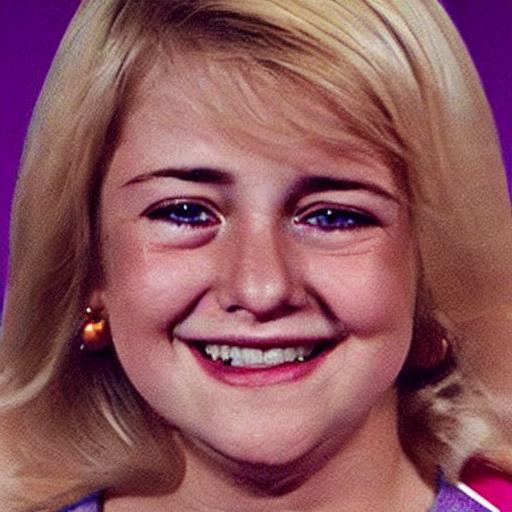}
 \\
\hline
  \rotatebox{90}{2D model} &
 \includegraphics[width=0.09\linewidth]{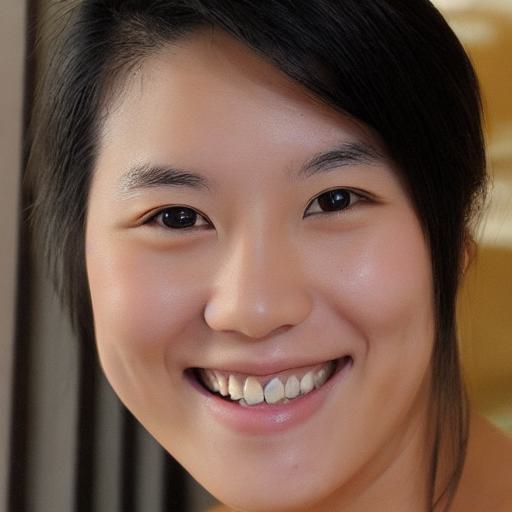}&
 \includegraphics[width=0.09\linewidth]{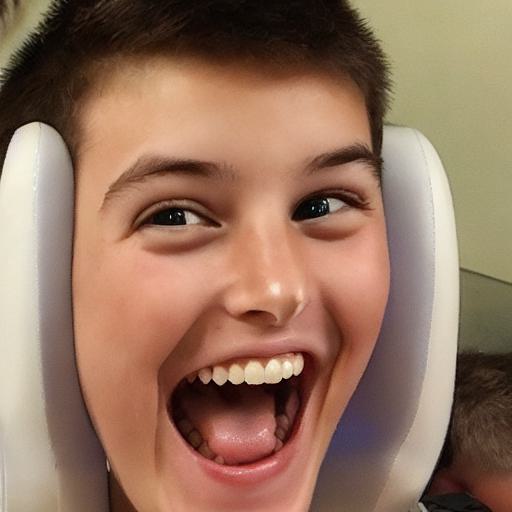}&
 \includegraphics[width=0.09\linewidth]{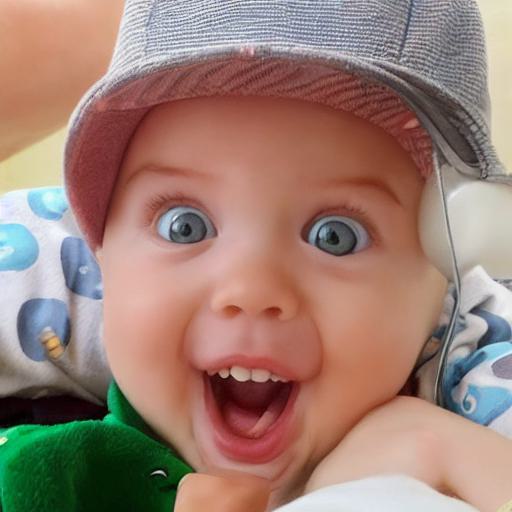}&
 \includegraphics[width=0.09\linewidth]{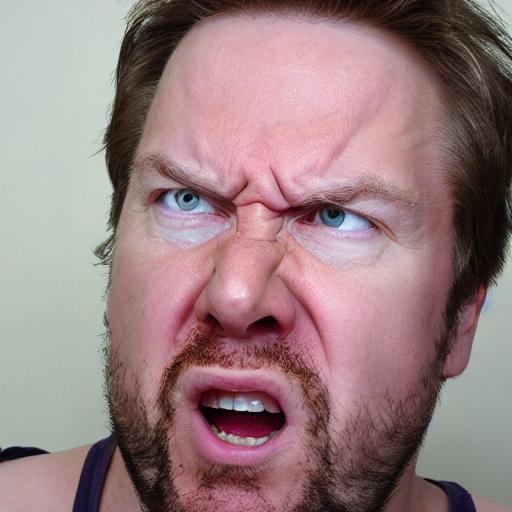}&
 \includegraphics[width=0.09\linewidth]{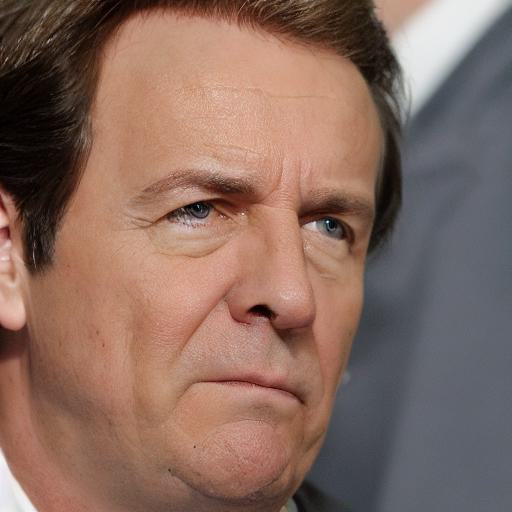}&
 \includegraphics[width=0.09\linewidth]{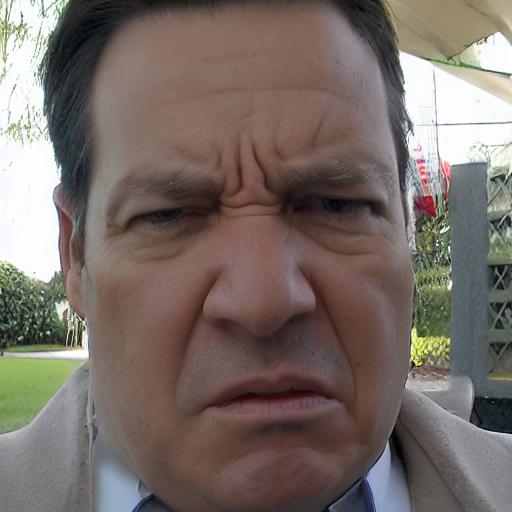}&
 \includegraphics[width=0.09\linewidth]{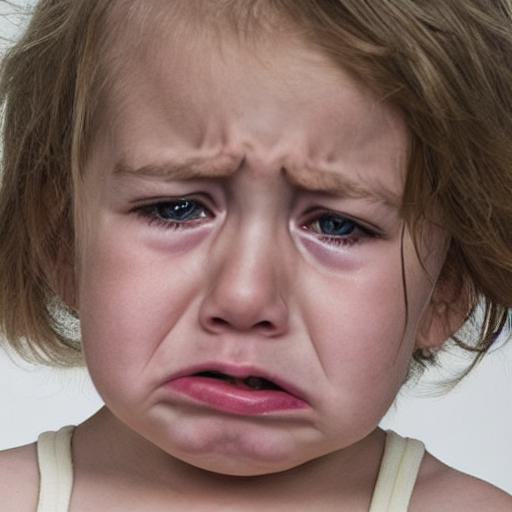}&
 \includegraphics[width=0.09\linewidth]{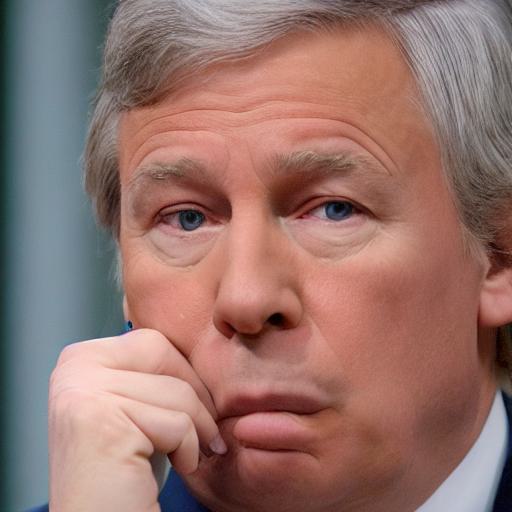}&
 \includegraphics[width=0.09\linewidth]{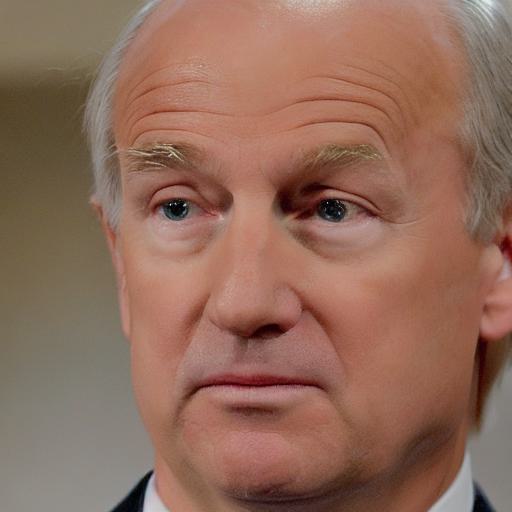}&
 \includegraphics[width=0.09\linewidth]{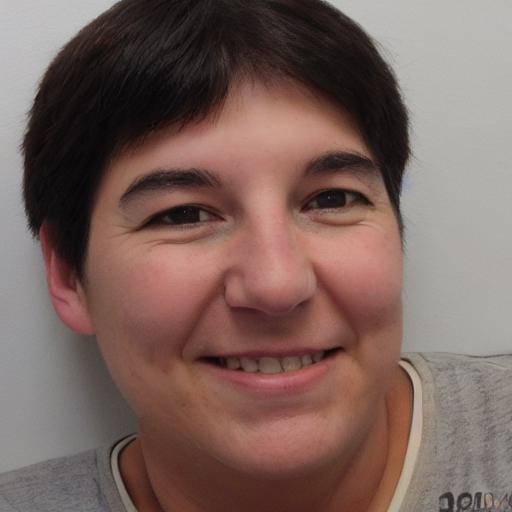}
\end{tabular}
    };
\end{tikzpicture}
\vspace{-0.8cm}
\caption{Top three rows: images sampled around a circle (angle of AV on top) at different learned $Z$ of our 3D model. Bottom: the same circle for 2D model. Not that the 3D model is clearly superior than 2D. The images on  first and last rows may directly be compared.  \label{fig:circle} }
\vspace{-0.5cm}
\end{figure*}

\subsection{ Unprojecting Images along $Z$}
\label{sec:unproj}
Once the conditional space is learned using the GANmut framework, we obtain the labels for $Z$ in rather a straightforward manner. Although the conditional space is implicitly learned by the earlier training, the images still need to be mapped to the conditional space to obtain the $Z$ labels. This could possibly be done more accurately by using techniques reported~\cite{xia2022gan}. However, we use a simple approach and obtain the sought labels directly from the discriminatory network. Let $\hat{Z} =  D(x)$ be the z-dimensional label predicted by the discriminator for a given image $x$, then we use $Y=[A,V,\hat{Z}]$ as the emotion label corresponding to that image in the proposed representation. 

\section{ Hybrid Text-to-Image Generation }
Describing compound emotions using natural language descriptions does not always lead to faithful representation of the intended emotion (see Figure ~\ref{fig:teaser}) in text-to-image models. To introduce granular control over such emotions, while taking advantage of large text-to-image diffusion models, we use a number encoder that encodes the 3D emotion vector $Y\in\mathbb{R}^3$ such that, when concatenated with the encoded text description, it can depict the described face with the intended emotion represented by the 3D vector. 

During training, the prior loss $\mathcal{L}_{prior}$ is tasked with mitigating language drift and overfitting on training images  while the reconstruction loss $\mathcal{L}_{recon}$ enables the desired control of the generated faces' expressions. Figure~\ref{fig:diffusion} shows how C2A2 effectively augments text-to-image models with a number encoder. We train our method on AffectNet dataset with and without 3D labels $Y$, which are obtained using the method described in the previous section.   

\noindent {\textbf{Number Encoder.} We use an MLP-based number encoder $E_y = \phi_\theta(Y)$ that projects the emotion vector $Y\in\mathbb{R}^3$  to a higher  dimensional embedding. Note that the DreamBooth~\cite{dreambooth} framework uses CLIP~\cite{clip} as the text encoder, following this we map our emotion vector to the embedding $E_y\in\mathbb{R}^{768}$. This embedding is merged with the embeddings from the text encoder to condition the image decoder. 

\noindent {\textbf{Learning number+text-to-image generation.}
To enable the joint condition of the numbers and text, we first map the emotion vectors to a higher dimensional embedding. These embeddings are then used together with the text embeddings, followed by the text-to-image generator training using $\mathcal{L}_{prior}$ and $\mathcal{L}_{recon}$ losses, similarly as in~\cite{dreambooth}.
During the inference, text and number embeddings are fused to generate images with continuous emotions and given text description. Please, refer Figure~\ref{fig:attributes} for such examples.

\begin{figure*}
\begin{tikzpicture}
    \node at (0,0) 
 {
 \addtolength{\tabcolsep}{-4.5pt}
\begin{tabular} {cccccccccccc}
 & r=0.10 & 0.21 & 0.32 & 0.42 & 0.53 &0.63 & 0.74& 0.84& 0.95& 1.0\\

 \rotatebox{90}{3D-Fearful} &
 \includegraphics[width=0.09\linewidth]{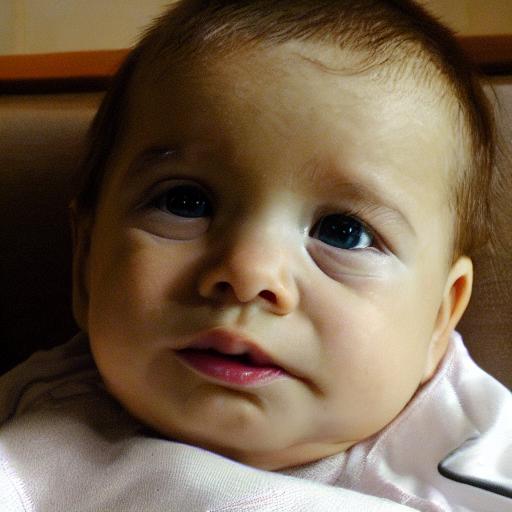}&
 \includegraphics[width=0.09\linewidth]{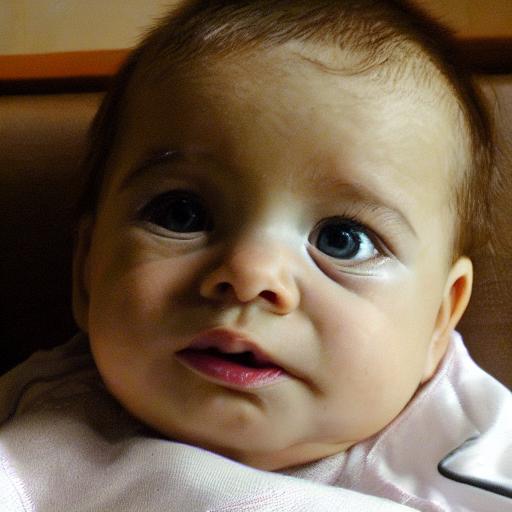}&
 \includegraphics[width=0.09\linewidth]{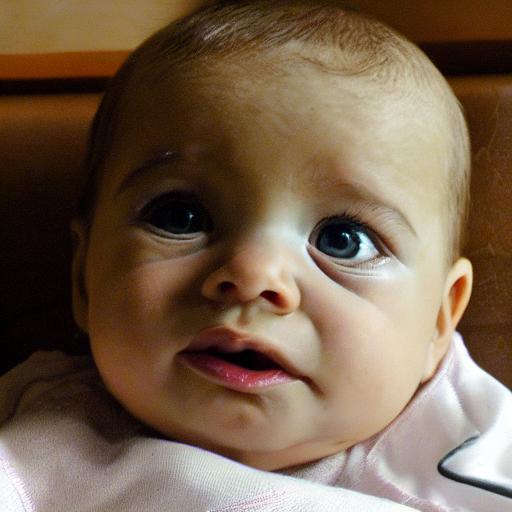}&
 \includegraphics[width=0.09\linewidth]{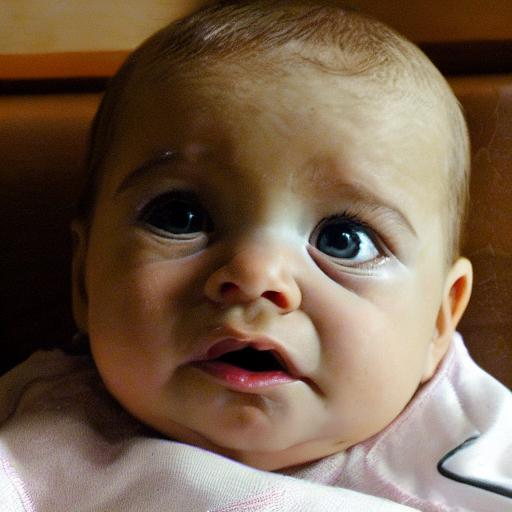}&
 \includegraphics[width=0.09\linewidth]{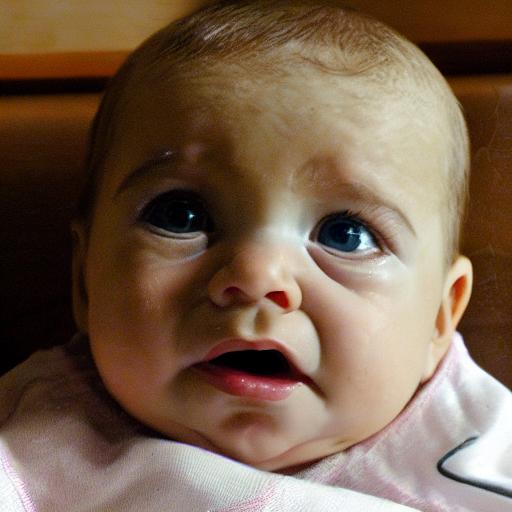}&
 \includegraphics[width=0.09\linewidth]{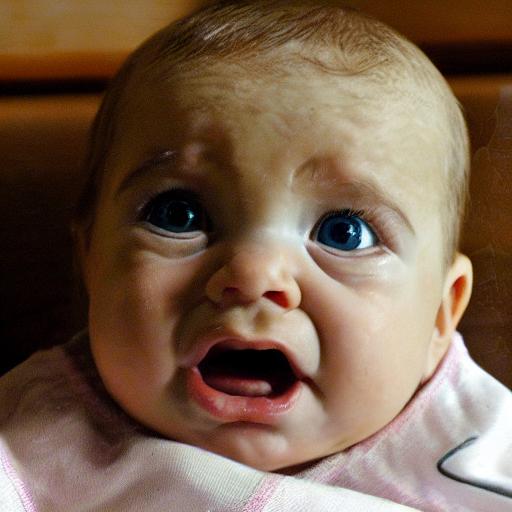}&
 \includegraphics[width=0.09\linewidth]{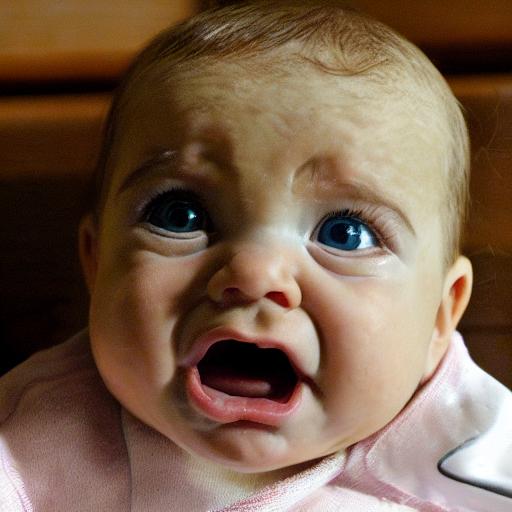}&
 \includegraphics[width=0.09\linewidth]{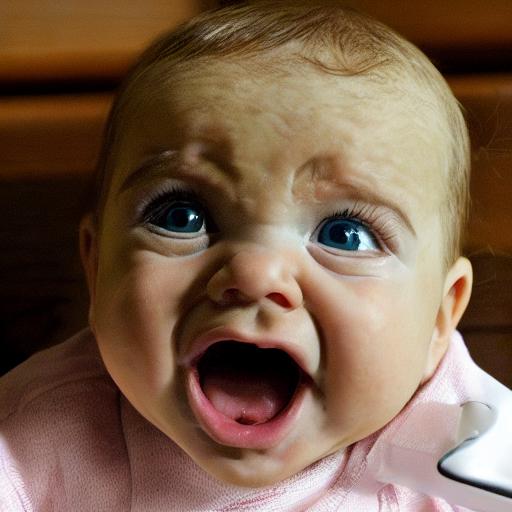}&
 \includegraphics[width=0.09\linewidth]{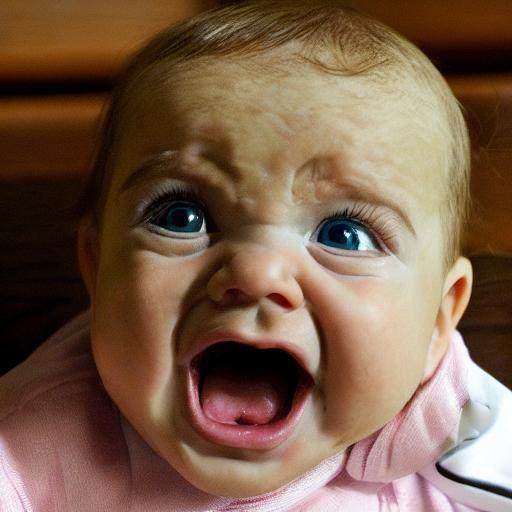}&
 \includegraphics[width=0.09\linewidth]{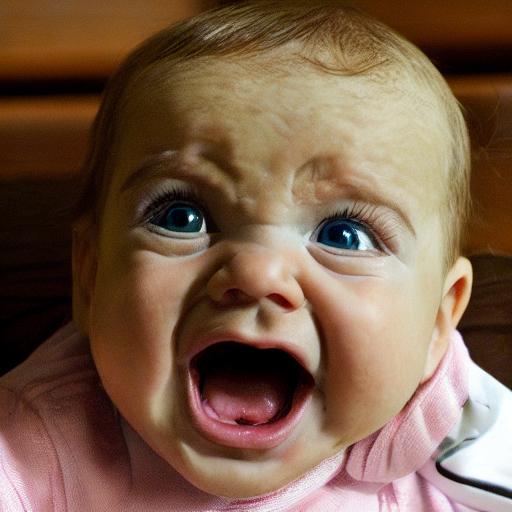}
 \\ 

 \rotatebox{90}{2D-Fearful} &
 \includegraphics[width=0.09\linewidth]{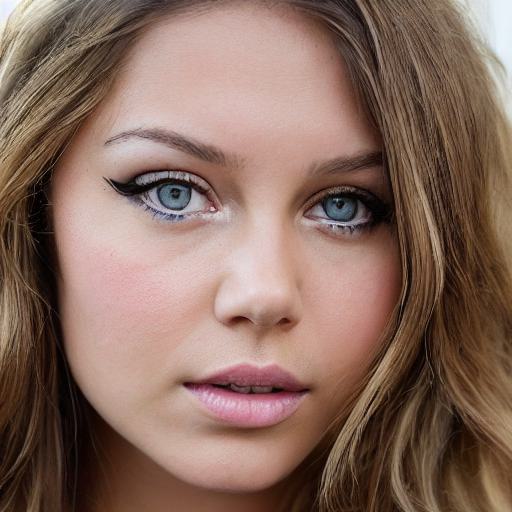}&
 \includegraphics[width=0.09\linewidth]{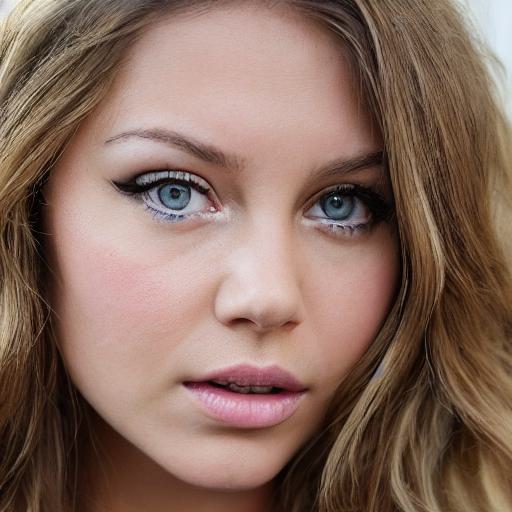}&
 \includegraphics[width=0.09\linewidth]{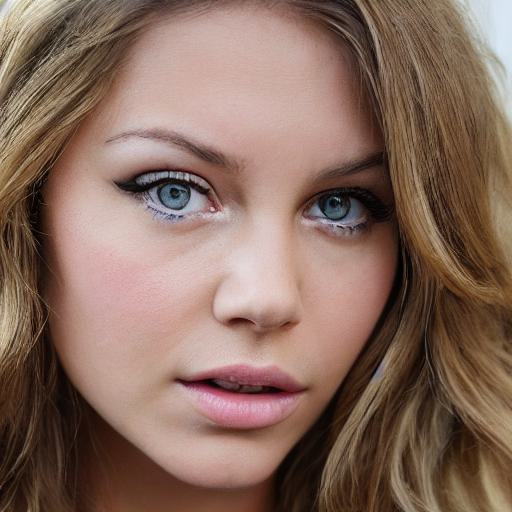}&
 \includegraphics[width=0.09\linewidth]{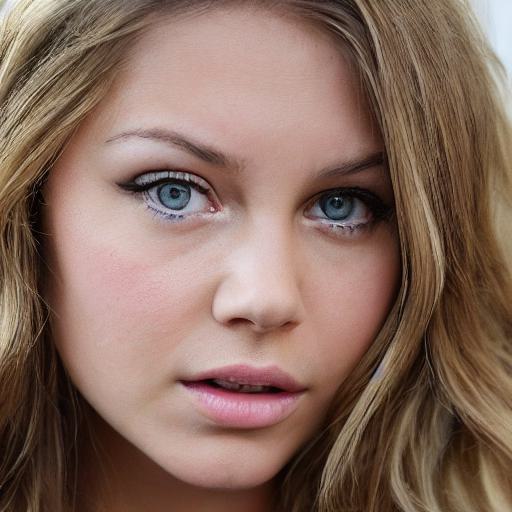}&
 \includegraphics[width=0.09\linewidth]{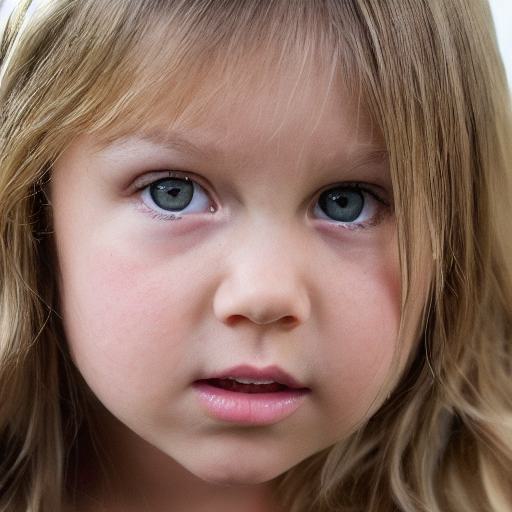}&
 \includegraphics[width=0.09\linewidth]{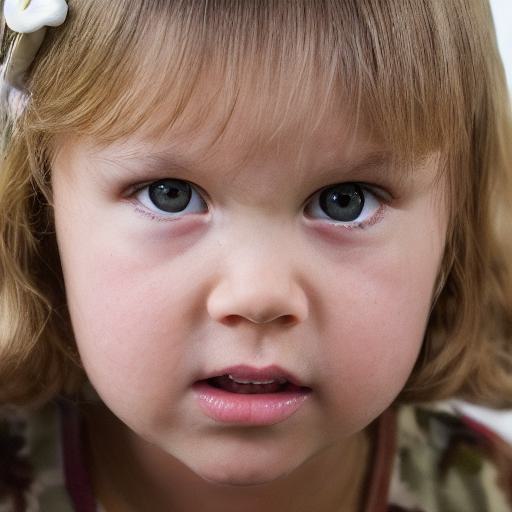}&
 \includegraphics[width=0.09\linewidth]{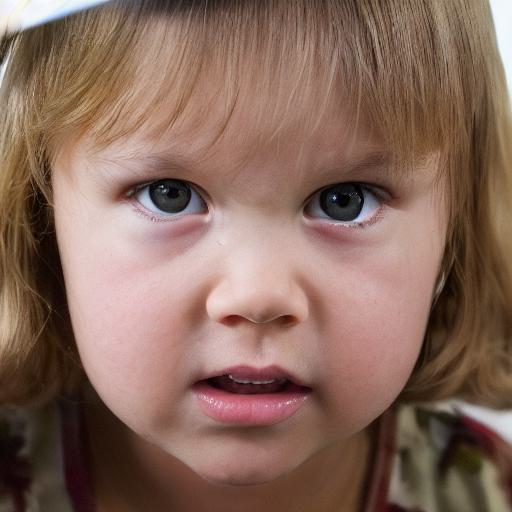}&
 \includegraphics[width=0.09\linewidth]{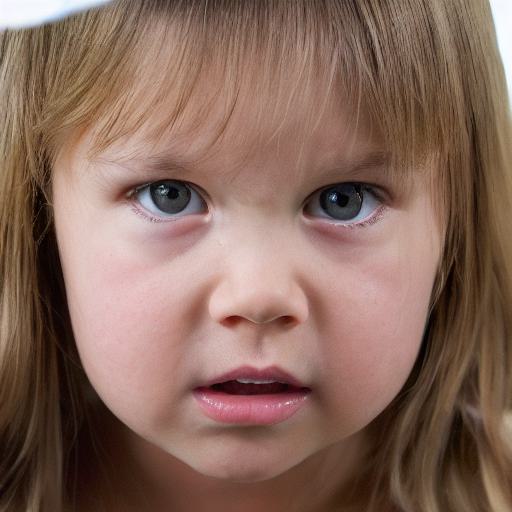}&
 \includegraphics[width=0.09\linewidth]{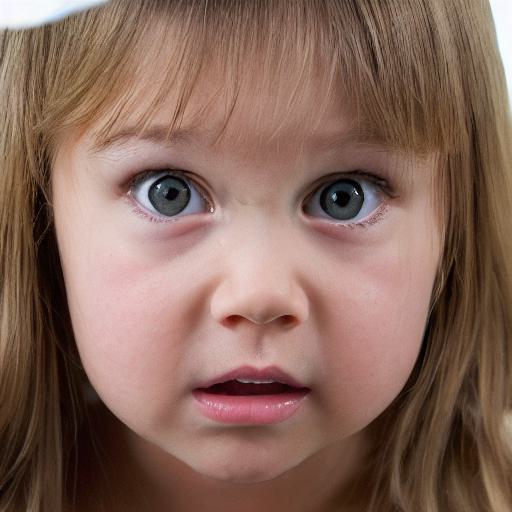}&
 \includegraphics[width=0.09\linewidth]{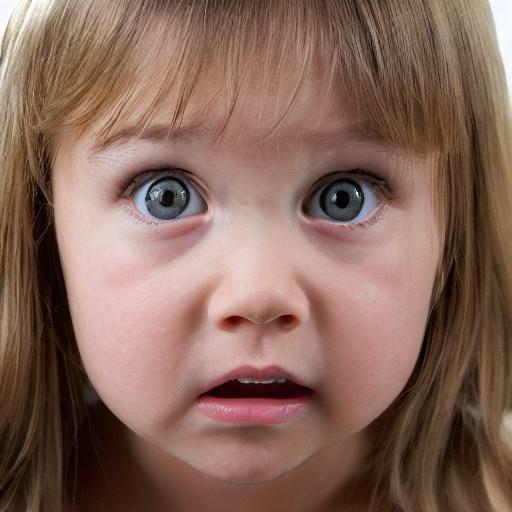}
 \\

 \rotatebox{90}{3D-Sad} &
 \includegraphics[width=0.09\linewidth]{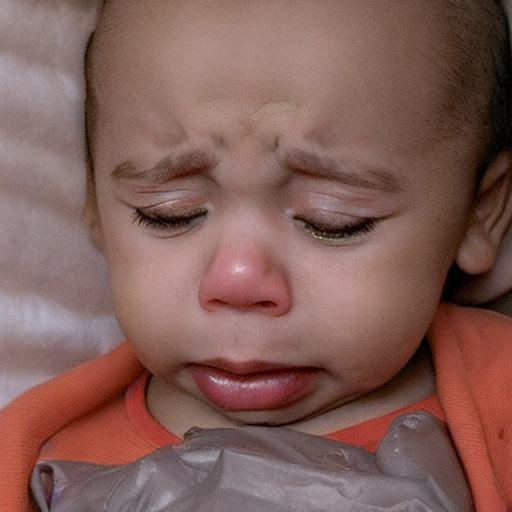}&
 \includegraphics[width=0.09\linewidth]{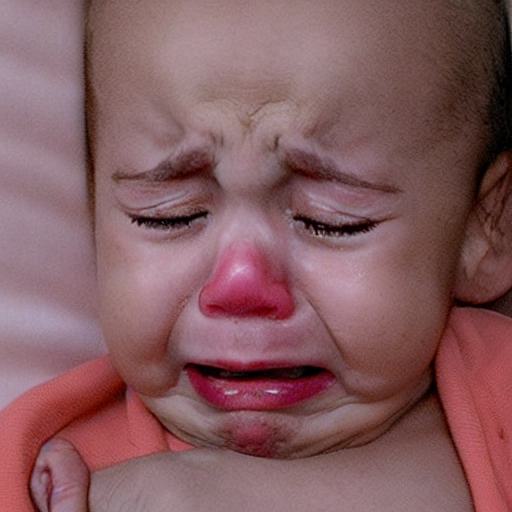}&
 \includegraphics[width=0.09\linewidth]{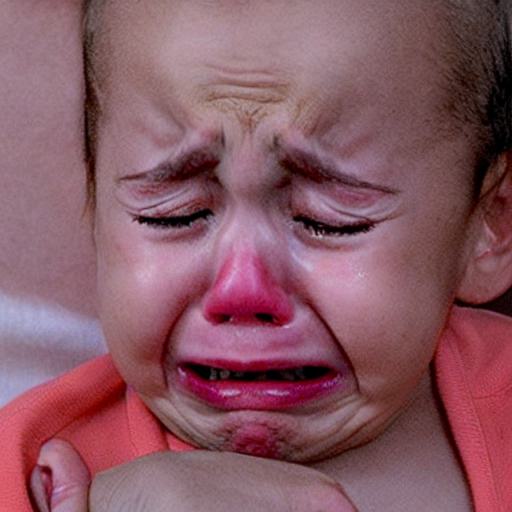}&
 \includegraphics[width=0.09\linewidth]{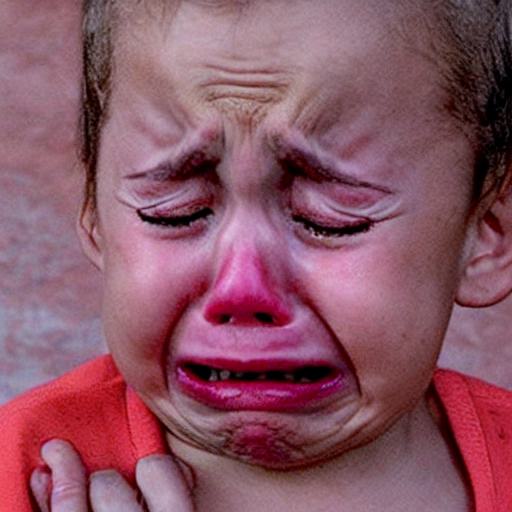}&
 \includegraphics[width=0.09\linewidth]{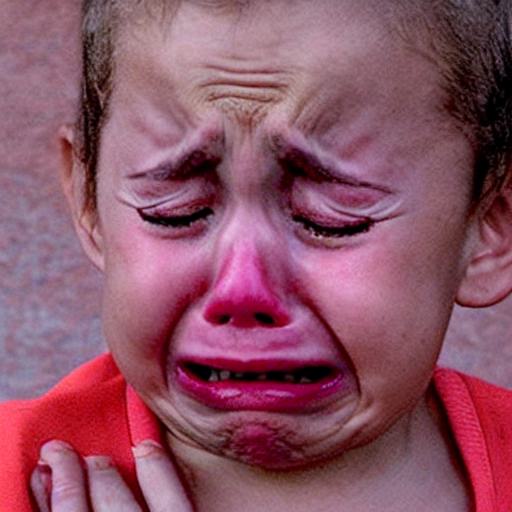}&
 \includegraphics[width=0.09\linewidth]{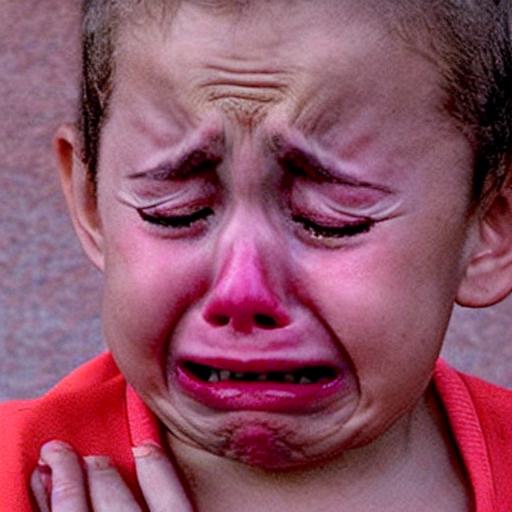}&
 \includegraphics[width=0.09\linewidth]{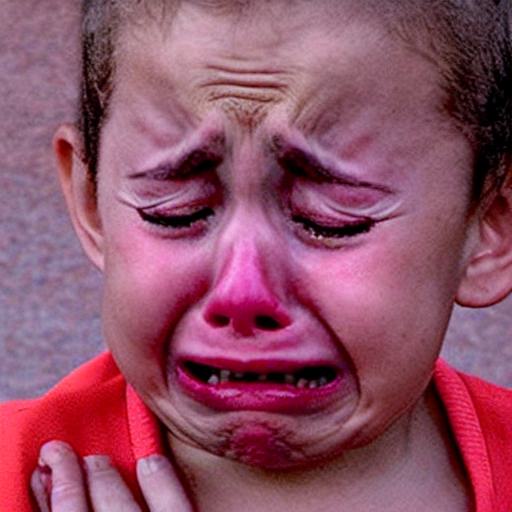}&
 \includegraphics[width=0.09\linewidth]{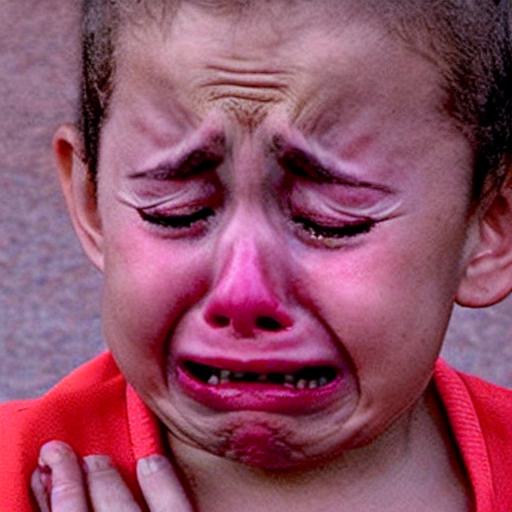}&
 \includegraphics[width=0.09\linewidth]{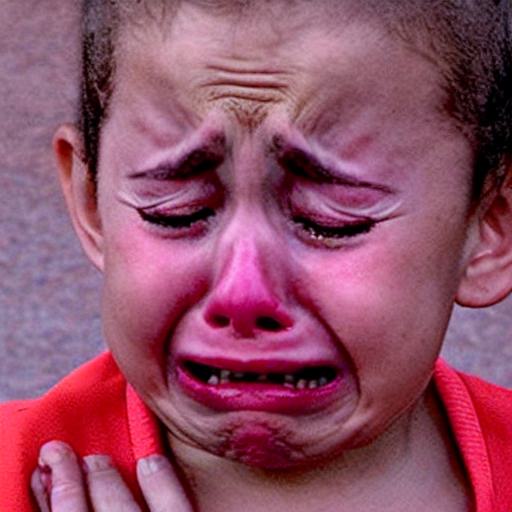}&
 \includegraphics[width=0.09\linewidth]{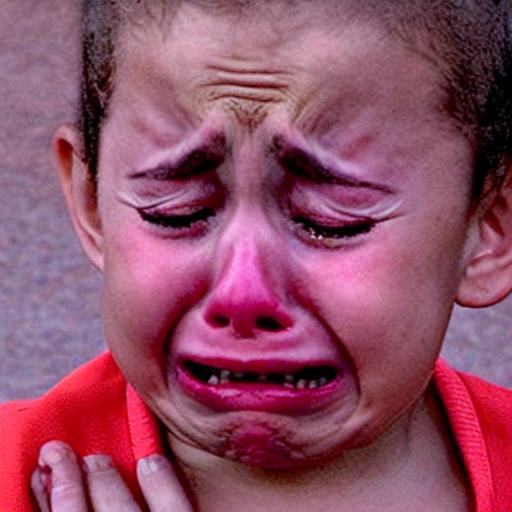}
 \\ 

 \rotatebox{90}{2D-Sad} &
 \includegraphics[width=0.09\linewidth]{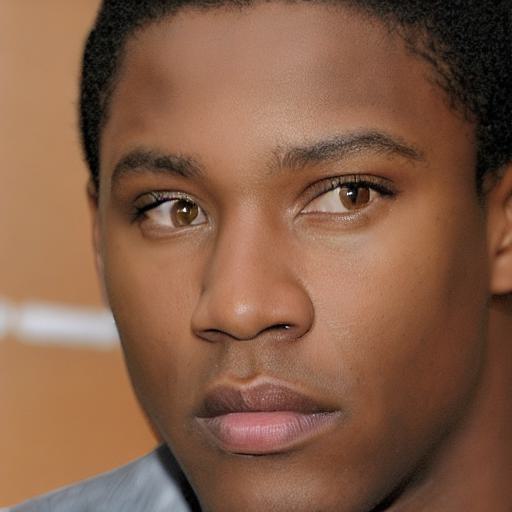}&
 \includegraphics[width=0.09\linewidth]{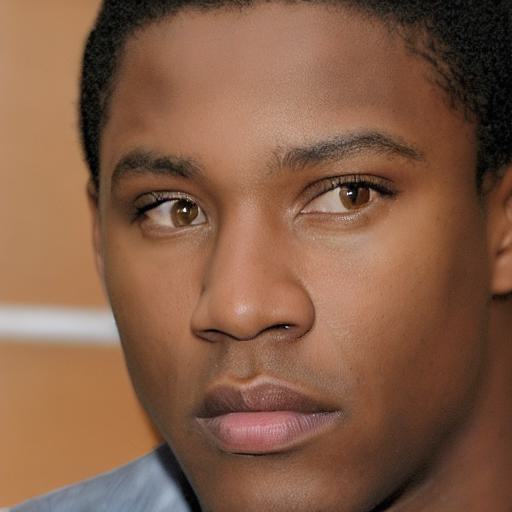}&
 \includegraphics[width=0.09\linewidth]{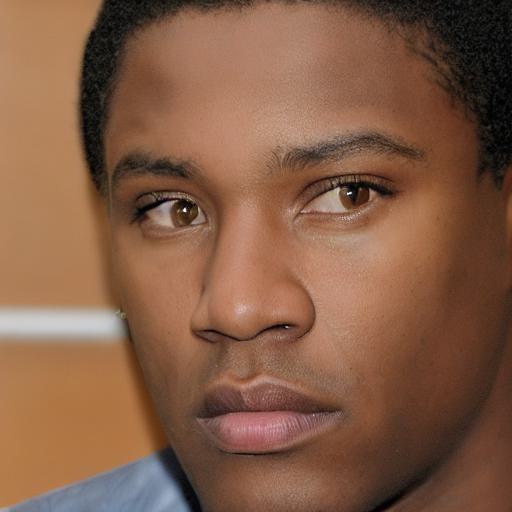}&
 \includegraphics[width=0.09\linewidth]{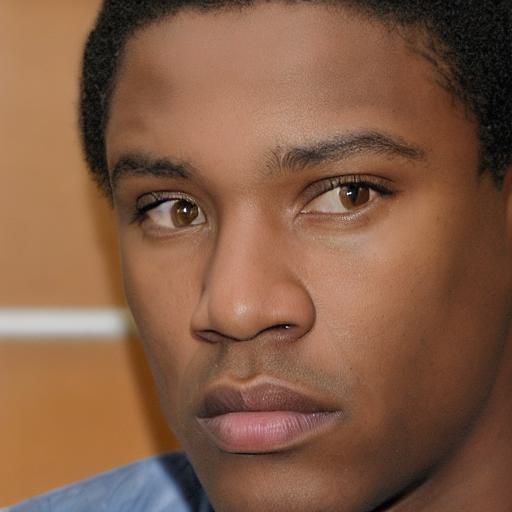}&
 \includegraphics[width=0.09\linewidth]{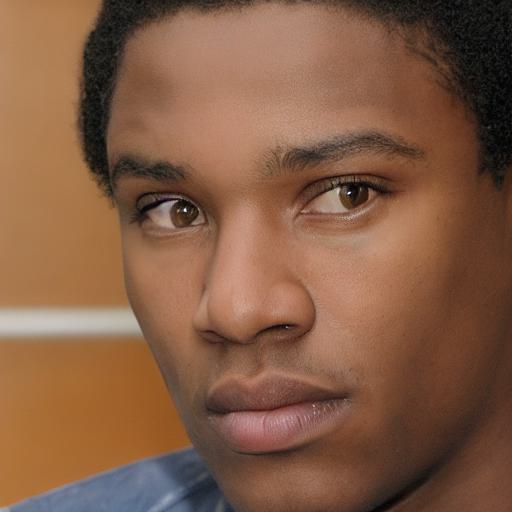}&
 \includegraphics[width=0.09\linewidth]{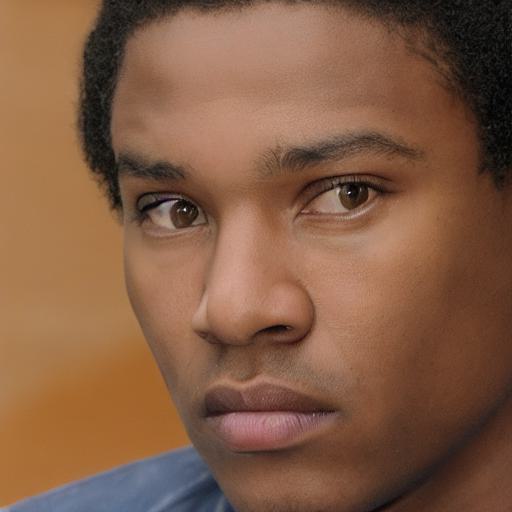}&
 \includegraphics[width=0.09\linewidth]{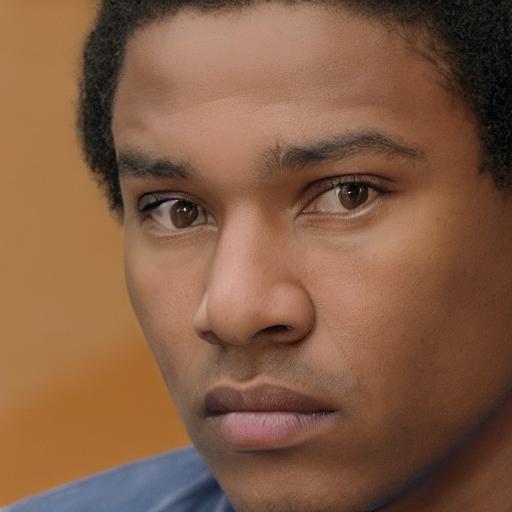}&
 \includegraphics[width=0.09\linewidth]{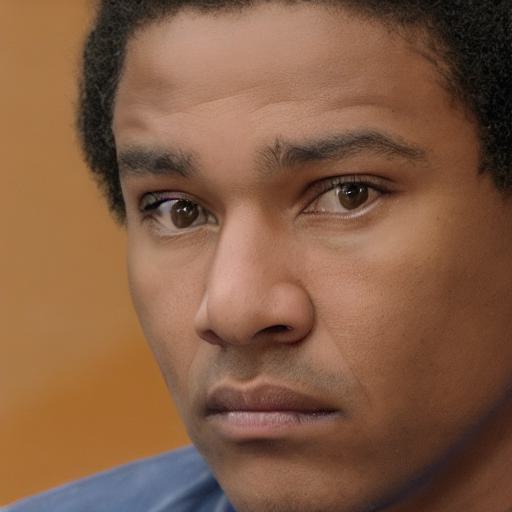}&
 \includegraphics[width=0.09\linewidth]{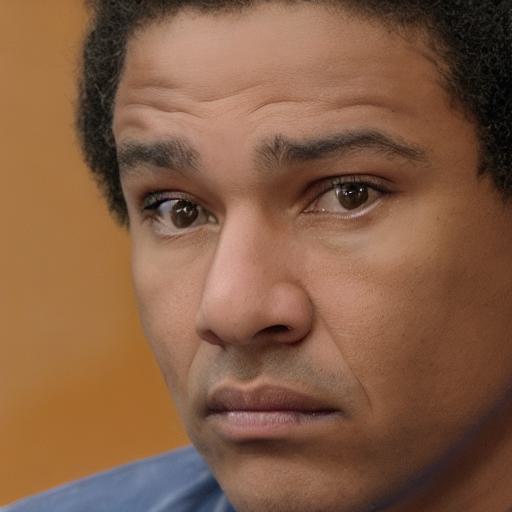}&
 \includegraphics[width=0.09\linewidth]{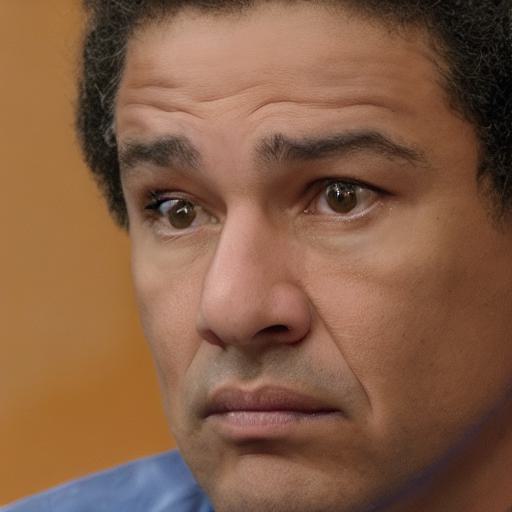}
 \\

\end{tabular}
    };
\end{tikzpicture}
\vspace{-0.8cm}
\caption{Both our 2D and 3D methods understand the emotions represented as continuous numbers. For 3D model, we showcase the behaviour towards the learned $Z$. These images illustrate that our learned representation is indeed continuous. Better viewed zoomed in. \label{fig:basic}}
\vspace{-3mm}
\end{figure*}

\begin{figure*}
\begin{tikzpicture}
    \node at (0,0) 
 {
 \addtolength{\tabcolsep}{-4.5pt}
\begin{tabular} {ccccccccccc}
 & Happy disgd. & Sad-fear & Sad-angry & Fear-disgd. & Angry-surpd. &Happy-fear & Sad-surpd.& Awed& Hatred\\

 \rotatebox{90}{3D (Ours)} &
 \includegraphics[width=0.1\linewidth]{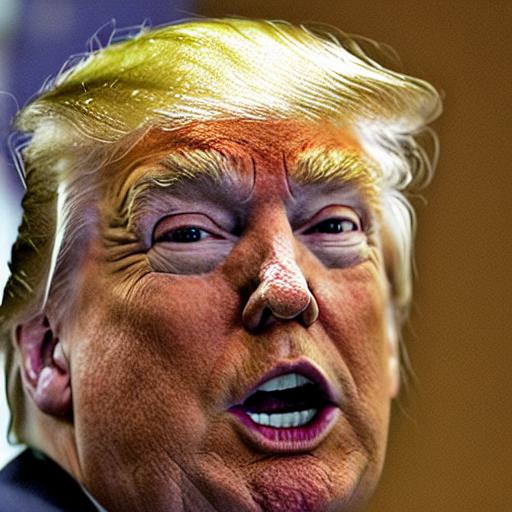}&
 \includegraphics[width=0.1\linewidth]{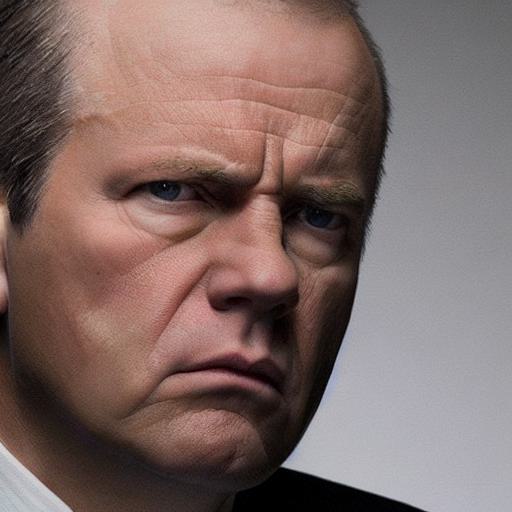}&
 \includegraphics[width=0.1\linewidth]{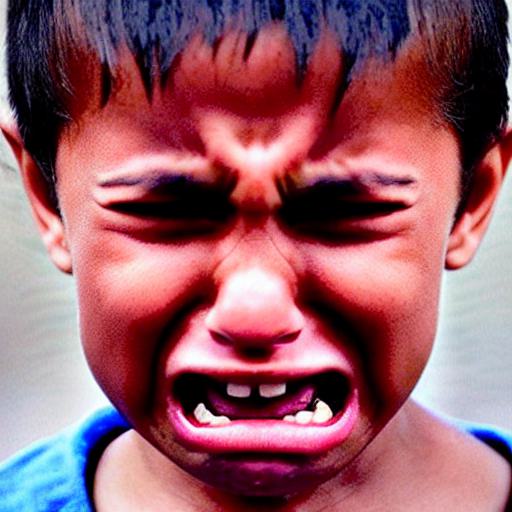}&
 \includegraphics[width=0.1\linewidth]{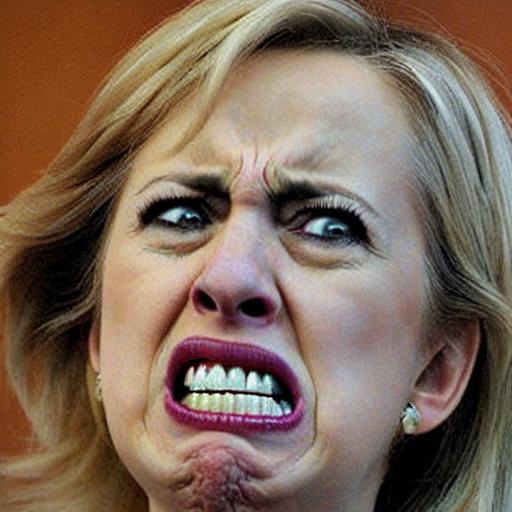}&
 \includegraphics[width=0.1\linewidth]{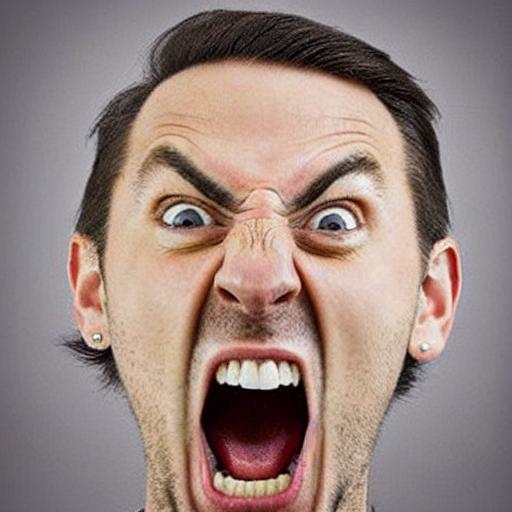}&
 \includegraphics[width=0.1\linewidth]{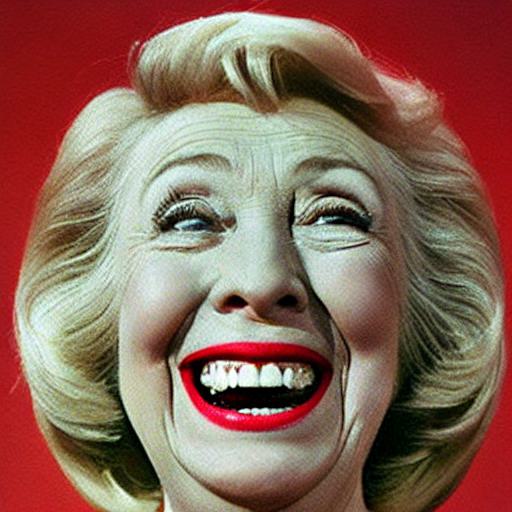}&
 \includegraphics[width=0.1\linewidth]{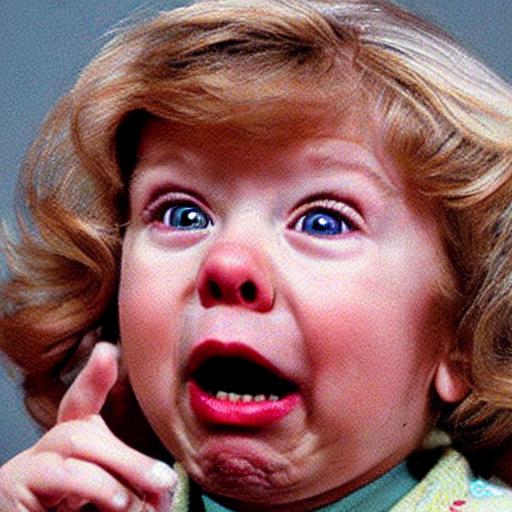}&
 \includegraphics[width=0.1\linewidth]{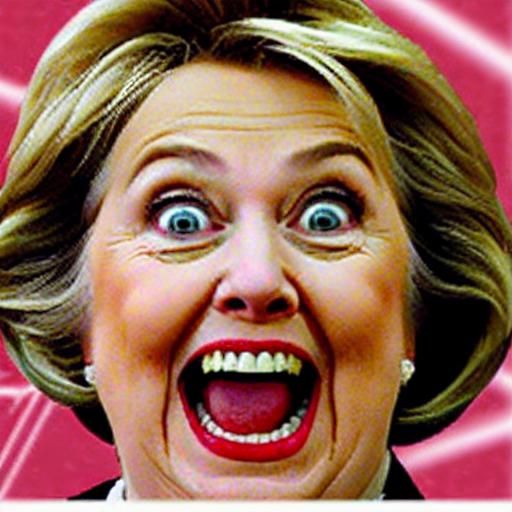}&
 \includegraphics[width=0.1\linewidth]{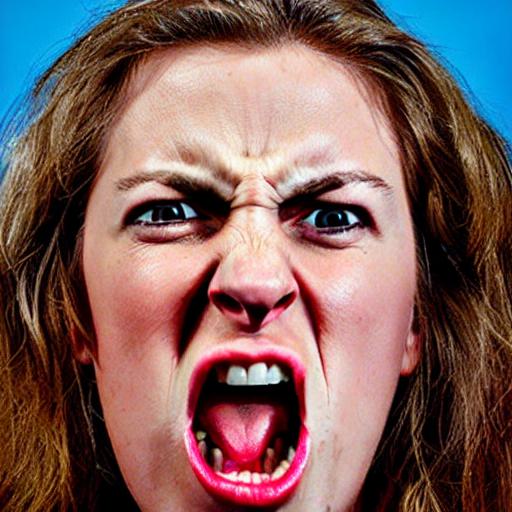}
 \\ 

 \rotatebox{90}{2D model} &
 \includegraphics[width=0.1\linewidth]{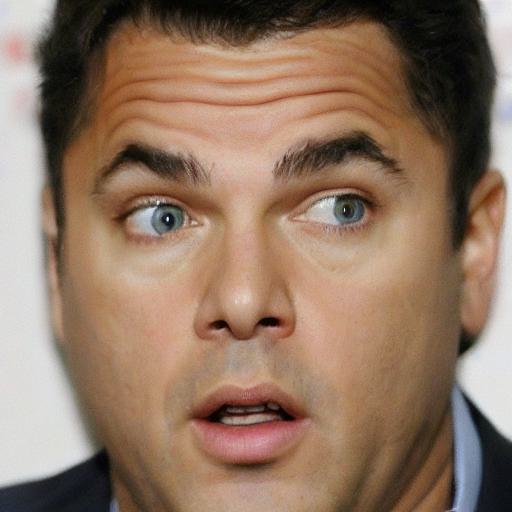}&
 \includegraphics[width=0.1\linewidth]{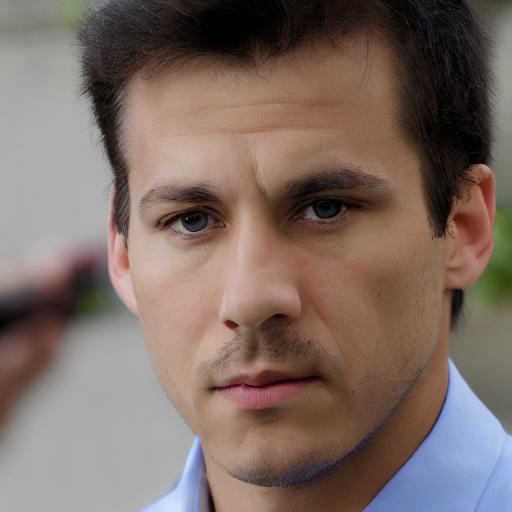}&
 \includegraphics[width=0.1\linewidth]{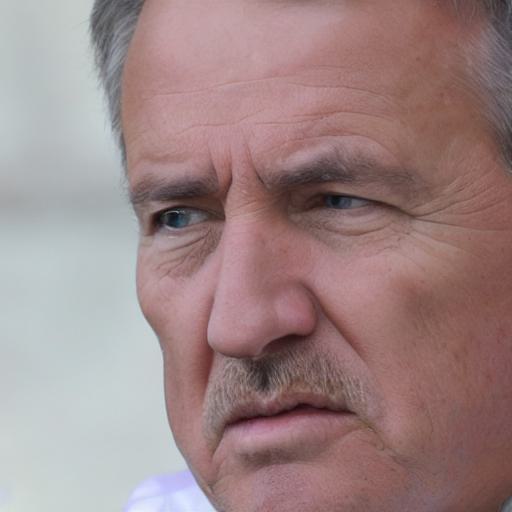}&
 \includegraphics[width=0.1\linewidth]{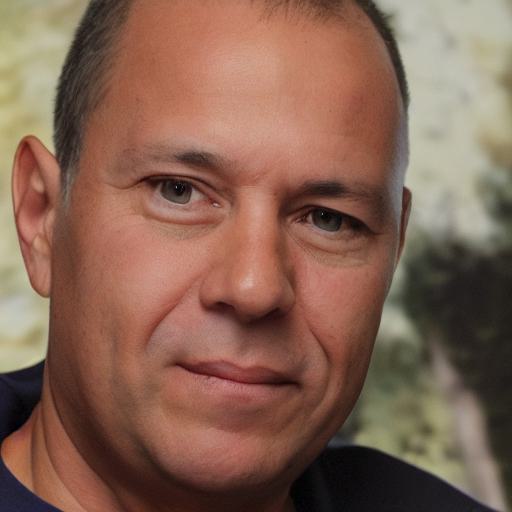}&
 \includegraphics[width=0.1\linewidth]{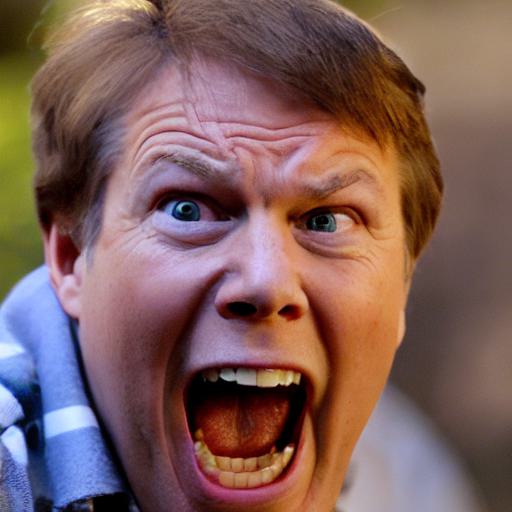}&
 \includegraphics[width=0.1\linewidth]{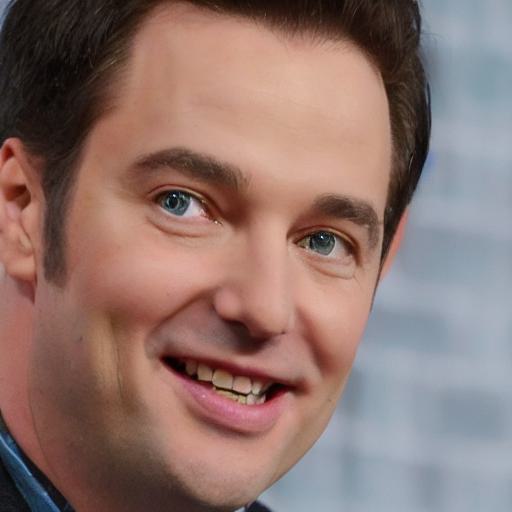}&
 \includegraphics[width=0.1\linewidth]{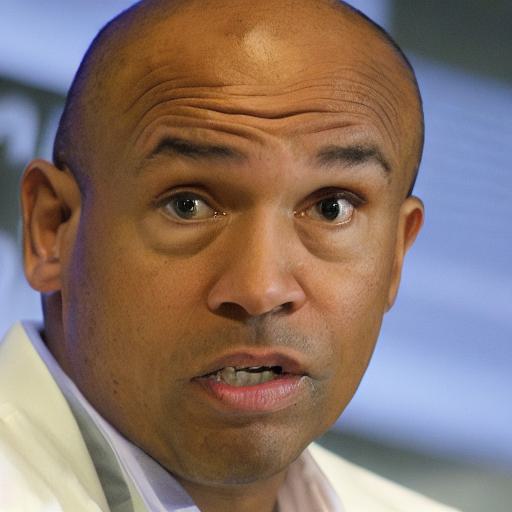}&
 \includegraphics[width=0.1\linewidth]{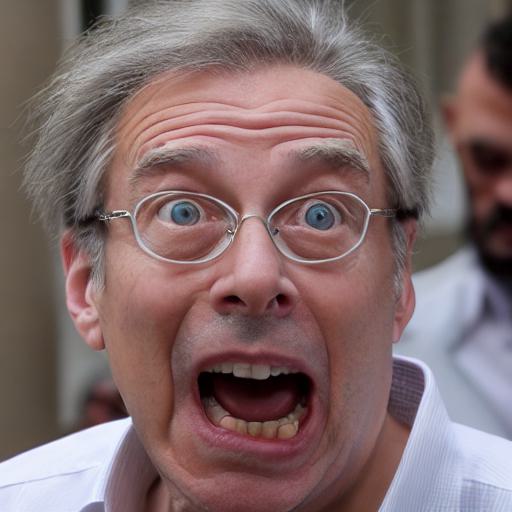}&
 \includegraphics[width=0.1\linewidth]{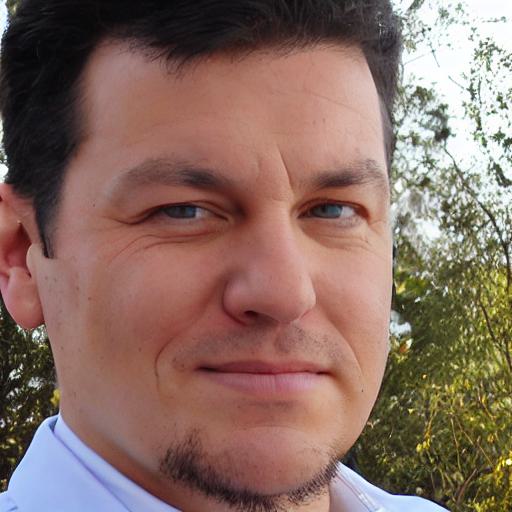}
 \\

 \rotatebox{90}{\scriptsize  DreamBooth} &
 \includegraphics[width=0.1\linewidth]{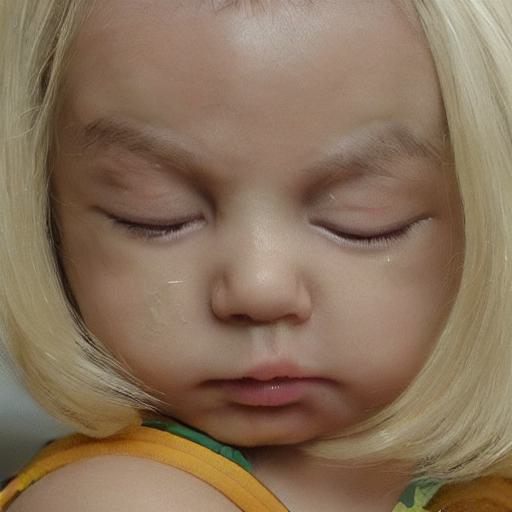}&
 \includegraphics[width=0.1\linewidth]{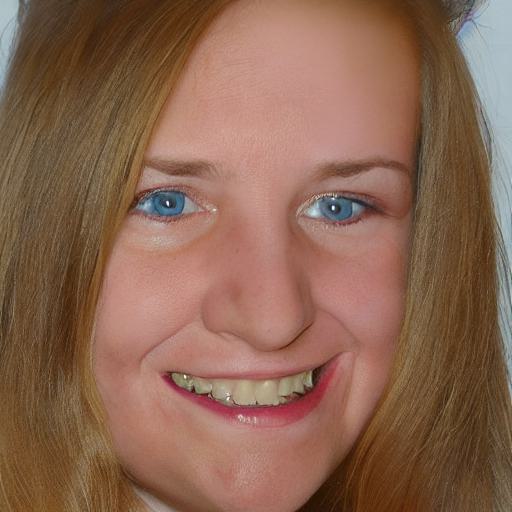}&
 \includegraphics[width=0.1\linewidth]{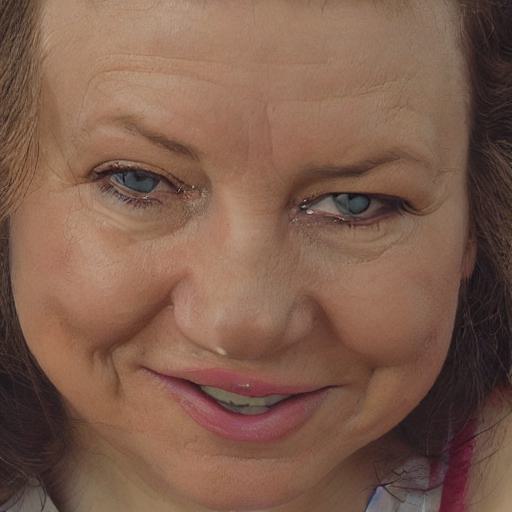}&
 \includegraphics[width=0.1\linewidth]{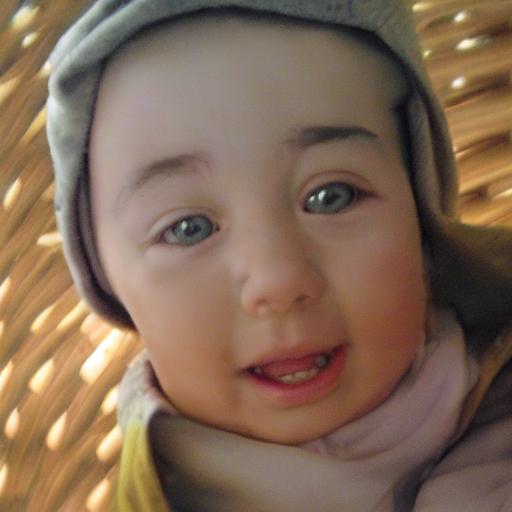}&
 \includegraphics[width=0.1\linewidth]{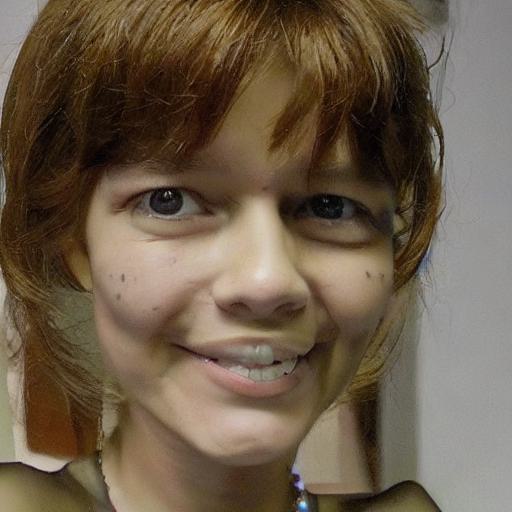}&
 \includegraphics[width=0.1\linewidth]{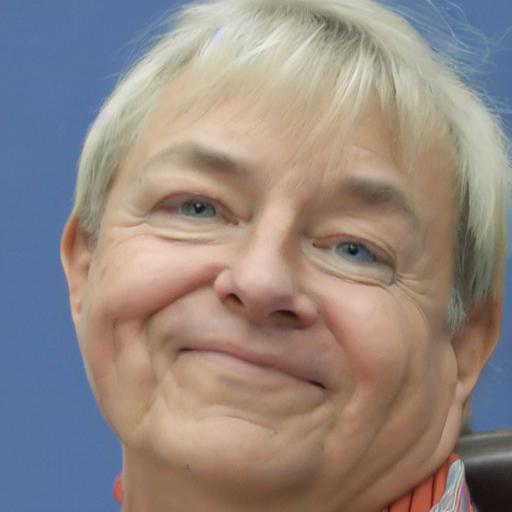}&
 \includegraphics[width=0.1\linewidth]{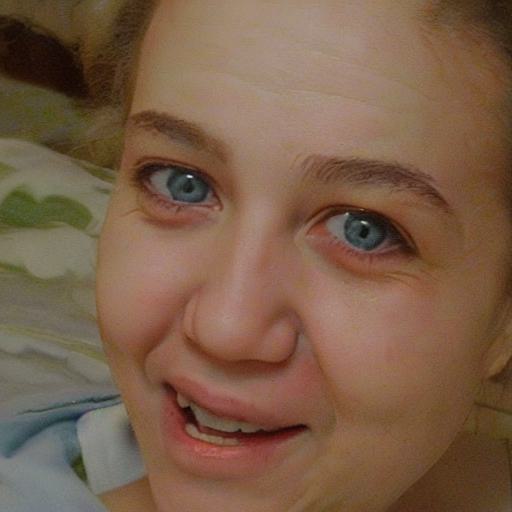}&
 \includegraphics[width=0.1\linewidth]{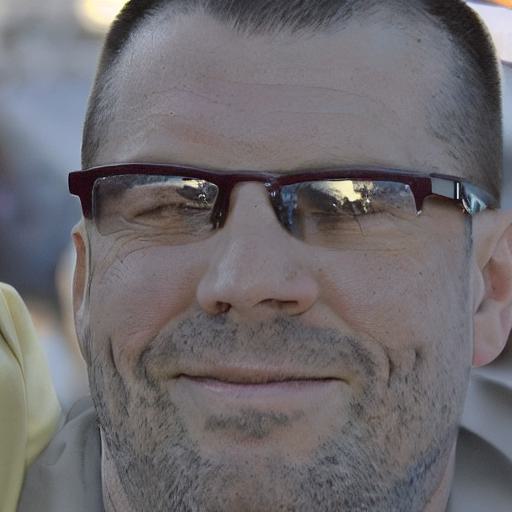}&
 \includegraphics[width=0.1\linewidth]{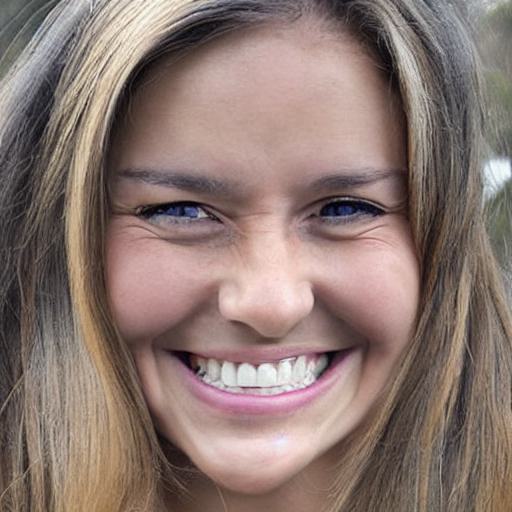}
 \\

  \rotatebox{90}{DALL.E 3} &
 \includegraphics[width=0.1\linewidth]{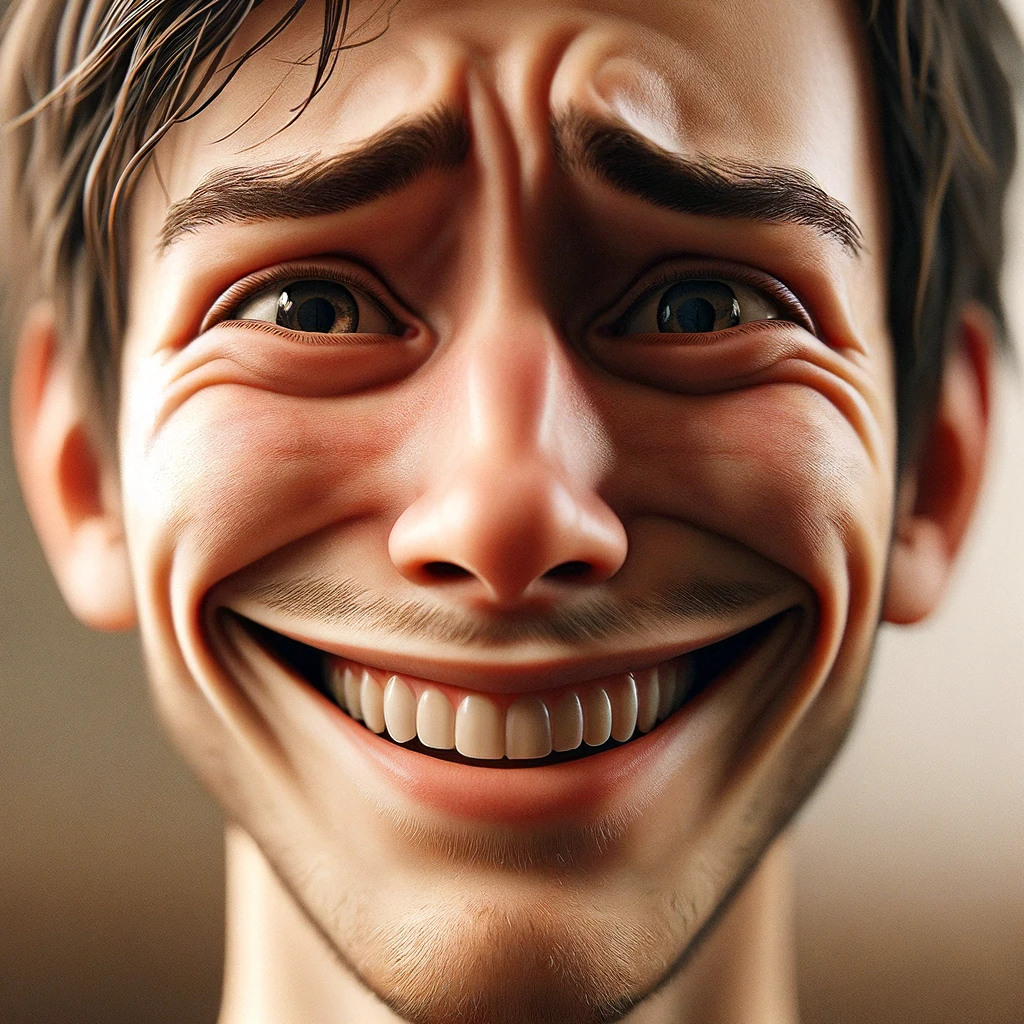}&
 \includegraphics[width=0.1\linewidth]{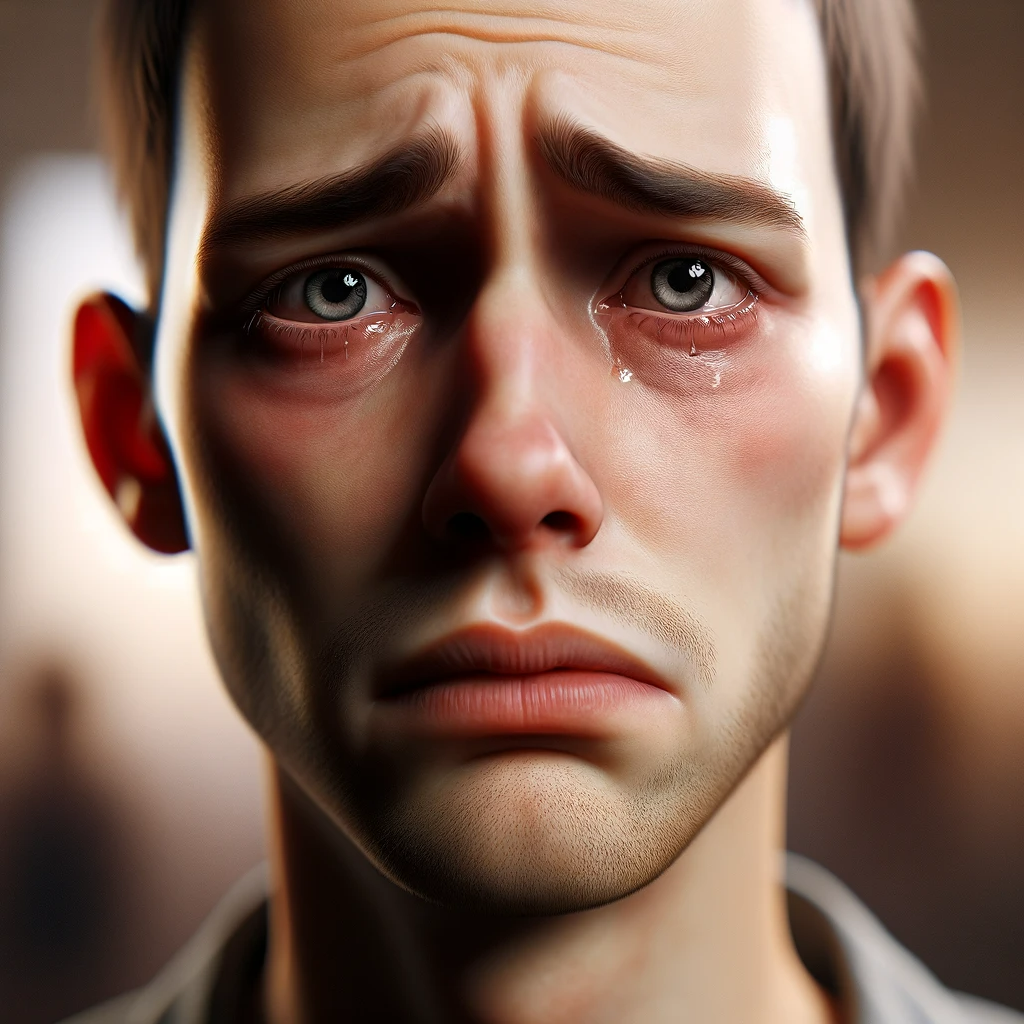}&
 \includegraphics[width=0.1\linewidth]{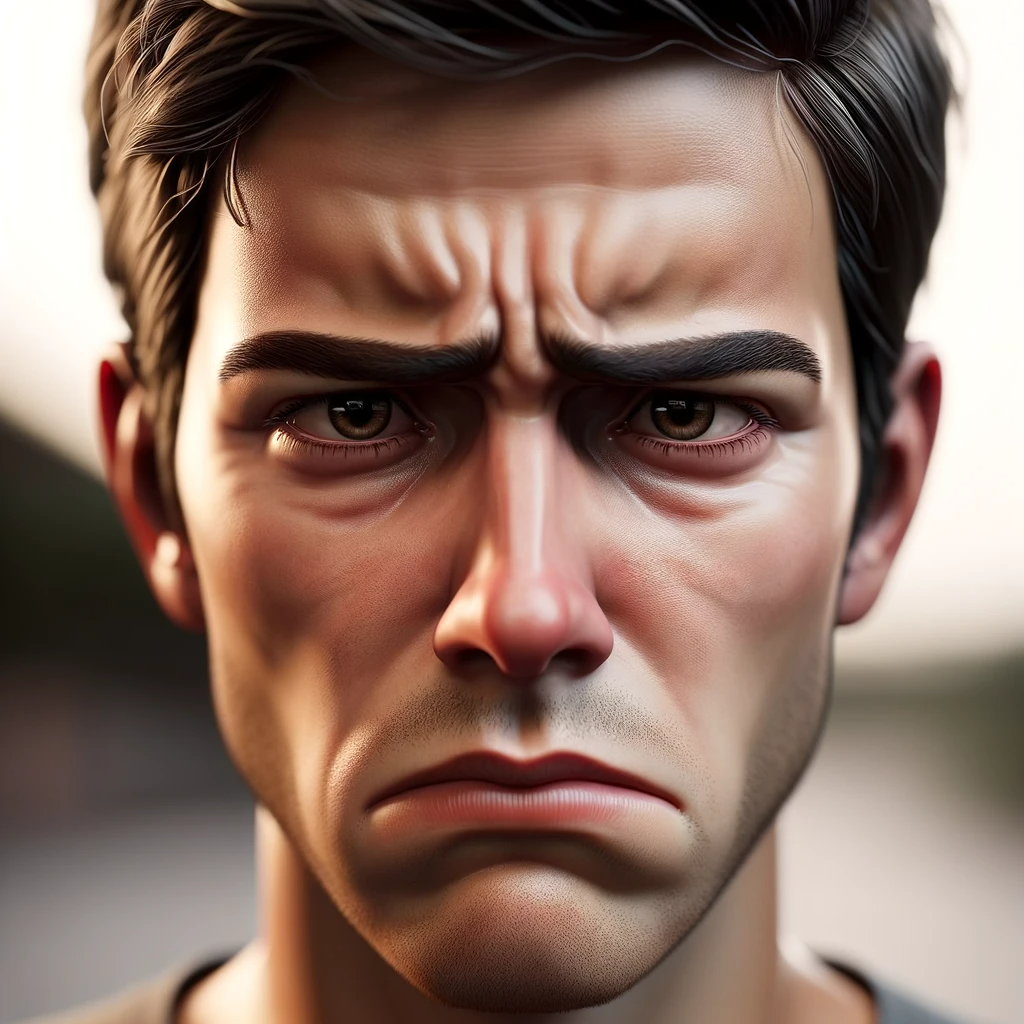}&
 \includegraphics[width=0.1\linewidth]{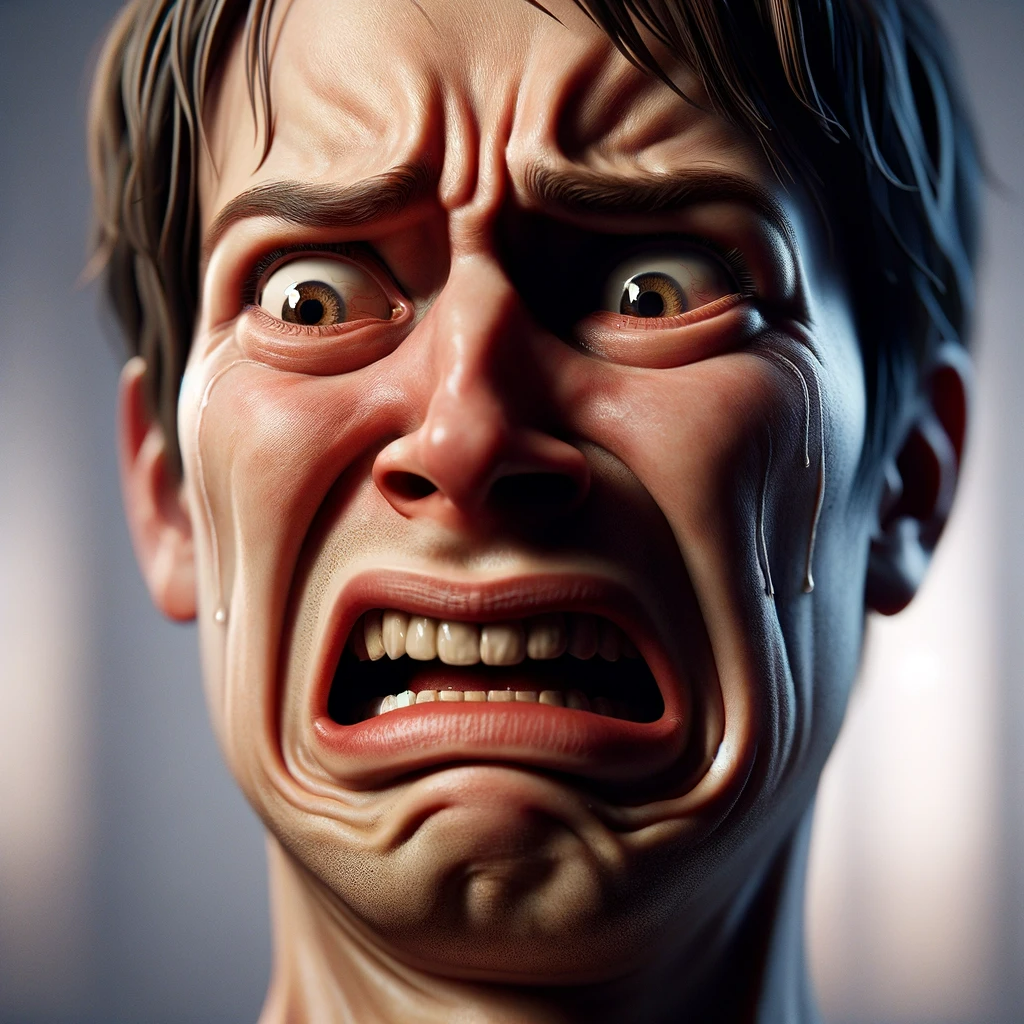}&
 \includegraphics[width=0.1\linewidth]{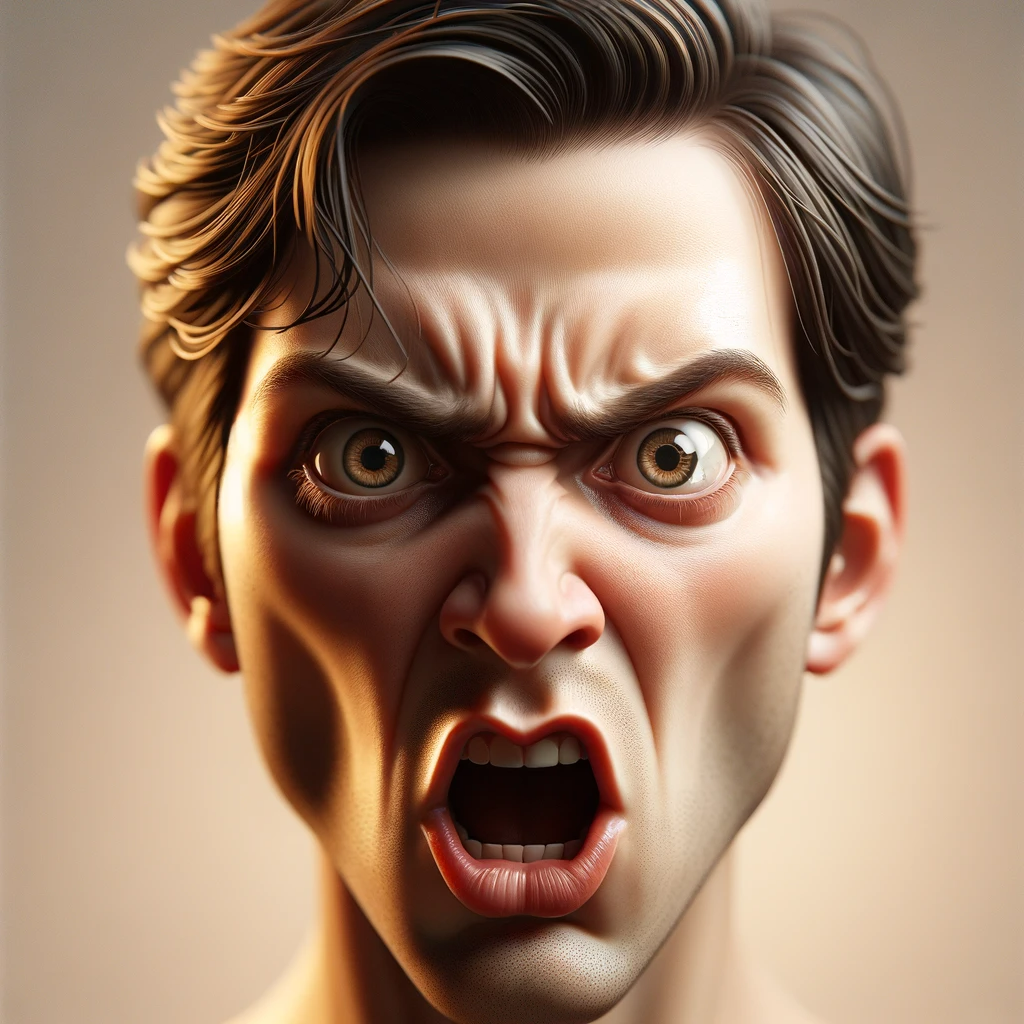}&
 \includegraphics[width=0.1\linewidth]{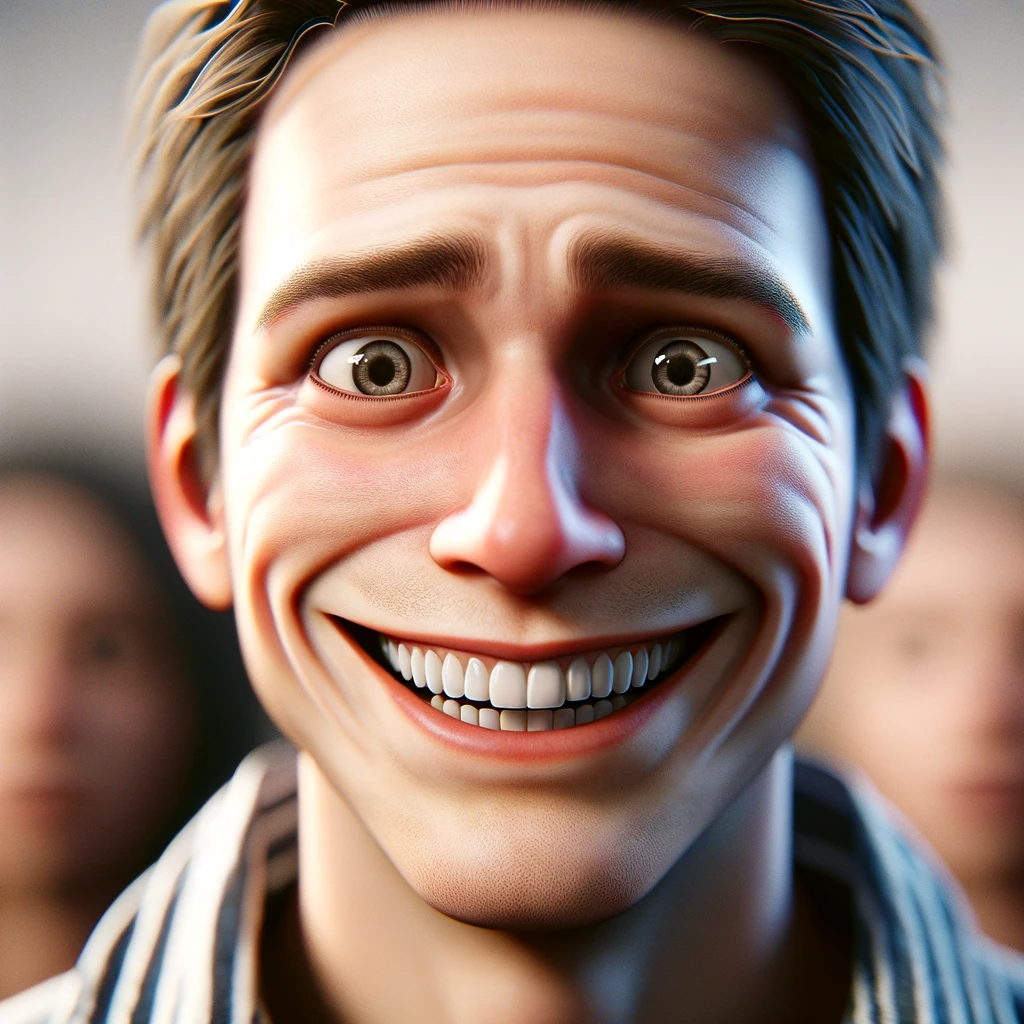}&
 \includegraphics[width=0.1\linewidth]{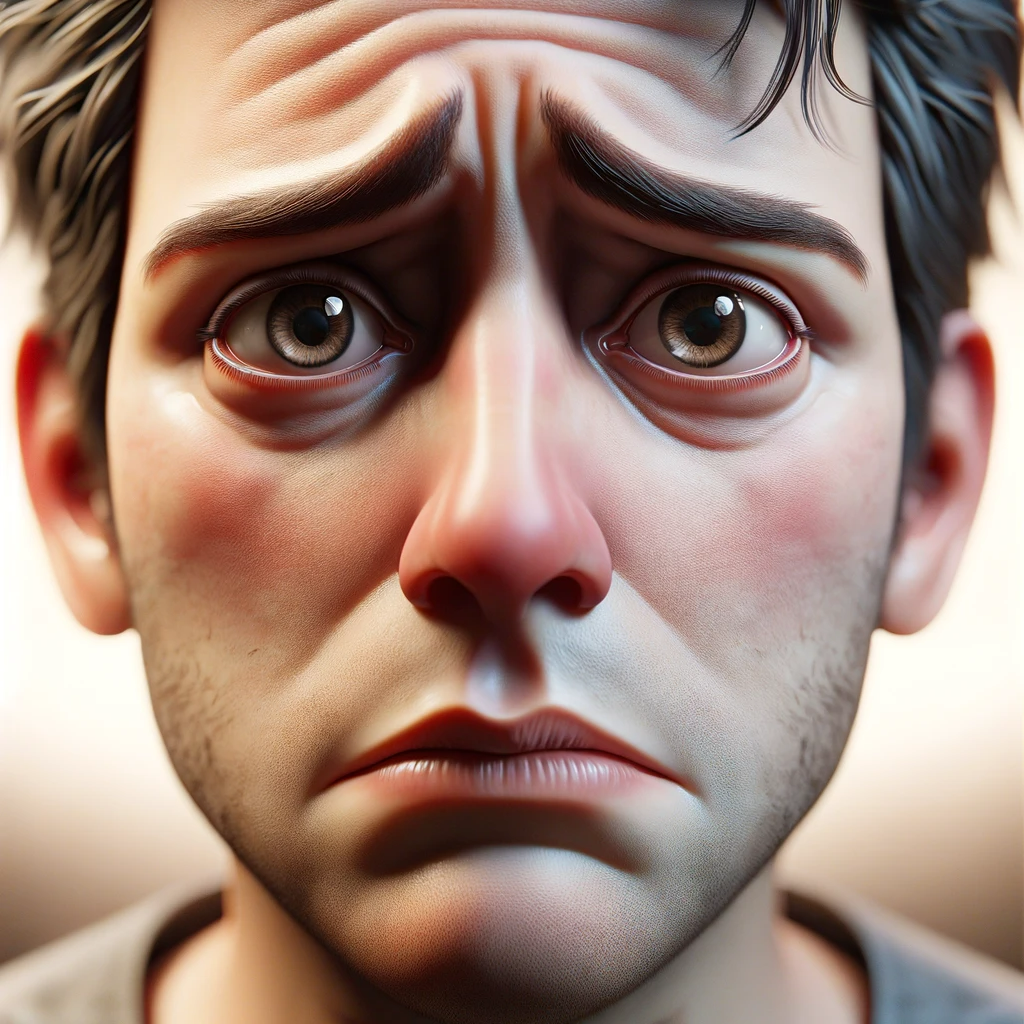}&
 \includegraphics[width=0.1\linewidth]{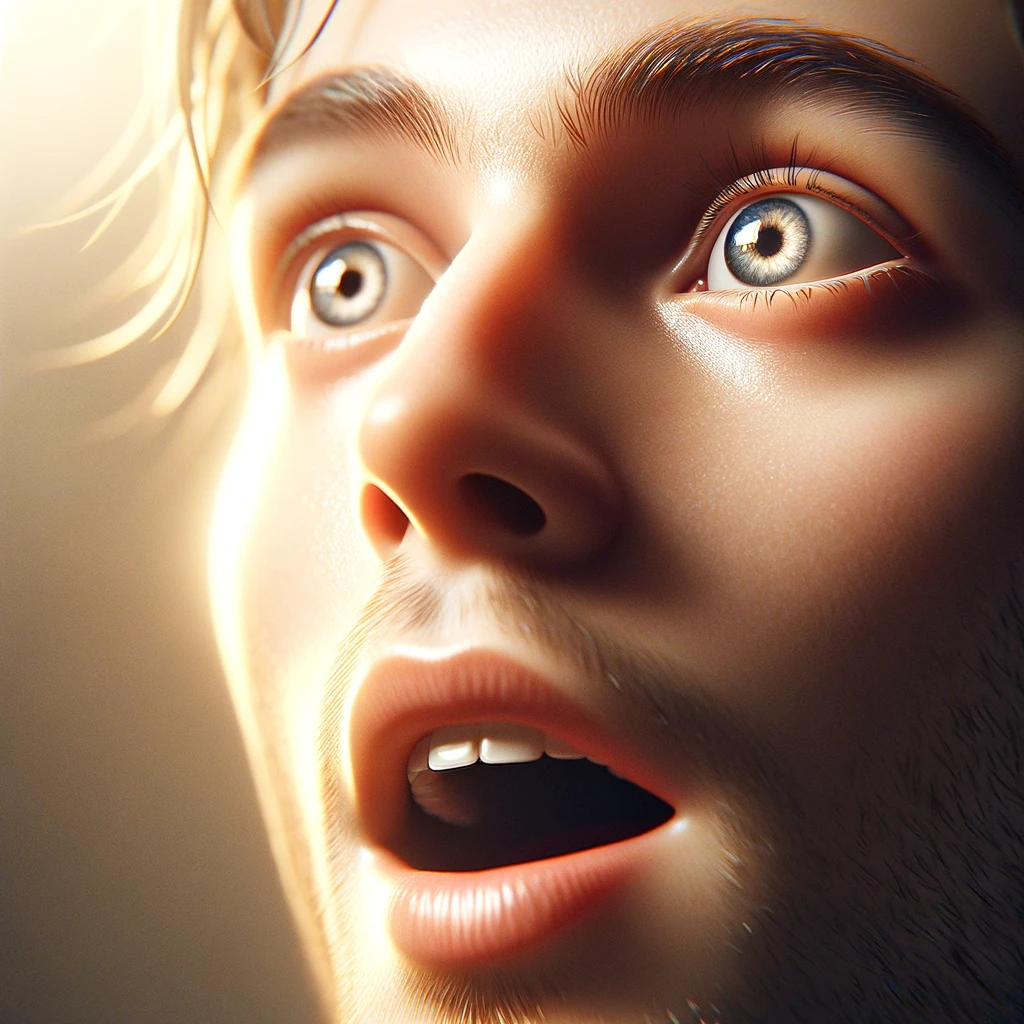}&
 \includegraphics[width=0.1\linewidth]{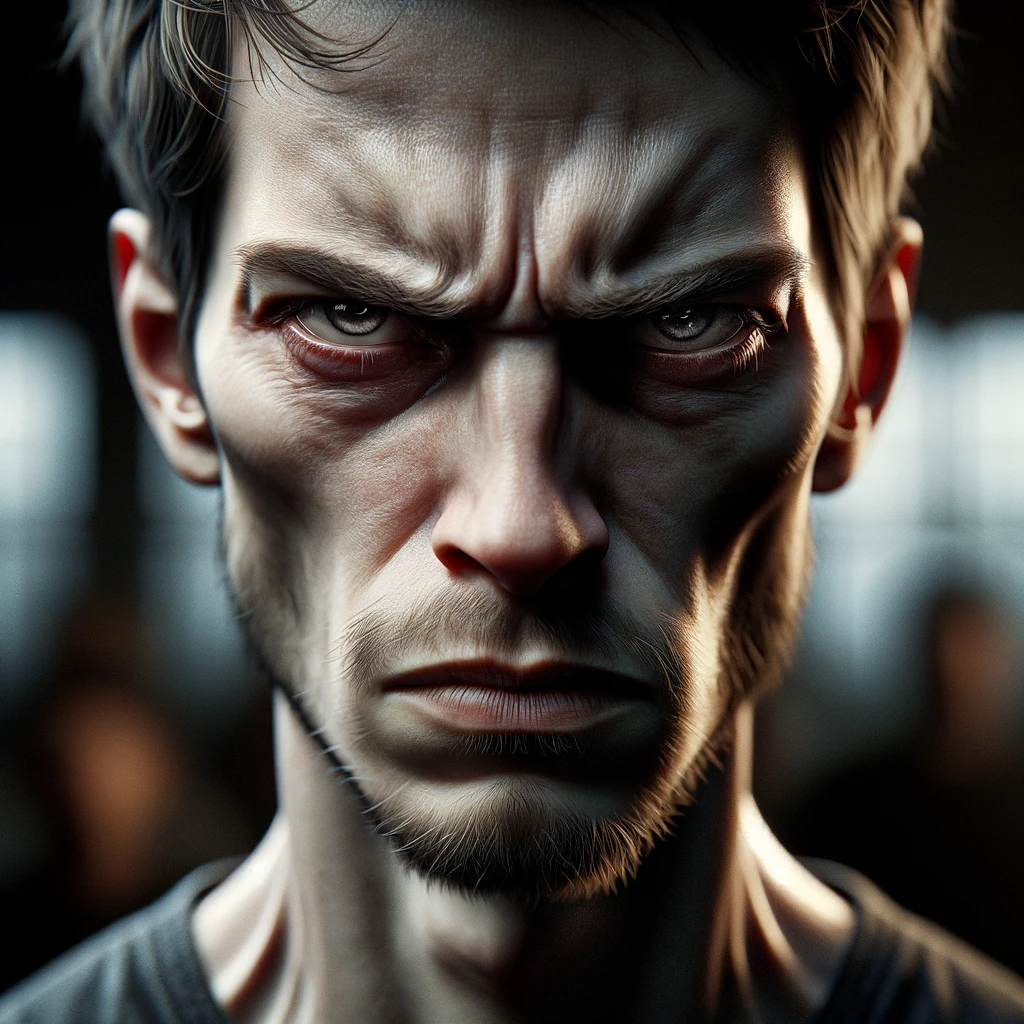}
 \\ 
  \rotatebox{90}{\scriptsize Stable Diffusion} &
 \includegraphics[width=0.1\linewidth]{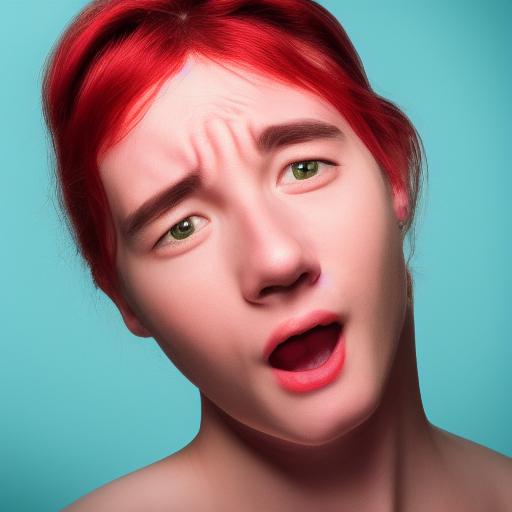}&
 \includegraphics[width=0.1\linewidth]{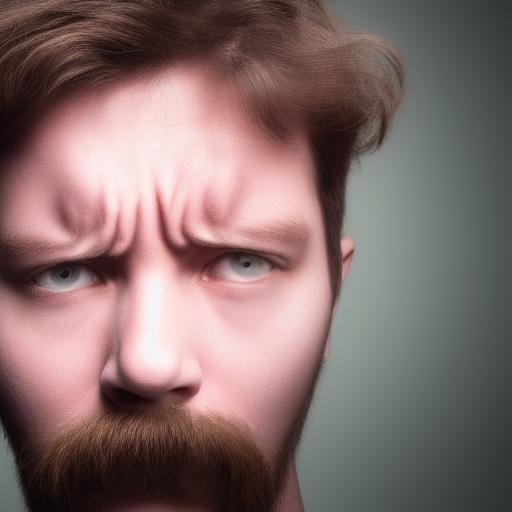}&
 \includegraphics[width=0.1\linewidth]{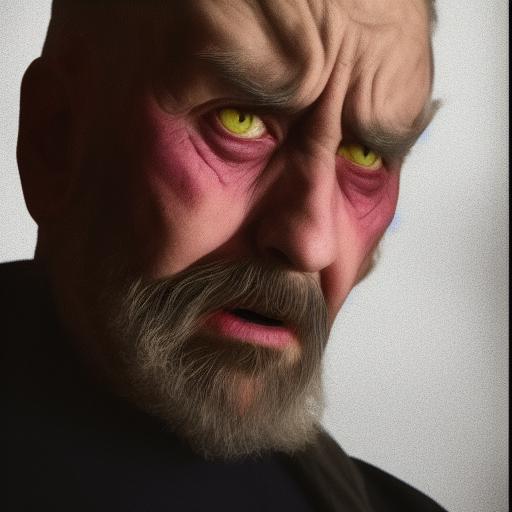}&
 \includegraphics[width=0.1\linewidth]{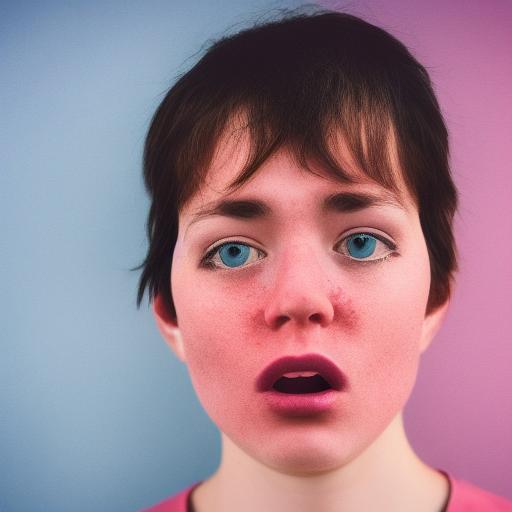}&
 \includegraphics[width=0.1\linewidth]{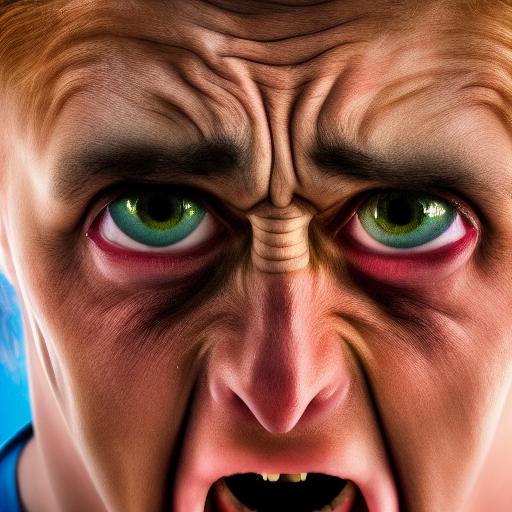}&
 \includegraphics[width=0.1\linewidth]{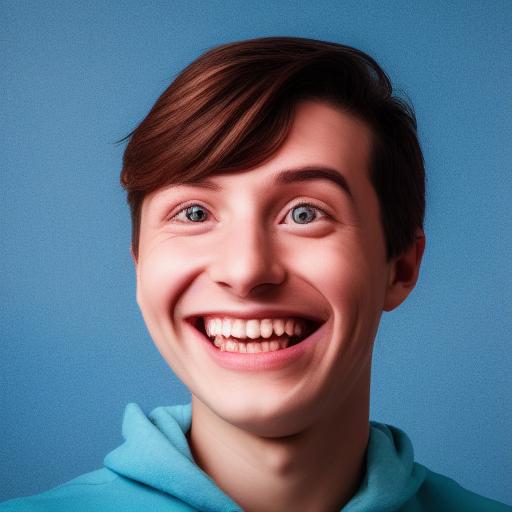}&
 \includegraphics[width=0.1\linewidth]{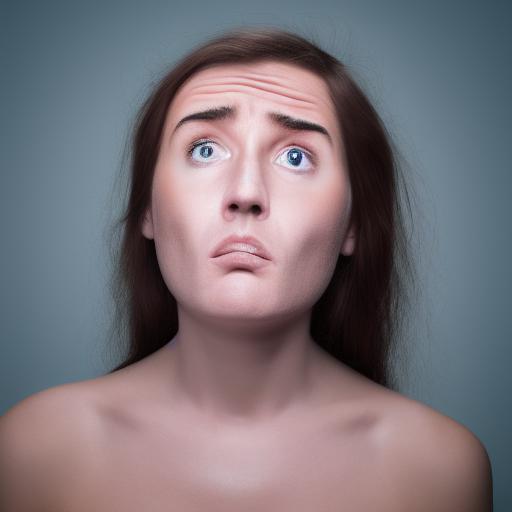}&
 \includegraphics[width=0.1\linewidth]{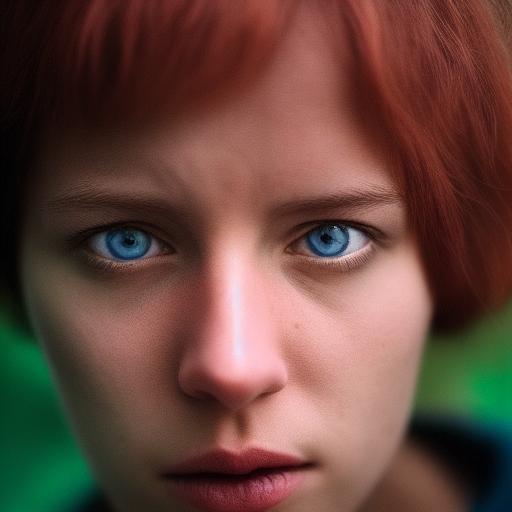}&
 \includegraphics[width=0.1\linewidth]{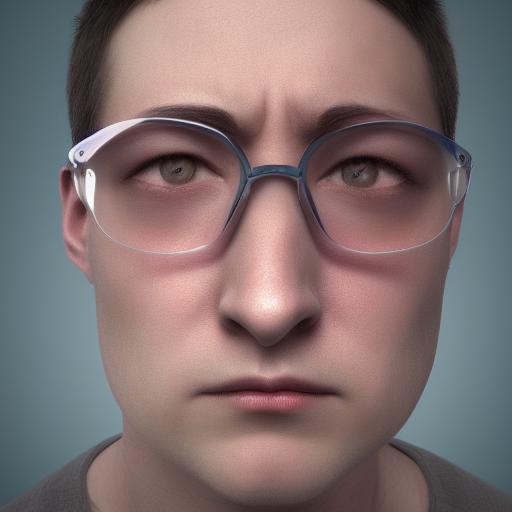}

\end{tabular}
    };
\end{tikzpicture}
\vspace{-0.8cm}
\caption{Nine compound emotions that can be represented by our 3-dimensional C2A2 but cannot be represented by 2-dimensional AV model (please refer Table~\ref{tab:2d-3D_emotions} to associated the emotions (top, here) to the 3D vs. 2D representations). When compared with other methods, our 3D-based representation is clearly superior, thanks to its richer representation and the continuous number understanding capabilities. \label{fig:full}}
\vspace{-0.3cm}
\end{figure*}

\begin{figure}
\vspace{-5mm}
\resizebox{1.0\linewidth}{!}
{
\begin{tikzpicture}
    \node at (0,0) 
 {
 \addtolength{\tabcolsep}{-4.5pt}
\begin{tabular} {ccccc}
 & Hat & Glasses & Mustache & Lipstick \\
 \rotatebox{90}{Happy-surpd.} &
 \includegraphics[width=0.22\linewidth]{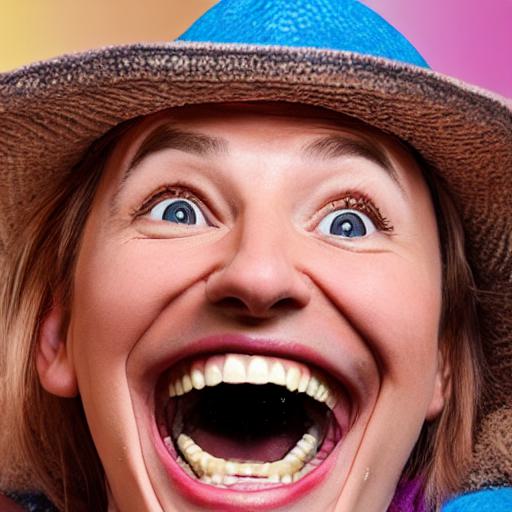}&
 \includegraphics[width=0.22\linewidth]{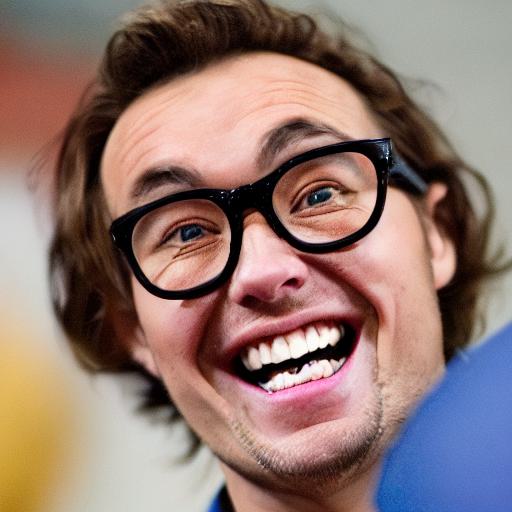}&
 \includegraphics[width=0.22\linewidth]{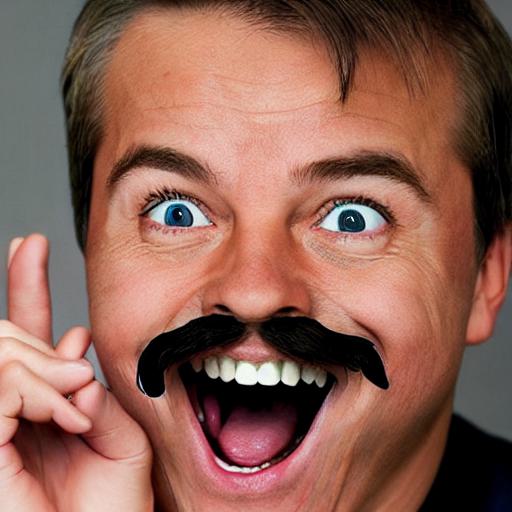}&
 \includegraphics[width=0.22\linewidth]{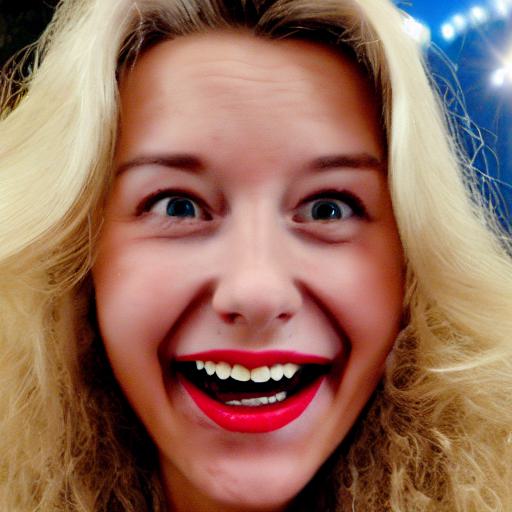}
 \\ 
 \rotatebox{90}{Sad-fearful} &
 \includegraphics[width=0.22\linewidth]{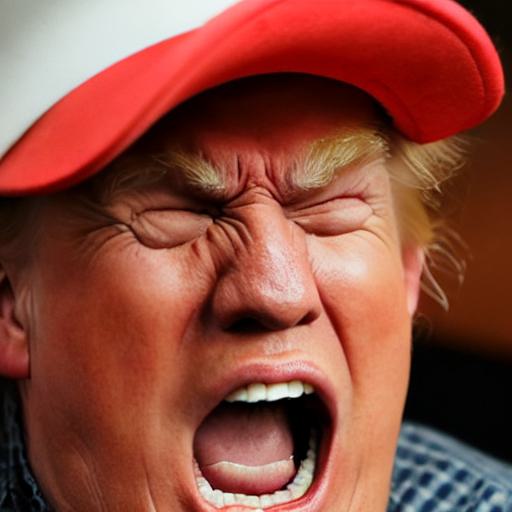}&
 \includegraphics[width=0.22\linewidth]{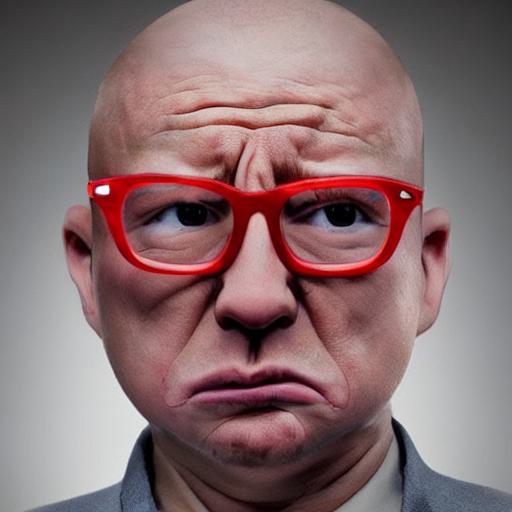}&
 \includegraphics[width=0.22\linewidth]{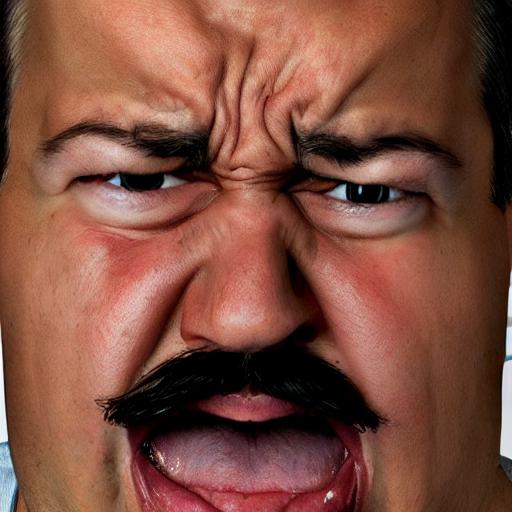}&
 \includegraphics[width=0.22\linewidth]{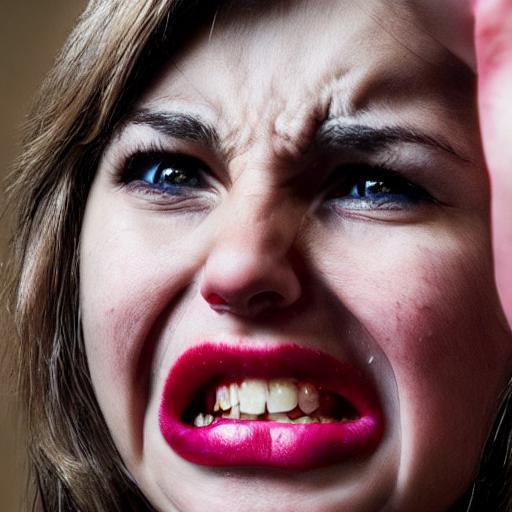} \\
 \rotatebox{90}{Happy-sad} &
 \includegraphics[width=0.22\linewidth]{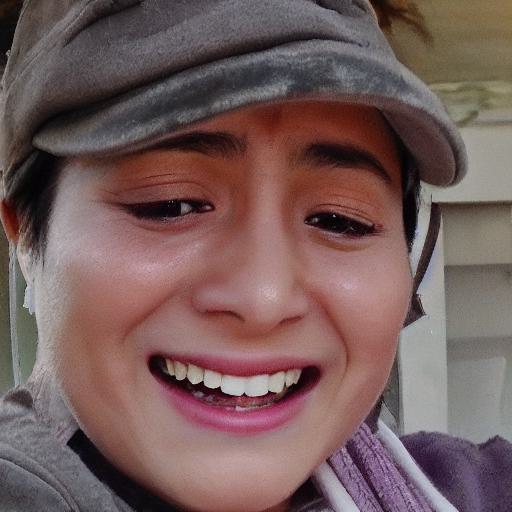}&
 \includegraphics[width=0.22\linewidth]{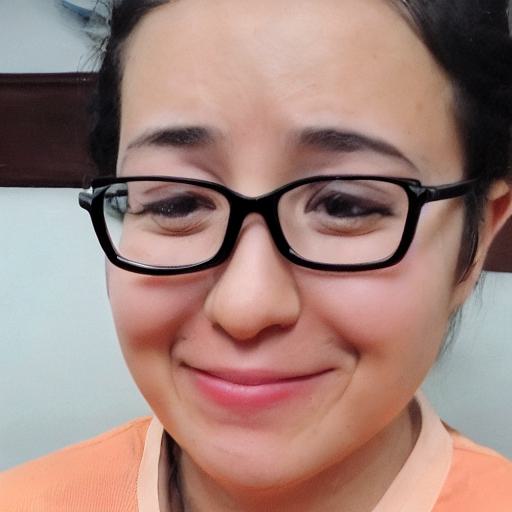}&
 \includegraphics[width=0.22\linewidth]{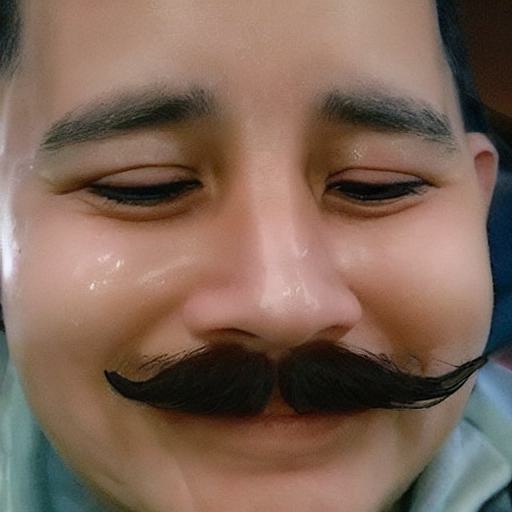}&
 \includegraphics[width=0.22\linewidth]{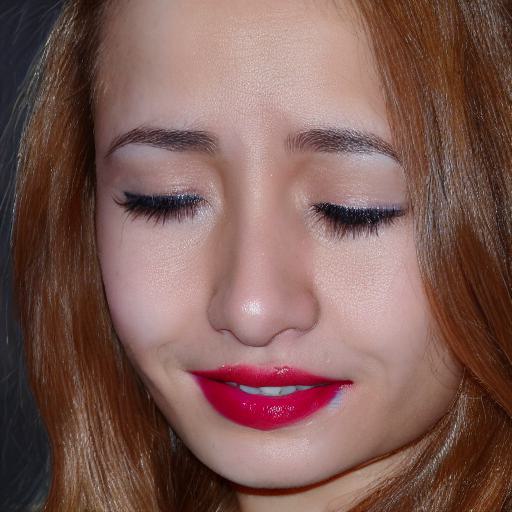} 
\end{tabular}
    };
\end{tikzpicture}
}
\vspace{-7mm}
\caption{Our method can preserve the attributes from the base network. This allows us to perform meaningful number+text-to-image generation. Expression for Compound emotions (left) represented in numbers are generated with text attributes (top).\label{fig:attributes}}
\vspace{-5mm}
\end{figure}

\section{Experiments}
\noindent\textbf{Datatest.} The training of our models and baselines utilize AffectNet~\cite{affectnet}, recognized as the most extensive dataset in affect computing, comprising $\approx$ 1M images sourced from the Internet. Searches on prominent search engines were conducted using 1250 keywords linked to emotions across six languages. Remarkably, 450K of these images received manual annotations from 12 specialists, categorizing them into basic emotions and AV labels. Given our objective to generate intricate and nuanced emotions in a highly varied context, this dataset emerged as the perfect selection.

\label{sec:experiments}

\noindent\textbf{Implementation details.} We implemented two parts proposed in this paper in two different settings. The first part of our implementation is based on GANmut~\cite{ganmut}.  
We set the number of training iterations to 1M.
We adopted the same training strategy of GANmut, and the same hyper-parameters. For the second part, we implemented a Dreambooth-like approach on top of the Stable Diffusion model. We performed 2 experiments where we generated emotions from the 2D and the 3D spaces. 
We also tried several variation of text+number inputs, with limited success. That is why we use both a text and a number embeddings. During the training and the generation process we used ``sks" as a placeholder token which is used to identify a specific emotion. We experimented with different text prompts such as ``a photo of a sks person" and ``a colourful photo of a sks person", and came to the conclusion that the best performing one is ``a high-quality realistic color photo of a sks person", so we used it as a text prompt in all of our experiments. For regularization we used 100K randomly chosen images from the training datasets with the same prompt and all of the coordinates corresponding to their emotions were set to 0.

\noindent\textbf{Evaluation metric. } 
Similar to GANmut, we employ the modified Fréchet Inception Distance \cite{conf/nips/HeuselRUNH17}, termed as Fréchet Emotion Distance (FED), for assessing emotions. For calculating FED, we trained VGGNet~\cite{VGG} on AffectNet for emotion classification. This involves inputting real and generated images into VGGNet and extracting features proximal to the ultimate classifier. We then assume Gaussian distributions for both feature sets and compute their Fréchet distance. Our goal is to minimize the FED value. To determine FED, we uniformly randomly sample images in every instance. More evaluation methods are also presented, detailed in their respective subsections. For approximating human emotion assessment, we utilize the softmax score from the trained VGGNet. Additionally, we also engage 8 psychologists for evaluation through an expert user study.

\begin{table*}[b]
\vspace{-3mm}
\resizebox{1.0\linewidth}{!}{
    \scriptsize
    \centering
    \begin{tabular}{|c|c|c|c|c|c|c|c|c|c|c|}
    \hline
    \textbf{Model} & \textbf{Happy-disgd.} & \textbf{Sad-Fearful}  & \textbf{Sad-Angry} & \textbf{Fearful-Disgd.} & \textbf{Angry-Surpd.} & \textbf{Happy-Fearful} & \textbf{Sad-Surpd.} & \textbf{Average} & \textbf{Average Overall}\\
    \hline
    \hline

    3D model (Ours)& \textbf{3.51}&	\textbf{4.06}	& \textbf{4.18}	& \textbf{4.52}& \textbf{4.74}&	\textbf{4.32}&	\textbf{4.02} & \textbf{4.19} & \textbf{4.03}
\\ 
    \hline

    2D-AV model & 2.71 &	3.00&	2.67&	3.09	&3.67&	3.41	&2.50 & 3.01 & 3.50
 \\

    \hline
    AUs model & 1.17 &1.55&1.77&	1.83	& 1.60 &	1.69	&1.92 & 1.76 & 1.87
 \\
     \hline
    Stable Diffusion & 1.32 &	2.15&	2.31&	2.31	&1.89&	2.50	&2.07 & 2.04 & 2.21
 \\
    \hline
    DALL.E 3 & 2.25 &3.43&3.45&	3.36	& 3.94 &	3.04	&2.60 & 3.31 & 3.72
 \\
 \hline
    DreamBooth & 1.94 & 1.32&1.07&	1.33	& 1.23&	1.14	&1.59 & 1.47& 1.59
 \\
    \hline
    \end{tabular}
        }
        \vspace{-2mm}
    \caption{Average ratings (out of 1-5) for 7 compound emotions that can be represented by our 3D C2A2 but cannot be represented by the 2D-AV model, the average of these 7 emotions (second last) and the average  among all emotions (last column) used in our study.}
    \label{tab:qant_results_2}
\end{table*}

\begin{figure}
    \centering
    \includegraphics[width=0.44\textwidth]{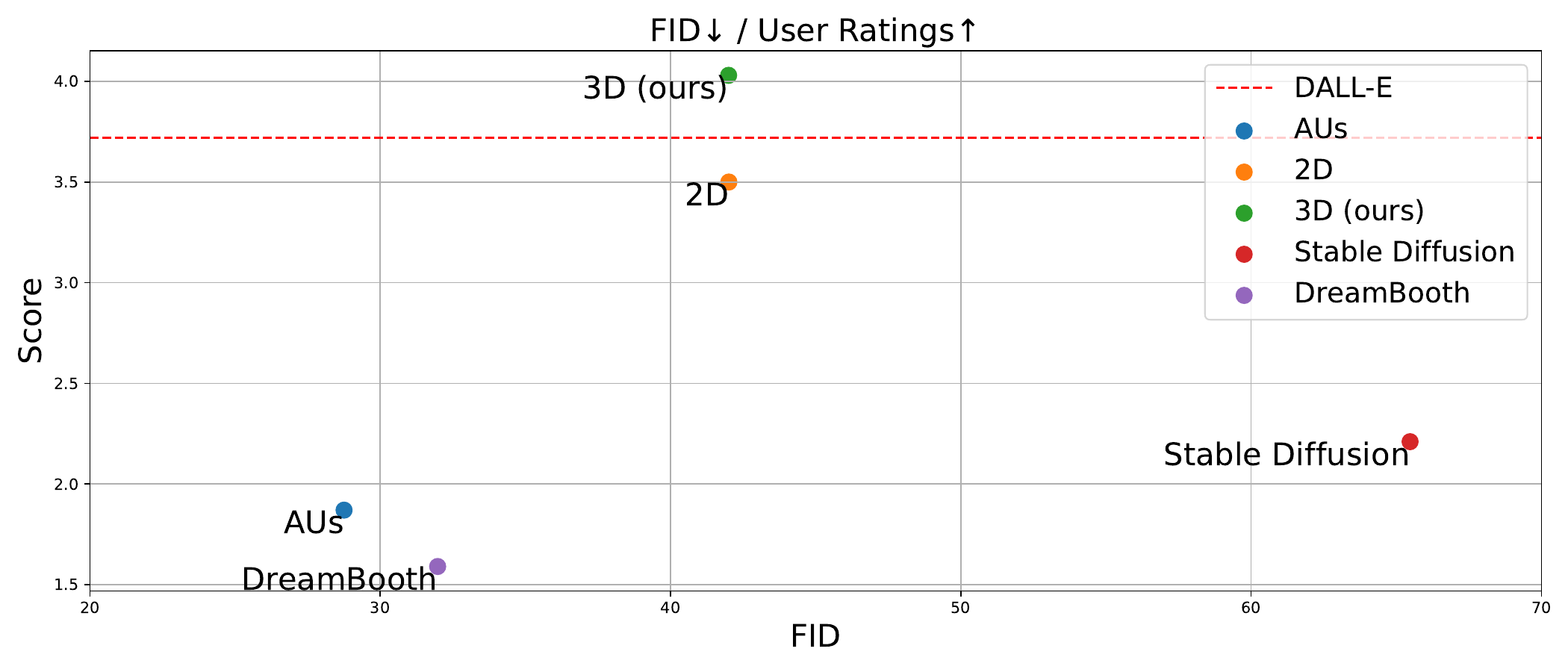}
    \vspace*{-3mm}
    \caption{FID and user ratings (Scores) for six different methods. Our 3D representation based method presents the best combination between low FID and high user rating score. Two methods with low FID often fail to represent the targeted expressions.}
    \label{fig:fid_user_plot}
\end{figure}

\begin{figure}
    \centering
    \vspace{-1mm}
    \includegraphics[width=0.44\textwidth]{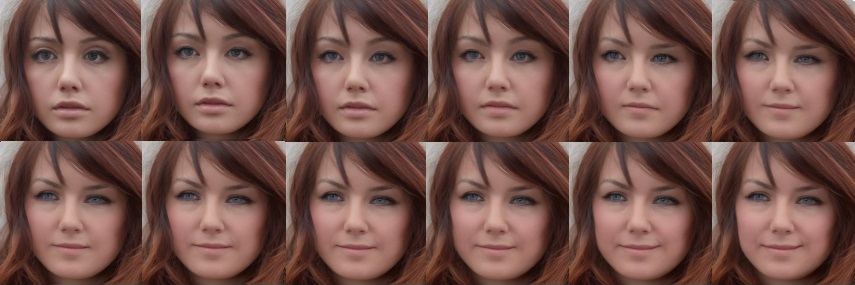}
    \includegraphics[width=0.44\textwidth]{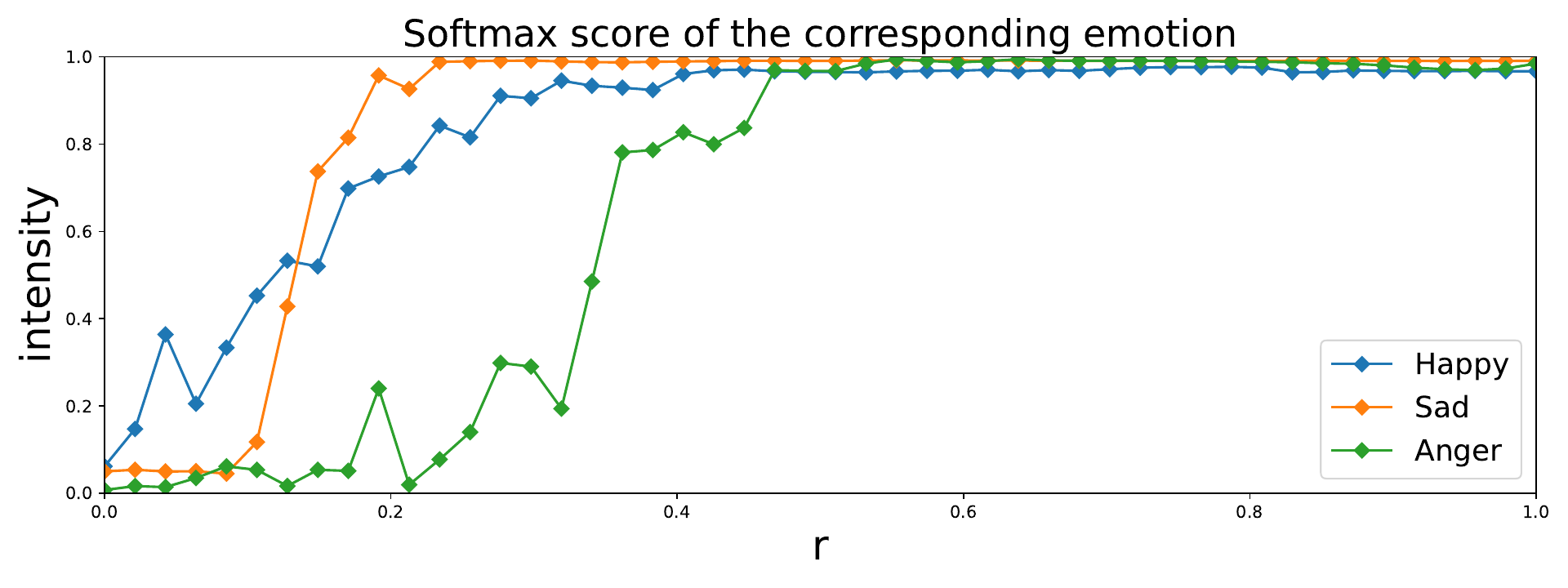}
     \vspace{-3mm}
    \caption{ Top: subtle control of expression between r=0.02 and r=0.3 along the ``happy" axis.  Bottom: softmax scores with increasing radius away from the neutral to 3 canonical emotions.}
     \vspace{-5mm}
    \label{fig:smooth}
    \vspace{0mm}
\end{figure}

\noindent\textbf{Baselines.} We use DreamBooth~\cite{dreambooth} trained on the AffectNet as our baseline for both quantitative and qualitative evaluations. Additionally, we use recent and popular image generation methods for further comparisons. More importantly, we perform exhaustive comparisons of our method that uses 2D AV representation against the proposed 3D representation. Please, refer to our supplementary material for more details, further analysis, and visualizations.
\subsection{Qualitative Results}
We present five sets of qualitative results, as depicted in Figures~\ref{fig:circle}, \ref{fig:basic}, \ref{fig:full}, \ref{fig:attributes}, and \ref{fig:smooth} demonstrating the efficacy of our methods. Figure~\ref{fig:basic} reveals that both our 2D and 3D models effectively interpret emotions represented as continuous numbers, with the 3D model exhibiting behavior towards the learned $Z$, indicating a continuous learned representation. Specifically, the top three rows of Figure~\ref{fig:basic} display images sampled around a circle at different learned $Z$ values of our 3D model, with AV values annotated above. The bottom part of the same figure contrasts this with the 2D model, using the same circle in the AV space to generate expressions. Notably, the 3D model's results are significantly superior to those of the 2D model, as they can be directly compared in the first and last rows.

Figure~\ref{fig:full} focuses on nine compound emotions that our 3D C2A2 model can represent but are beyond the scope of the 2D-AV model. This comparison, detailed in Table~\ref{tab:2d-3D_emotions}, underscores the superiority of our 3D-based representation, attributing to its richer representation and understanding of continuous numbers. For comparison, we also provide results obtained using Stable Diffusion~\cite{ldm}, DreamBooth~\cite{dreambooth}, and closed-source popular DALL-E 3.
Figure~\ref{fig:attributes} illustrates how our method preserves attributes from the base network, enabling meaningful number+text-to-image generation. Here, expressions for compound emotions, represented as numbers, are generated alongside text attributes, showcasing the versatility and robustness of our approach in handling complex emotional representations. Lastly, Figure~\ref{fig:smooth} illustrates an example of subtle control. 

\begin{table}[]
    \centering
    \begin{tabular}{|c|c|c|c|}
    \hline
    \textbf{Method} & \textbf{FED} ($\downarrow$) & \textbf{ERE} ($\downarrow$) & \textbf{SS} ($\downarrow$)   \\
    \hline
    \hline
    DreamBooth~\cite{dreambooth}& 61.147 &-- &--\\
    \hline
    2D-AV model & 20.347 &0.0774 &0.905 \\
    \hline
    3D model (Ours)& \textbf{16.060} &\textbf{0.0536} &\textbf{0.806}\\ 
    \hline
    \end{tabular}
 \vspace{-2mm}
    \caption{Fréchet Emotion Distance (FED), Emotion Reconstruction Error (ERE), and Smoothness Score (SS) for three different method. Our 3-dimensional representation of emotions clear outperform the other alternatives, when trained on the same settings.}
    \vspace{-5mm}
    \label{tab:qant_results}
\end{table}

\subsection{Quantitative Results}
We conduct two sets of quantitative evaluations. First we provide the results using the metric from GANmut~\cite{ganmut}. Later, the evaluation by the human psychologists is provided, along with FID, confirming the significance of our 3D representation model, by a large margin, in generating meaningful images. The user rating and FID trade-off is summarized in Figure~\ref{fig:fid_user_plot} for six different methods.

\noindent \textbf{FED, ERE, and Smoothness.}
Together with the FED, we use the Emotion Reconstruction Error (ERE) and the Smoothness Score (SS). To calculate FED, we sample 50K images and compute the FID using the emotion features. For ERE, we conduct a uniform search for target emotion, using a sample budget of 500 images. Multiple runs are performed for each of seven basic emotions. Then ERE is the averaged emotion reconstruction errors between target and closest images. The SS is determined by using  the VGGNet classifier with increasing emotion intensity. 
We follow ~\cite{ganmut} for these metrics, where readers can also find more details.

The FED, ERE, and SS, obtained by DreamBooth~\cite{dreambooth}, 2-dimensional AV based representation, and the proposed 3-dimensional representation are reported in Table~\ref{tab:qant_results}. Note that we train DreamBooth on AffectNet to obtain the reported results, which does not use 2D or 3D emotion labels. These results again highlighting the superiority of the our 3-dimensional C2A2 emotion model and method developed to perform number+text-to-image generation.

\noindent {\textbf{Expert user (psychologist) study.}}
We asked 8 expert psychologists to rate 315 images (per method) on the scale from 1 to 5 (5 being the highest) by how much they agree with the coordinates of the images. We also provided them with the nearest emotion to every image. More details of our study is in the supplementary materials. The obtained results are reported in Table~\ref{tab:qant_results_2}, which shows that the experts clearly prefer the images generated by the proposed 3D representation, for individual compound emotions as well as overall emotions. This further validates our emotion representation and expression generation methods.

\section{Conclusion}
We proposed a novel, unified and interpretable emotion representation that is capable of expressing nine additional compound emotions from the continuous 3-dimensional space. The continuous and 3D aspect of our representation allowed us to generate images with multitude of expressions. To facilitate such generation, we proposed two methods; one to recover the additional emotion axis $Z$, and another to generate images using the continuous vectors representing emotions in a DreamBooth-like setting. Both our qualitative and quantitative results showcase the superiority of the proposed emotion representation and the method for number+text-to-image generation. Furthermore, we showcase the capability of our method in generating compound expressions together with other facial attributes. We seek to extend our method along the temporal dimension by learning from the video examples as our future work. \\
\noindent {\textbf{Limitations and ethical statement.}} The emotion conditioning in our model is not entirely disentangled. This is particularly evident with attributes akin to age, shown in Figure~\ref{fig:basic}. Our method in processing text descriptions  beyond facial attributes needs further exploration. In this work, we controlled the expression without preserving the identity of the person. A future extension could benefit from adding identity preserving properties to our model. Central to our research ethos is a commitment to ethical and responsible data usage, with a strong focus on fostering socially responsible applications.

\noindent {\textbf{Acknowledgements.}}
This research was partially funded by the Ministry of Education and Science of Bulgaria (support for INSAIT, part of the Bulgarian National Roadmap for Research Infrastructure).
\label{sec:conclusion}

{
    \small
    \bibliographystyle{ieeenat_fullname}
    \bibliography{main}

\begin{thebibliography}{52}
\providecommand{\natexlab}[1]{#1}
\providecommand{\url}[1]{\texttt{#1}}
\expandafter\ifx\csname urlstyle\endcsname\relax
  \providecommand{\doi}[1]{doi: #1}\else
  \providecommand{\doi}{doi: \begingroup \urlstyle{rm}\Url}\fi

\bibitem[Acharya et~al.(2018)Acharya, Huang, Pani~Paudel, and Van~Gool]{acharya2018covariance}
Dinesh Acharya, Zhiwu Huang, Danda Pani~Paudel, and Luc Van~Gool.
\newblock Covariance pooling for facial expression recognition.
\newblock In \emph{Proceedings of the IEEE conference on computer vision and pattern recognition workshops}, pages 367--374, 2018.

\bibitem[Balaji et~al.(2022)Balaji, Nah, Huang, Vahdat, Song, Zhang, Kreis, Aittala, Aila, Laine, Catanzaro, Karras, and Liu]{balaji2022ediffi}
Yogesh Balaji, Seungjun Nah, Xun Huang, Arash Vahdat, Jiaming Song, Qinsheng Zhang, Karsten Kreis, Miika Aittala, Timo Aila, Samuli Laine, Bryan Catanzaro, Tero Karras, and Ming-Yu Liu.
\newblock ediff-i: Text-to-image diffusion models with an ensemble of expert denoisers, 2022.

\bibitem[Calder and Young(2016)]{calder2016understanding}
Andrew~J Calder and Andrew~W Young.
\newblock Understanding the recognition of facial identity and facial expression.
\newblock \emph{Facial Expression Recognition}, pages 41--64, 2016.

\bibitem[Du et~al.(2014)Du, Tao, and Martinez]{du2014compound}
Shichuan Du, Yong Tao, and Aleix~M Martinez.
\newblock Compound facial expressions of emotion.
\newblock \emph{Proceedings of the National Academy of Sciences}, 111\penalty0 (15):\penalty0 E1454--E1462, 2014.

\bibitem[d’Apolito{,} Danda Pani Paudel{,} Zhiwu Huang{,} Andres Romero{,} Luc Van~Gool(2021)]{ganmut}
Stefano d’Apolito{,} Danda Pani Paudel{,} Zhiwu Huang{,} Andres Romero{,} Luc Van~Gool.
\newblock Ganmut: Learning interpretable conditional space for gamut of emotions.
\newblock \emph{In Proceedings of the IEEE/CVF Conference on Computer Vision and Pattern Recognition}, pages 568--577, 2021.

\bibitem[Ekman and Friesen(1971)]{basicEmotion}
P Ekman and W Friesen.
\newblock Constants across cultures in the face and emotion.
\newblock \emph{Journal of Personality and Social Psychology}, 1971.

\bibitem[Ekman(1978)]{FacsProposal}
W.~V. Ekman, P.and~Friesen.
\newblock The facial actin coding system: a technique for the measurements of facial movements.
\newblock 1978.

\bibitem[et~al.(2018.{\natexlab{a}})]{stargan}
Choi{,}~Yunjey{,} et al.
\newblock Stargan: Unified generative adversarial networks for multi-domain image-to-image translation.
\newblock \emph{Proceedings of the IEEE conference on computer vision and pattern recognition.}, 2018.{\natexlab{a}}.

\bibitem[et~al.(2022.{\natexlab{a}})]{AUgraph}
Luo{,}~Cheng{,} et al.
\newblock Learning multi-dimensional edge feature-based au relation graph for facial action unit recognition.
\newblock \emph{arXiv preprint arXiv:2205.01782}, 2022.{\natexlab{a}}.

\bibitem[et~al.(2018.{\natexlab{b}})]{ganimation}
Pumarola{,}~Albert{,} et al.
\newblock Ganimation: Anatomically-aware facial animation from a single image.
\newblock \emph{Proceedings of the European conference on computer vision (ECCV).}, 2018.{\natexlab{b}}.

\bibitem[et~al.(2022.{\natexlab{b}})]{ned}
Papantoniou{,} Foivos~Paraperas{,} et al.
\newblock Neural emotion director: Speech-preserving semantic control of facial expressions in" in-the-wild" videos.
\newblock \emph{Proceedings of the IEEE/CVF Conference on Computer Vision and Pattern Recognition.}, 2022.{\natexlab{b}}.

\bibitem[et~al.(2018.{\natexlab{c}})]{moods}
Vielzeuf{,}~Valentin{,} et al.
\newblock The many moods of emotion.
\newblock \emph{arXiv preprint arXiv:1810.13197}, 2018.{\natexlab{c}}.

\bibitem[Feng et~al.(2022)Feng, He, Fu, Jampani, Akula, Narayana, Basu, Wang, and Wang]{feng2022training}
Weixi Feng, Xuehai He, Tsu-Jui Fu, Varun Jampani, Arjun Akula, Pradyumna Narayana, Sugato Basu, Xin~Eric Wang, and William~Yang Wang.
\newblock Training-free structured diffusion guidance for compositional text-to-image synthesis.
\newblock \emph{arXiv preprint arXiv:2212.05032}, 2022.

\bibitem[Gal et~al.(2022)Gal, Alaluf, Atzmon, Patashnik, Bermano, Chechik, and Cohen-Or]{t2i}
Rinon Gal, Yuval Alaluf, Yuval Atzmon, Or Patashnik, Amit~H Bermano, Gal Chechik, and Daniel Cohen-Or.
\newblock An image is worth one word: Personalizing text-to-image generation using textual inversion.
\newblock \emph{arXiv preprint arXiv:2208.01618}, 2022.

\bibitem[Goodfellow et~al.(2020)Goodfellow, Pouget-Abadie, Mirza, Xu, Warde-Farley, Ozair, Courville, and Bengio]{goodfellow2020generative}
Ian Goodfellow, Jean Pouget-Abadie, Mehdi Mirza, Bing Xu, David Warde-Farley, Sherjil Ozair, Aaron Courville, and Yoshua Bengio.
\newblock Generative adversarial networks.
\newblock \emph{Communications of the ACM}, 63\penalty0 (11):\penalty0 139--144, 2020.

\bibitem[Gu et~al.(2022)Gu, Chen, Bao, Wen, Zhang, Chen, Yuan, and Guo]{gu2022vector}
Shuyang Gu, Dong Chen, Jianmin Bao, Fang Wen, Bo Zhang, Dongdong Chen, Lu Yuan, and Baining Guo.
\newblock Vector quantized diffusion model for text-to-image synthesis.
\newblock In \emph{Proceedings of the IEEE/CVF Conference on Computer Vision and Pattern Recognition}, pages 10696--10706, 2022.

\bibitem[Hasani{,} and Mahoor.(2017.)]{affectnet}
Mollahosseini{,} Ali{,}~Behzad Hasani{,} and Mohammad~H. Mahoor.
\newblock Affectnet: A database for facial expression, valence, and arousal computing in the wild.
\newblock \emph{IEEE Transactions on Affective Computing 10.1}, pages 18--31., 2017.

\bibitem[Heusel et~al.(2017)Heusel, Ramsauer, Unterthiner, Nessler, and Hochreiter]{conf/nips/HeuselRUNH17}
Martin Heusel, Hubert Ramsauer, Thomas Unterthiner, Bernhard Nessler, and Sepp Hochreiter.
\newblock Gans trained by a two time-scale update rule converge to a local nash equilibrium.
\newblock In \emph{NIPS}, pages 6626--6637, 2017.

\bibitem[Ho et~al.(2020)Ho, Jain, and Abbeel]{ddpm}
Jonathan Ho, Ajay Jain, and Pieter Abbeel.
\newblock Denoising diffusion probabilistic models.
\newblock \emph{Advances in neural information processing systems}, 33:\penalty0 6840--6851, 2020.

\bibitem[Kannala{,} and Rahtu.(2020.)]{icface}
Tripathy{,} Soumya{,}~Juho Kannala{,} and Esa Rahtu.
\newblock Icface: Interpretable and controllable face reenactment using gans.
\newblock \emph{Proceedings of the IEEE/CVF winter conference on applications of computer vision.}, 2020.

\bibitem[Kim and Song(2022)]{kim2022emotion}
Daeha Kim and Byung~Cheol Song.
\newblock Emotion-aware multi-view contrastive learning for facial emotion recognition.
\newblock In \emph{European Conference on Computer Vision}, pages 178--195. Springer, 2022.

\bibitem[Kraut and Johnston(1979)]{Kraut1979SocialAE}
Robert~E. Kraut and Robert~E. Johnston.
\newblock Social and emotional messages of smiling: An ethological approach.
\newblock 1979.

\bibitem[Kumari et~al.(2023)Kumari, Zhang, Zhang, Shechtman, and Zhu]{Kumari_2023}
Nupur Kumari, Bingliang Zhang, Richard Zhang, Eli Shechtman, and Jun-Yan Zhu.
\newblock Multi-concept customization of text-to-image diffusion.
\newblock In \emph{2023 IEEE/CVF Conference on Computer Vision and Pattern Recognition (CVPR)}. IEEE, 2023.

\bibitem[Larsen and Diener(1992)]{larsen1992promises}
Randy~J Larsen and Edward Diener.
\newblock Promises and problems with the circumplex model of emotion.
\newblock 1992.

\bibitem[Lee et~al.(2023)Lee, Liu, Ryu, Watkins, Du, Boutilier, Abbeel, Ghavamzadeh, and Gu]{lee2023aligning}
Kimin Lee, Hao Liu, Moonkyung Ryu, Olivia Watkins, Yuqing Du, Craig Boutilier, Pieter Abbeel, Mohammad Ghavamzadeh, and Shixiang~Shane Gu.
\newblock Aligning text-to-image models using human feedback.
\newblock \emph{arXiv preprint arXiv:2302.12192}, 2023.

\bibitem[Liu et~al.(2022)Liu, Li, Du, Torralba, and Tenenbaum]{liu2022compositional}
Nan Liu, Shuang Li, Yilun Du, Antonio Torralba, and Joshua~B Tenenbaum.
\newblock Compositional visual generation with composable diffusion models.
\newblock In \emph{European Conference on Computer Vision}, pages 423--439. Springer, 2022.

\bibitem[Motamed et~al.(2023)Motamed, Paudel, and Gool]{motamed2023lego}
Saman Motamed, Danda~Pani Paudel, and Luc~Van Gool.
\newblock Lego: Learning to disentangle and invert concepts beyond object appearance in text-to-image diffusion models, 2023.

\bibitem[Nichol et~al.(2021{\natexlab{a}})Nichol, Dhariwal, Ramesh, Shyam, Mishkin, McGrew, Sutskever, and Chen]{glide}
Alex Nichol, Prafulla Dhariwal, Aditya Ramesh, Pranav Shyam, Pamela Mishkin, Bob McGrew, Ilya Sutskever, and Mark Chen.
\newblock Glide: Towards photorealistic image generation and editing with text-guided diffusion models.
\newblock \emph{arXiv preprint arXiv:2112.10741}, 2021{\natexlab{a}}.

\bibitem[Nichol et~al.(2021{\natexlab{b}})Nichol, Dhariwal, Ramesh, Shyam, Mishkin, McGrew, Sutskever, and Chen]{nichol2021glide}
Alex Nichol, Prafulla Dhariwal, Aditya Ramesh, Pranav Shyam, Pamela Mishkin, Bob McGrew, Ilya Sutskever, and Mark Chen.
\newblock Glide: Towards photorealistic image generation and editing with text-guided diffusion models.
\newblock \emph{arXiv preprint arXiv:2112.10741}, 2021{\natexlab{b}}.

\bibitem[Paul{,} and Friesen(1978)]{FACS}
Ekman{,} Paul{,} and Wallace~V. Friesen.
\newblock Facial action coding system.
\newblock 1978.

\bibitem[Radford et~al.(2021)Radford, Kim, Hallacy, Ramesh, Goh, Agarwal, Sastry, Askell, Mishkin, Clark, et~al.]{clip}
Alec Radford, Jong~Wook Kim, Chris Hallacy, Aditya Ramesh, Gabriel Goh, Sandhini Agarwal, Girish Sastry, Amanda Askell, Pamela Mishkin, Jack Clark, et~al.
\newblock Learning transferable visual models from natural language supervision.
\newblock In \emph{International conference on machine learning}, pages 8748--8763. PMLR, 2021.

\bibitem[Ramesh et~al.(2021)Ramesh, Pavlov, Goh, Gray, Voss, Radford, Chen, and Sutskever]{ramesh2021zero}
Aditya Ramesh, Mikhail Pavlov, Gabriel Goh, Scott Gray, Chelsea Voss, Alec Radford, Mark Chen, and Ilya Sutskever.
\newblock Zero-shot text-to-image generation.
\newblock In \emph{International Conference on Machine Learning}, pages 8821--8831. PMLR, 2021.

\bibitem[Ramesh et~al.(2022)Ramesh, Dhariwal, Nichol, Chu, and Chen]{ramesh2022hierarchical}
Aditya Ramesh, Prafulla Dhariwal, Alex Nichol, Casey Chu, and Mark Chen.
\newblock Hierarchical text-conditional image generation with clip latents.
\newblock \emph{arXiv preprint arXiv:2204.06125}, 1\penalty0 (2):\penalty0 3, 2022.

\bibitem[Rombach et~al.(2022{\natexlab{a}})Rombach, Blattmann, Lorenz, Esser, and Ommer]{ldm}
Robin Rombach, Andreas Blattmann, Dominik Lorenz, Patrick Esser, and Bj{\"o}rn Ommer.
\newblock High-resolution image synthesis with latent diffusion models.
\newblock In \emph{Proceedings of the IEEE/CVF conference on computer vision and pattern recognition}, pages 10684--10695, 2022{\natexlab{a}}.

\bibitem[Rombach et~al.(2022{\natexlab{b}})Rombach, Blattmann, Lorenz, Esser, and Ommer]{rombach2022high}
Robin Rombach, Andreas Blattmann, Dominik Lorenz, Patrick Esser, and Bj{\"o}rn Ommer.
\newblock High-resolution image synthesis with latent diffusion models.
\newblock In \emph{Proceedings of the IEEE/CVF conference on computer vision and pattern recognition}, pages 10684--10695, 2022{\natexlab{b}}.

\bibitem[Rosenwein(2010)]{rosenwein2010problems}
Barbara~H Rosenwein.
\newblock Problems and methods in the history of emotions.
\newblock \emph{Passions in context}, 1\penalty0 (1):\penalty0 1--32, 2010.

\bibitem[Ruiz et~al.(2023)Ruiz, Li, Jampani, Pritch, Rubinstein, and Aberman]{dreambooth}
Nataniel Ruiz, Yuanzhen Li, Varun Jampani, Yael Pritch, Michael Rubinstein, and Kfir Aberman.
\newblock Dreambooth: Fine tuning text-to-image diffusion models for subject-driven generation.
\newblock In \emph{Proceedings of the IEEE/CVF Conference on Computer Vision and Pattern Recognition}, pages 22500--22510, 2023.

\bibitem[Russell(1980)]{Circumplex}
James Russell.
\newblock A circumplex model of affect.
\newblock \emph{Journal of Personality and Social Psychology}, 39:\penalty0 1161--1178, 1980.

\bibitem[Russell and Mehrabian(1977)]{PADrusselmehrabis}
James Russell and Albert Mehrabian.
\newblock Evidence for a three-factor theory of emotions.
\newblock \emph{Journal of Research in Personality}, 11:\penalty0 273--294, 1977.

\bibitem[Russell et~al.(1989)Russell, Lewicka, and Niit]{cross_culture}
James Russell, Maria Lewicka, and Toomas Niit.
\newblock A cross-cultural study of a circumplex model of affect.
\newblock \emph{Journal of Personality and Social Psychology}, 57:\penalty0 848--856, 1989.

\bibitem[Russell et~al.(2003)Russell, Bachorowski, and Fernandez-Dols]{EmotionExpression}
James Russell, Jo-Anne Bachorowski, and José-Miguel Fernandez-Dols.
\newblock Facial and vocal expressions of emotion.
\newblock \emph{Annual Review of Psychology}, 54:\penalty0 329--349, 2003.

\bibitem[Saharia et~al.(2022)Saharia, Chan, Saxena, Li, Whang, Denton, Ghasemipour, Gontijo~Lopes, Karagol~Ayan, Salimans, et~al.]{saharia2022photorealistic}
Chitwan Saharia, William Chan, Saurabh Saxena, Lala Li, Jay Whang, Emily~L Denton, Kamyar Ghasemipour, Raphael Gontijo~Lopes, Burcu Karagol~Ayan, Tim Salimans, et~al.
\newblock Photorealistic text-to-image diffusion models with deep language understanding.
\newblock \emph{Advances in Neural Information Processing Systems}, 35:\penalty0 36479--36494, 2022.

\bibitem[Scherer et~al.(2000)]{scherer2000psychological}
Klaus~R Scherer et~al.
\newblock Psychological models of emotion.
\newblock \emph{The neuropsychology of emotion}, 137\penalty0 (3):\penalty0 137--162, 2000.

\bibitem[Simonyan and Zisserman(2014)]{VGG}
Karen Simonyan and Andrew Zisserman.
\newblock Very deep convolutional networks for large-scale image recognition.
\newblock \emph{arXiv 1409.1556}, 2014.

\bibitem[Sohl-Dickstein et~al.(2015)Sohl-Dickstein, Weiss, Maheswaranathan, and Ganguli]{sohl2015deep}
Jascha Sohl-Dickstein, Eric Weiss, Niru Maheswaranathan, and Surya Ganguli.
\newblock Deep unsupervised learning using nonequilibrium thermodynamics.
\newblock In \emph{International conference on machine learning}, pages 2256--2265. PMLR, 2015.

\bibitem[Song et~al.(2020{\natexlab{a}})Song, Meng, and Ermon]{song2020denoising}
Jiaming Song, Chenlin Meng, and Stefano Ermon.
\newblock Denoising diffusion implicit models.
\newblock \emph{arXiv preprint arXiv:2010.02502}, 2020{\natexlab{a}}.

\bibitem[Song et~al.(2020{\natexlab{b}})Song, Sohl-Dickstein, Kingma, Kumar, Ermon, and Poole]{song2020score}
Yang Song, Jascha Sohl-Dickstein, Diederik~P Kingma, Abhishek Kumar, Stefano Ermon, and Ben Poole.
\newblock Score-based generative modeling through stochastic differential equations.
\newblock \emph{arXiv preprint arXiv:2011.13456}, 2020{\natexlab{b}}.

\bibitem[Tang et~al.(2023)Tang, Yuan, and Zhang]{tang2023emotional}
Lilu Tang, Peijun Yuan, and Dan Zhang.
\newblock Emotional experience during human-computer interaction: A survey.
\newblock \emph{International Journal of Human--Computer Interaction}, pages 1--11, 2023.

\bibitem[Wu et~al.(2023)Wu, Liu, Zhao, Kale, Bui, Yu, Lin, Zhang, and Chang]{Wu_2023_CVPR}
Qiucheng Wu, Yujian Liu, Handong Zhao, Ajinkya Kale, Trung Bui, Tong Yu, Zhe Lin, Yang Zhang, and Shiyu Chang.
\newblock Uncovering the disentanglement capability in text-to-image diffusion models.
\newblock In \emph{Proceedings of the IEEE/CVF Conference on Computer Vision and Pattern Recognition (CVPR)}, pages 1900--1910, 2023.

\bibitem[Xia et~al.(2022)Xia, Zhang, Yang, Xue, Zhou, and Yang]{xia2022gan}
Weihao Xia, Yulun Zhang, Yujiu Yang, Jing-Hao Xue, Bolei Zhou, and Ming-Hsuan Yang.
\newblock Gan inversion: A survey.
\newblock \emph{IEEE Transactions on Pattern Analysis and Machine Intelligence}, 45\penalty0 (3):\penalty0 3121--3138, 2022.

\bibitem[Yu et~al.(2022)Yu, Xu, Koh, Luong, Baid, Wang, Vasudevan, Ku, Yang, Ayan, et~al.]{yu2022scaling}
Jiahui Yu, Yuanzhong Xu, Jing~Yu Koh, Thang Luong, Gunjan Baid, Zirui Wang, Vijay Vasudevan, Alexander Ku, Yinfei Yang, Burcu~Karagol Ayan, et~al.
\newblock Scaling autoregressive models for content-rich text-to-image generation.
\newblock \emph{arXiv preprint arXiv:2206.10789}, 2\penalty0 (3):\penalty0 5, 2022.

\bibitem[Zou et~al.(2023)Zou, Faisan, Yu, Valette, and Seo]{zou20234d}
Kaifeng Zou, Sylvain Faisan, Boyang Yu, S{\'e}bastien Valette, and Hyewon Seo.
\newblock 4d facial expression diffusion model.
\newblock \emph{arXiv preprint arXiv:2303.16611}, 2023.

\end{thebibliography}
}

\end{document}